%% file: main.tex
\documentclass[11pt]{article}

\usepackage[margin=1in]{geometry}
\usepackage{graphicx}
\usepackage{tikz}
\usetikzlibrary{arrows.meta,positioning,shapes.geometric,calc,fit,backgrounds}
\usepackage{booktabs}
\usepackage{array}
\usepackage{longtable}
\usepackage{amsmath}
\usepackage{xurl}
\usepackage{fvextra}
\usepackage{microtype}
\usepackage[section]{placeins}
\usepackage[numbers,sort&compress]{natbib}
\usepackage[colorlinks=true,linkcolor=blue,citecolor=blue,urlcolor=blue]{hyperref}
\DefineVerbatimEnvironment{CodeBlock}{Verbatim}{fontsize=\small,breaklines=true,breakanywhere=true}
\newcommand{\filepath}[1]{\path{#1}}
\newcommand{\metricbreak}[2]{\shortstack[l]{\texttt{#1}\\\texttt{#2}}}
\newcommand{\metricbreakthree}[3]{\shortstack[l]{\texttt{#1}\\\texttt{#2}\\\texttt{#3}}}
\newcolumntype{L}[1]{>{\raggedright\arraybackslash}p{#1}}

\newcommand{\bench}{DreamBench-SWE}

\input{analysis/fold/v21_external_audit_macros}

\title{\bench{}: A Multi-Session Memory-Hygiene Benchmark for Software Agents}
\author{Sarthak Singh\\Independent Researcher}
\date{}

\begin{document}

\maketitle

\input{sections/01_abstract}
\input{sections/02_introduction}
\input{sections/03_related_work}
\input{sections/04_problem_formulation}
\input{sections/05_method}
\input{sections/06_experiments}
\input{sections/07_results}

\input{sections/08_analysis}
\input{sections/09_limitations}
\input{sections/10_conclusion}

\FloatBarrier
\bibliographystyle{plainnat}
\bibliography{bibliography/sources}

\appendix
\input{sections/appendix_verbatim}
\input{sections/appendix_artifact}
\input{sections/appendix_v2_protocol}
\input{sections/appendix_construct_taxonomy}
\input{sections/appendix_clustered_stats}
\input{sections/appendix_reproducibility}
\input{sections/appendix_notation}
\input{sections/appendix_ethics_impact}

\end{document}

%% file: analysis/fold/v21_external_audit_macros.tex
\newcommand{\vOneCompletedUnits}{360}
\newcommand{\vOnePlannedUnits}{360}
\newcommand{\vOneValidCells}{720}
\newcommand{\vOnePlannedCells}{720}
\newcommand{\vOneBZeroPassed}{21}
\newcommand{\vOneBZeroValid}{180}
\newcommand{\vOneBZeroRate}{0.1167}
\newcommand{\vOneBZeroCILow}{0.0348}
\newcommand{\vOneBZeroCIHigh}{0.1986}
\newcommand{\vOneBFivePassed}{82}
\newcommand{\vOneBFiveValid}{180}
\newcommand{\vOneBFiveRate}{0.4556}
\newcommand{\vOneBFiveCILow}{0.3304}
\newcommand{\vOneBFiveCIHigh}{0.5807}
\newcommand{\vOneDFHybridPassed}{83}
\newcommand{\vOneDFHybridValid}{180}
\newcommand{\vOneDFHybridRate}{0.4611}
\newcommand{\vOneDFHybridCILow}{0.3545}
\newcommand{\vOneDFHybridCIHigh}{0.5677}
\newcommand{\vOneMemZeroLitPassed}{97}
\newcommand{\vOneMemZeroLitValid}{180}
\newcommand{\vOneMemZeroLitRate}{0.5389}
\newcommand{\vOneMemZeroLitCILow}{0.4165}
\newcommand{\vOneMemZeroLitCIHigh}{0.6613}
\newcommand{\vOneBFiveBZeroSigned}{+61}
\newcommand{\vOneBFiveBZeroRawP}{0.000324063}
\newcommand{\vOneBFiveBZeroHolmP}{0.00129625}
\newcommand{\vOneDFHybridBZeroSigned}{+62}
\newcommand{\vOneDFHybridBZeroRawP}{1.56049e-06}
\newcommand{\vOneDFHybridBZeroHolmP}{9.36294e-06}
\newcommand{\vOneMemZeroLitBZeroSigned}{+76}
\newcommand{\vOneMemZeroLitBZeroRawP}{7.85204e-06}
\newcommand{\vOneMemZeroLitBZeroHolmP}{3.92602e-05}
\newcommand{\vOneLitBFiveSigned}{+15}
\newcommand{\vOneLitBFiveRawP}{0.0273438}
\newcommand{\vOneLitBFiveMajorityP}{0.21875}
\newcommand{\vOneLitBFiveSeedOneP}{0.21875}
\newcommand{\vOneLitBFiveSeedTwoP}{0.03125}
\newcommand{\vOneLitBFiveSeedThreeP}{0.125}
\newcommand{\vOneLitDFRawP}{0.162649}
\newcommand{\vOneLitDFMajorityP}{0.607239}
\newcommand{\vOneBFiveDrift}{0.0389}
\newcommand{\vOneDFHybridDrift}{0.0667}
\newcommand{\vOneCnineBZeroPassed}{12}
\newcommand{\vOneCnineBZeroValid}{12}
\newcommand{\vOneCtenBZeroPassed}{6}
\newcommand{\vOneCtenBZeroValid}{6}

%% file: sections/01_abstract.tex
\begin{abstract}
DreamBench-SWE is a multi-session benchmark for software-agent memory hygiene in
which later software tasks depend on non-inferable evidence from earlier sessions and
are scored by executable hidden oracles.  We report the original scaled v2 fold and a
separately preregistered v2.1 successor audit designed after that study but frozen before
successor outcome inspection.  The successor run completed
\vOneCompletedUnits/\vOnePlannedUnits{} work units and
\vOneValidCells/\vOnePlannedCells{} S3 cells across four conditions.  In the original
fold, the primary DF-hybrid--B5 contrast was null (95/180 versus 89/180;
clustered p=.518, Holm p=1), not evidence of equivalence, and C9/C10 retained
B0-headroom limitations.  In the successor, no external memory
achieved \vOneBZeroPassed/\vOneBZeroValid{} passes (rate \vOneBZeroRate),
deterministic verbatim event memory \vOneBFivePassed/\vOneBFiveValid{}
(rate \vOneBFiveRate), the typed-plus-raw reference probe
\vOneDFHybridPassed/\vOneDFHybridValid{} (rate \vOneDFHybridRate), and one pinned hosted
Mem0 literal-storage configuration \vOneMemZeroLitPassed/\vOneMemZeroLitValid{}
(rate \vOneMemZeroLitRate).  The registered six-slot Family A
retained unavailable slots at p=1; all three available comparisons against no memory
rejected after Holm correction.  Both preregistered mechanism contrasts were unavailable
after pre-evaluation conformance rejection.  The secondary literal-storage-versus-verbatim
comparison was nonconfirmatory and sensitivity-dependent, while the comparison with the
reference probe did not reject.  The audit therefore supports DreamBench-SWE as a
discriminating executable profile benchmark and characterizes one exact hosted-memory
configuration, but it does not establish an external-system mechanism, superiority among
memory-bearing conditions, equivalence, or broad product generality.  The original
v2.0.5 findings and artifacts remain unchanged.
\end{abstract}

%% file: sections/02_introduction.tex
\section{Introduction}

Software-engineering agents increasingly operate across repeated sessions on the same
repository. A later run may need to remember that a generated file is not
source-of-truth, that a reviewer preference applies only to one module, that a test
failure was flaky rather than causal, or that an older architecture fact was superseded
by a recent migration. In such settings, memory quality is not the same as memory
volume. A memory system can harm the agent when it retrieves a similar but stale fact,
when it overgeneralizes feedback beyond its scope, or when it turns an uncertain failure
diagnosis into an active instruction.

This paper studies memory hygiene as a benchmarkable failure mode for multi-session
software-engineering agents.  We use ``sleep'' and ``dream'' only as operational labels:
a sleep phase is an offline computation over logged trajectories and memory state, and a
dream artifact is a derived replay or repair artifact produced by that computation.  The
contribution is not biological analogy.  The paper is anchored on \bench{}, a controlled
benchmark whose traps require earlier-session memory and whose S3 outcomes are scored by
executable oracles.  We use an evaluated maintenance pipeline as a reference probe for studying how raw
evidence, typed consolidation, contradiction repair, counterfactual replay, stale-memory
suppression, and retrieval gating affect behavior on that benchmark.

The reference probe system is built around one invariant: raw episodes are never
destroyed.  Raw trajectories contain the task prompt, repository state, tool calls,
command output, diffs, memory reads, memory writes, feedback, and outcomes.  Sleep-phase
operators may create derived memories, mark them stale, supersede them, or block them
from retrieval, but they do not erase the evidence from which those derived memories
came.  This invariant makes the probe auditable, supports episodic-only controls, and
treats consolidation as a risky derived operation rather than an automatic improvement.

A typical trap is deliberately simple but hard to infer: S1 or S2 records a reviewer rule
such as ``future exports in this module must use this exact private dialect marker,''
while S3 asks for an implementation that only passes if the marker is reproduced
byte-for-byte.  The marker is hidden from the S3 prompt and absent from the wake-agent
container filesystem, so success requires memory.  Other traps test superseded
architecture facts, generated-file boundaries, scoped reviewer feedback, spurious
failure lessons, or abstention after insufficient evidence.  This design makes the
memory dependency executable and auditable, but the v2 fold also shows where the
instrument must be narrowed: C9 and C10 reveal B0-headroom limitations rather than broad
anti-hoarding validity.

The paper is organized around five research questions:

\begin{enumerate}
  \item \textbf{RQ1: Benchmark.} Can a controlled multi-session SWE benchmark expose
  memory failures that are hidden from single-session task success?
  \item \textbf{RQ2: Memory-system ordering.} Under the frozen v2 trap-clustered P1
  test, how does a hybrid typed-plus-raw reference probe compare with a strong verbatim
  event-memory system?
  \item \textbf{RQ3: Structured memory anchor.} Does typed provenance-linked memory
  separate from reflection-only vector retrieval, and what caveats survive clustered
  sensitivity?
  \item \textbf{RQ4: Hygiene diagnostics.} Which executable trap-tripped diagnostics
  and offline hygiene metrics explain memory-system failures on the benchmark?
  \item \textbf{RQ5: Successor external audit.} In a separately preregistered audit,
  do concurrent memory-bearing controls and an admitted external-system configuration
  separate from no memory, and do the frozen cross-configuration mechanism contrasts
  remain evaluable?
\end{enumerate}

Table~\ref{tab:intro-verdicts} summarizes the original v1 hypothesis family and the v2
re-anchor.  The v1 negative result remains explicit: H1, interpreted as ``reference-probe hybrid
separates from the strong verbatim B5 baseline,'' is \emph{not established} in the v1
fold.  The scaled v2 primary clustered P1 result is also null: reference-probe hybrid is
$95/180=0.528$ and B5 is $89/180=0.494$, with signed statistic $+6$, permutation
$p{=}0.518$, Holm $p{=}1.0$, and reject=false.  This is a failure to reject, not
equivalence.  Whether a future expansion reverses that outcome or not, P1 is evidence
about memory-system behavior on \bench{}, not a replacement for the benchmark headline.

\begin{table}[!htbp]
\centering
\scriptsize
\resizebox{\textwidth}{!}{%
\begin{tabular}{p{0.18\linewidth}p{0.14\linewidth}p{0.31\linewidth}p{0.25\linewidth}p{0.08\linewidth}}
\toprule
\textbf{Claim family} & \textbf{Tested?} & \textbf{Evidence} & \textbf{Result} & \textbf{Section} \\
\midrule
Benchmark validity & V2 fold complete & 60 traps, 3 seeds, 1890 condition-level result files; admission funnel 30 authored, 30 dry-valid, 30 live-valid, 30 admitted & Completed measurement fold, narrowed by C9 and C10 B0-headroom failures & \ref{sec:results} \\
H1 / v2 P1: hybrid typed-plus-raw memory vs.\ strong verbatim event memory & V1 tested; v2 clustered P1 tested & V1 reference-probe hybrid $52/66=0.788$ vs.\ B5 $48/66=0.727$, Holm $p{=}1.0$; v2 reference-probe hybrid $95/180=0.528$ vs.\ B5 $89/180=0.494$, signed $+6$, permutation $p{=}0.518$, Holm $p{=}1.0$ & Separation not established; failure to reject, not equivalence & \ref{sec:results-mcnemar} \\
Structured memory anchor & V1 and v2 tested & V1 typed-only vs.\ B3 positive with clustered caveats; v2 DF vs.\ B3 signed $+13$, permutation $p{=}0.279$, Holm $p{=}1.0$ & V2 does not reject; v1 anchor remains historical, not headline & \ref{sec:results-mcnemar} \\
Hygiene diagnostics & Diagnostic & V2 oracle metrics: \texttt{IrrelevantImportRate} $1434/3024=0.474$, \texttt{OverscopeRate} $378/2268=0.167$, \texttt{RepeatedErrorRate} $1512/2646=0.571$, \texttt{StaleUseRate} $756/3780=0.200$ & Executable diagnostics expose failure modes; no judge-based hygiene headline & \ref{sec:results-hygiene} \\
Cost and resource diagnostics & Diagnostic only & V2 cost per successful task: reference-probe raw-only $0.0356$, reference-probe hybrid $0.0383$, B5 $0.0395$ & Resource evidence constrains interpretation; it is not a headline claim & \ref{sec:limitations} \\
V2.1 successor discrimination & Separate preregistered audit & Family A: B5, DF-hybrid, and B5-MEM0-LIT each reject versus B0 after fixed-six-slot Holm correction; Family B has no available contrast & Benchmark discrimination and one bounded external profile; no mechanism, superiority, equivalence, or product ranking & \ref{sec:results-v21-external} \\
\bottomrule
\end{tabular}
}%
\caption{Re-anchored claim ledger.  Confirmatory claims are limited to canonical analyzer outputs; nulls are failures to reject, not equivalence claims.}
\label{tab:intro-verdicts}
\end{table}

We make three contributions, and only the first is the headline.  \textbf{First},
\bench{} is a dynamic memory-hygiene benchmark for multi-session software tasks.
\bench{} consists of hand-curated memory traps: each S3 session turns on hidden,
non-inferable information recoverable only from earlier-session memory and is scored by
an executable oracle.  Wake agents and sleep judges run in containers whose filesystems
exclude the benchmark repository, hidden oracles, reference solutions, and sequence
records; we do not claim network isolation.  All runs are validity-gated and
contamination-scanner-reviewed.  This benchmark stands regardless of the measured
memory-system ordering, but the v2 fold narrows its anti-hoarding claims where C9 and
C10 fail B0-headroom criteria.  \textbf{Second}, the reference probe is a
wake/sleep/read architecture for training-free external memory maintenance around a
fixed software-engineering agent, evaluated through typed-only, raw-only, and hybrid
variants.  \textbf{Third}, we report the v1 fold and the scaled v2 fold.  The v2 fold
has a null primary P1 comparison, non-rejections across P2--P6, a clean admission
funnel, explicit C9/C10 validity caveats, and executable hygiene diagnostics.

The v2.1 successor audit answers RQ5 without retroactively changing the original v2
family.  Its complete canonical run separates both concurrent positive controls and the
one admitted hosted literal-storage configuration from B0 after fixed-six-slot Holm
correction.  The registered mechanism family is unavailable because its counterpart
conditions failed frozen pre-evaluation conformance.  The resulting claim is benchmark
discrimination plus one bounded external-system profile, not a mechanism, product
ranking, superiority claim, or equivalence result
(Section~\ref{sec:results-v21-external}).

%% file: sections/03_related_work.tex
\section{Related Work}

\paragraph{Evaluation gap.}
\bench{} is closest in spirit to SWE-bench, SWE-bench Verified, SWE-Bench Pro,
SWE-agent, and Meta-Harness, but those benchmarks and harness studies evaluate
issue resolution, agent-computer interfaces, or harness optimization rather than
longitudinal memory maintenance
\citep{jimenez2024swebench,openai2024swebenchverified,deng2025swebenchpro,yang2024sweagent,lee2026metaharnessendtoendoptimizationmodel}.
Conversational and incremental-memory benchmarks such as LoCoMo, LongMemEval,
BEAM, AMA-Bench, and MemoryAgentBench test long-history recall, temporal reasoning,
updates, selective forgetting, and dialogue or agentic QA
\citep{maharana2024locomo,wu2025longmemeval,tavakoli2025beam,zhao2026amabench,hu2025memoryagentbench}.
More recent benchmarks couple memory to action.  MemoryArena uses interdependent
multi-session tasks across web navigation, planning, search, and formal reasoning;
EvoMemBench spans in- and cross-episode, knowledge- and execution-oriented settings
and compares 15 memory methods; WorldMemArena localizes multimodal memory failures
across writing, maintenance, retrieval, and use; and STATE-Bench provides stateful
enterprise workflows plus an Agent Learning track
\citep{he2026memoryarena,wang2026evomembench,liu2026worldmemarena,microsoft2026statebench}.
These works preclude a generic priority claim for multi-session, execution-oriented,
or lifecycle-aware memory evaluation.  The narrower \bench{} contribution is the
intersection of controlled repository continuation and memory hygiene: hidden,
non-inferable earlier-session evidence must cause real code edits that pass
executable repository oracles, under CSPRNG token injection, scored-artifact
isolation, and a C1--C10 taxonomy of SWE-specific pathologies.

\paragraph{Agent memory architecture.}
CoALA gives useful vocabulary for episodic, semantic, and procedural memories
\citep{sumers2024coala}; the reference probe uses that vocabulary as schema discipline rather
than as a claim of cognitive novelty.  MemGPT owns the virtual-context framing:
an LLM agent manages memory tiers to extend useful context beyond the immediate
window \citep{packer2024memgptllmsoperatingsystems}.  Letta's memory-model framing
pushes this further toward trained token-space memory curators and sleep-time memory
generation for future tasks \citep{letta2026memorymodels}.  Sleep-time Compute owns
the general idea of spending offline compute on context before later queries arrive,
including a case study in an agentic SWE setting \citep{lin2025sleeptimecomputeinferencescaling}.
The reference probe adds a narrower, training-free mechanism: typed provenance over SWE
trajectories, whole-store contradiction repair, grounded counterfactual replay, stale
suppression, and a retrieval gate whose output is judged by executable hidden oracles
rather than by memory plausibility alone.

\paragraph{External memory systems and agent-native memory.}
Mem0 is the single external memory-system family represented in the original v2
diagnostics, through two pinned hosted configurations.
It owns production-oriented online memory extraction, consolidation, retrieval, and a
graph variant for relational memory and obsolete-triple handling
\citep{chhikara2025mem0}.  Our B5 condition is only a deterministic verbatim
event-memory substitute.  B5-MEM0 uses hosted-default extraction, whereas
B5-MEM0-LIT disables fact inference as an exact-literal diagnostic; neither is a
tuned-Mem0 ranking claim.  The distinction is that the reference probe performs offline
typed maintenance over a preserved SWE evidence store, while \bench{} specifically
tests whether a later coding session preserves exact hidden CSPRNG tokens and avoids
stale or contradictory memory use under executable oracles.
The separately preregistered v2.1 audit applies a stricter conformance boundary: only
B5-MEM0-LIT was admitted and evaluated; native B5-MEM0 and both planned Supermemory
conditions were rejected before evaluation.  The successor therefore profiles one exact
external configuration and reports conformance attrition rather than treating unavailable
systems as observed losses.  The rejected Supermemory lanes were pinned to the official
repository's \texttt{server-v0.0.5} release and release-tag commit
\citep{supermemoryrepo2026}; that citation establishes conformance provenance only, not
evaluated Supermemory performance.

Agent-Native Memory (ANM) studies agent memory as a data-management system, decomposing
memory into representation and storage, extraction, retrieval and routing, and
maintenance, and evaluating cost and robustness tradeoffs across representative systems
\citep{zhou2026agentnativememory}.  That work owns the broad systems characterization.
\bench{} adds a task-family-specific benchmark in which the ground truth is not an
open-ended dialogue answer but a hidden, non-inferable code contract, and the reference probe adds
a concrete training-free maintenance pipeline for typed SWE memories.

\paragraph{SWE-agent memory and task-state substrates.}
Shen et al.'s structurally aligned subtask-level memory is the closest published
SWE-agent memory comparison point we cite.  It owns subtask-granularity storage,
retrieval, and updating aligned with a software agent's functional decomposition, and
reports a mean Pass@1 gain of $+4.7$ percentage points over vanilla agents on
SWE-bench Verified across four backbones \citep{shen2026structurally}.  Our B6
condition is a subtask-memory substitute inspired by that granularity, not a fidelity
reimplementation and not a claim about Shen et al.'s system.  \bench{} asks a
different question: whether memory systems survive multi-session hygiene traps with
hidden exact-token contracts, stale facts, contradictory feedback, and executable
repository oracles.

beads is adjacent persistent SWE memory infrastructure rather than a direct memory
hygiene benchmark.  It owns a Dolt-backed structured issue and project-memory graph
for coding agents, with dependency links, \texttt{supersedes} links, and compaction of
closed tasks \citep{yegge_beads}.  The reference probe adds semantic maintenance over agent traces:
typed provenance, contradiction repair, grounded replay, stale suppression, and
retrieval gating.  A beads-style task graph could be a useful substrate or tracker,
but the claim tested here is not that durable task storage exists; it is whether the
maintenance operators improve future SWE behavior under \bench{}.

\paragraph{Reflection, replay, and procedural lessons.}
Generative Agents and Reflexion convert experience into later-use memories or verbal
lessons; in this paper, Reflexion-style lesson reuse is represented by B3
reflection-only vector retrieval \citep{park2023generative,shinn2023reflexion}.
Agent Workflow Memory, ExpeL, Voyager, and self-improving coding-agent work study
procedural lesson extraction, executable skill accumulation, or editing the agent
implementation itself \citep{wang2024agentworkflowmemory,zhao2024expel,wang2023voyager,robeyns2025selfimproving}.
The reference probe does not claim novelty for trajectory-to-lesson conversion.  The added claim is
that lessons are typed, provenance-linked, lifecycle-managed, and checked against
future executable SWE traps rather than treated as free-form reusable advice.

\paragraph{Memory safety, evaluation, and degradation.}
Faulty Memory shows that continuously updated useful memories can degrade, making raw
evidence and consolidation controls important \citep{zhang2026faulty}.  MemGate treats
memory search as a trust boundary, TRUSTMEM studies learned write/revise/prune transitions,
and classical JTMS/ATMS/AGM belief revision gives the formal lineage for
justification-aware inconsistency management
\citep{zhang2026beyondsimilarity,yang2026trustmem,doyle1979truthmaintenance,dekleer1986assumptionbased,alchourron1985logic}.
BinEval motivates decomposing judge questions into inspectable binary decisions
\citep{cho2026askdontjudge}.  In this paper, those hygiene labels remain diagnostic:
the primary judge-free outcome is executable S3 pass@1, and the benchmark logs stale,
contradictory, \texttt{RepeatedErrorRate}, harmful-retrieval, and scope diagnostics
relative to raw episodic controls.

\paragraph{Research and knowledge-curation agents.}
STORM and Co-STORM use retrieval, multi-perspective question asking, agent conversation,
and dynamic knowledge structures to synthesize Wikipedia-like articles and support
learning \citep{shao-etal-2024-assisting,jiang-etal-2024-unknown}.  Tongyi
DeepResearch, DeepSearcher, Local Deep Research, Chain-of-Retrieval Augmented
Generation, MiroThinker, and MiroFlow are adjacent research or retrieval-and-synthesis
frameworks that organize evidence through iterative search, local-corpus retrieval,
verification, or report generation
\citep{tongyi2025deepresearch,deepsearcher2025,learningcircuit_localdeepresearch,wang2025corag,miromind2025mirothinker,miromind2026mirothinker17h1,su2026miroflow}.
Those systems broaden or curate external knowledge for a research question; \bench{}
instead evaluates maintenance over a coding agent's own prior software-work history.

Table~\ref{tab:related-comparison} summarizes the closest memory work by claim boundary.
The only external memory-system family with a scored condition is Mem0: the original
v2 diagnostics contain B5-MEM0 and B5-MEM0-LIT, while the stricter successor audit
admits only B5-MEM0-LIT.  The other rows are positioning references rather than
practical baselines.

{\scriptsize
\setlength\LTleft{0pt}
\setlength\LTright{0pt}
\begin{longtable}{p{0.20\linewidth}p{0.33\linewidth}p{0.39\linewidth}}
\caption{Closest prior work and claim boundary.  The table does not assert generic
benchmark priority.  Rows other than Mem0 are positioning references, not additional
external memory-system baselines in the reported v2 fold.}
\label{tab:related-comparison}\\
\toprule
\textbf{Work} & \textbf{What prior work owns} & \textbf{What \bench{}/reference probe adds} \\
\midrule
\endfirsthead
\toprule
\textbf{Work} & \textbf{What prior work owns} & \textbf{What \bench{}/reference probe adds} \\
\midrule
\endhead
\bottomrule
\endlastfoot
LoCoMo, LongMemEval, BEAM, AMA-Bench, MemoryAgentBench \citep{maharana2024locomo,wu2025longmemeval,tavakoli2025beam,zhao2026amabench,hu2025memoryagentbench} & Long-history conversational memory, temporal reasoning, updates, incremental interaction, selective forgetting, and agentic QA. & Repository-state continuation in which earlier-session evidence controls executable code-oracle success rather than only answer quality. \\
MemoryArena \citep{he2026memoryarena} & Interdependent multi-session Memory-Agent-Environment loops in web navigation, planning, search, and formal reasoning. & SWE-specific hidden-evidence traps, production code edits, repository oracles, and contamination controls. \\
EvoMemBench \citep{wang2026evomembench} & In-/cross-episode and knowledge-/execution-oriented evaluation of 15 representative memory methods. & A narrower controlled test of repository continuity and hygiene pathologies, with trap-clustered inference over executable S3 outcomes. \\
WorldMemArena \citep{liu2026worldmemarena} & Multimodal action-world interaction with stage-level diagnosis of memory writing, maintenance, retrieval, and use. & Text-and-code repository tasks whose decisive evidence is hidden from S3 and whose success is checked by software oracles. \\
STATE-Bench \citep{microsoft2026statebench} & Stateful enterprise workflows with database actions, simulated users, and an Agent Learning track for reusable memories, skills, or prompt optimizations. & Multi-session software-repository memory hygiene with frozen earlier evidence, code diffs, and hidden repository assertions. \\
Mem0 / Mem0$^g$ \citep{chhikara2025mem0} & Online extraction, consolidation, retrieval, and graph memory for long-term agent memory; B5-MEM0 and B5-MEM0-LIT are two pinned configurations of one hosted external system family. & Multi-session SWE hygiene traps with hidden CSPRNG tokens and executable oracles; offline typed provenance, contradiction repair, grounded replay, stale suppression, and a retrieval gate. \\
MemGPT, Letta, Sleep-time Compute \citep{packer2024memgptllmsoperatingsystems,letta2026memorymodels,lin2025sleeptimecomputeinferencescaling} & Virtual context, memory blocks or memory models, and offline compute over context before later inference. & A training-free SWE maintenance pipeline whose replay is grounded in prior trajectory evidence and evaluated by hidden repository oracles. \\
ANM \citep{zhou2026agentnativememory} & Data-management view of agent memory modules, workload-dependent tradeoffs, cost, robustness, and maintenance analysis. & A SWE-specific construct taxonomy, CSPRNG-token traps, and executable multi-session oracles for memory-hygiene failures. \\
Shen et al. \citep{shen2026structurally} & Subtask-level SWE-agent memory aligned with functional decomposition; reported $+4.7$pp mean Pass@1 over vanilla agents on SWE-bench Verified. & The benchmark target differs by requiring exact continuation of concealed identifiers, stale/contradictory memory, generated-file and scope traps, and B6 as a non-fidelity substitute only. \\
beads \citep{yegge_beads} & Dolt-backed task and memory graph for coding agents, with dependency, \texttt{supersedes}, and compaction mechanisms. & Semantic maintenance over trajectory evidence: typed derived memories, contradiction repair, grounded counterfactual replay, stale suppression, and retrieval gating. \\
Reference probe & Scheduled offline maintenance over raw SWE evidence. & The proposed mechanism and the object evaluated by the three-seed confirmatory fold. Raw episodes are preserved even when derived memories are suppressed or logically deleted. \\
\end{longtable}
}

%% file: sections/04_problem_formulation.tex
\section{Problem Formulation}

Let $A$ be a software-engineering agent, $R$ a repository, $T$ a task, and $S_i$ the $i$th session in a multi-session sequence. A session is a bounded agent run with a task prompt, repository state, tool access, a model configuration, a memory policy, and a termination condition. A sequence is an ordered list of sessions over one repository or a controlled family of related repositories.

\paragraph{Trajectory.}
An agent trajectory is an append-only record
\[
  \tau = (\mathrm{meta}, e_1, e_2, \ldots, e_n, \mathrm{outcome}),
\]
where $\mathrm{meta}$ contains identifiers such as task id, session id, repository id, starting commit, model id, condition id, seed, and budgets. Each event $e_i$ is an observed state, model message, tool call, tool result, file edit, test result, memory read, memory write, reviewer message, or termination event. The outcome records the final diff, tests run, pass or fail status when executable verification exists, cost, latency, and labels. A raw episode is a persisted trajectory or trajectory segment with immutable evidence semantics.

\paragraph{Memory item.}
A derived memory item $m$ contains content, type, provenance, write reason, id, status, repository scope, file scope, symbol scope, task scope, timestamps, validity interval, confidence, utility, risk, staleness, contradiction links, supersession links, retrieval tags, usage accounting, and optional outcome impact. The memory types are \texttt{episodic}, \texttt{semantic\_project}, \texttt{procedural}, \texttt{failure}, \texttt{human\_feedback}, \texttt{constraint}, \texttt{contradiction}, and \texttt{dream\_artifact}. The statuses are \texttt{active}, \texttt{superseded}, \texttt{stale}, \texttt{requires\_review}, and \texttt{deleted}. The \texttt{deleted} status is a logical soft-delete for derived memories: it blocks retrieval while preserving the memory record and provenance. It is not used to physically remove derived records or raw trajectory evidence in reference-probe experiments.

\paragraph{Read admissibility.}
For task context $q$ and candidate memory $m$, the reference probe separates candidate generation from admission. Candidate generation can use semantic similarity, tags, file-path overlap, symbol overlap, task labels, and project scope. Admission rejects $m$ when its status, validity interval, scope, supersession state, risk, staleness, provenance, or type policy makes it inadmissible for $q$. Among admissible memories, the system ranks under a fixed memory-token budget.

\paragraph{Memory-hygiene objective.}
Let $R_i$ be the set of memory items admitted into session $i$, $U_i \subseteq R_i$ the items judged useful, $H_i \subseteq R_i$ the items judged harmful, and $S_i \subseteq R_i$ the items that should have been suppressed because they are stale, superseded, out of scope, or review-required. Memory hygiene measures whether the memory system retrieves useful, grounded, current, scoped, low-risk memories while suppressing harmful or stale memories. This differs from generic recall: retrieving more memories can reduce hygiene when the extra memories are wrong for the current repository state.

\paragraph{Metrics.}
TaskSuccess is the fraction of sessions whose final patch satisfies the task oracle. Pass@1 is the fraction of executable sessions whose first completed attempt passes the configured verification suite. UsefulMemoryPrecision is
\[
  \frac{\sum_i |U_i|}{\sum_i |R_i|}.
\]
HarmfulMemoryRate is
\[
  \frac{\sum_i |H_i|}{\sum_i |R_i|}.
\]
In the reported confirmatory fold, HarmfulMemoryRate is logged but non-discriminating:
it is 0.000 for every condition, so the fold does not support harmful-retrieval reduction
claims.  RepeatedErrorRate is the count of repeated error events divided by the count of
opportunities to avoid repetition. Stale or superseded activation is tracked
diagnostically when stale memory is available, but it is not a standalone
confirmatory-fold metric in the reported hygiene panel. ContradictionRepairAccuracy is the
fraction of evaluable contradictions whose expected old-item status, new-item status, and
relation are all correct. Cost metrics separate wake tokens, sleep tokens, judge tokens,
wake latency, sleep latency, and judge latency.

\paragraph{Research question.}
The central question is RQ0: for multi-session software engineering agents, what do
offline maintenance, raw-evidence retention, and their combination change under fixed
model, token budget, tool budget, and task stream?  The confirmatory fold isolates the
typed pipeline, raw-evidence capsule, and hybrid ladder.  The component ablations
(A0, A2, A4, A5, A6, and A11) are descriptive in v2: A0 episodic-only is
$20/180=0.111$ versus typed-only at $80/180=0.444$; A6 no retrieval hard gate is
$72/180=0.400$; A4 no counterfactual replay is $77/180=0.428$; A5 no stale
suppression is $79/180=0.439$; A2 no contradiction repair is $88/180=0.489$;
and A11 forced consolidation is $70/180=0.389$.  Because A2 scores above
typed-only and \texttt{CONTRADICTION} records appear in retrieval for 0/180
hybrid and 0/180 raw-only S3 records, contradiction-repair and
operator-specific causal claims remain hypotheses unless supported by paired
component tests or audits.

%% file: sections/05_method.tex
\section{Method: Reference Probe}
\label{sec:method}

The reference probe wraps a fixed wake-phase software-engineering agent with an external memory store and an offline maintenance loop. The wake agent can be a SWE-agent-style harness or another code-capable tool-using agent. The probe does not train a new base model and does not require the wake agent to expose internal weights or hidden state.

\subsection{Wake/Sleep/Read Loop}

For each benchmark sequence, the reference probe initializes a repository snapshot and a condition-specific memory namespace. At session $S_i$, the read phase retrieves candidate memories and applies the retrieval gate. The wake phase gives the admitted memories, task prompt, repository state, and tool budget to the agent. The trajectory logger records task metadata, model configuration, tool calls, command output, diffs, memory reads, memory writes, feedback, final outcome, cost, and latency. After the session, the sleep schedule determines whether offline maintenance runs over raw episodes since the previous pass. Figure~\ref{fig:arch} summarizes the wake/sleep/read loop.

\paragraph{Raw-episode invariant.}
The system maintains two stores: an append-only raw episode log $E$ and a derived memory store $M$. A raw episode is the evidence record for one session or replay: prompt, repository state, tool transcript, command output, diffs, memory reads, memory writes, feedback, outcome, and cost. No reference-probe operator overwrites, truncates, summarizes in place, hard-deletes, or garbage-collects raw episodes during an experiment. Sleep operators can only append derived memories, append audit records, or change the lifecycle status of derived memories. The status \texttt{deleted} is therefore a logical soft-delete that applies only to derived memories; the derived record, its provenance, and all raw episodes remain addressable for audit, replay, and episodic-only controls.

\paragraph{Frozen raw-evidence repairs and system variants.}
The confirmatory fold uses three named repairs from
\path{analysis/investigation-evidence/PREREGISTRATION.md}, frozen on 2026-07-03 before
the confirmatory fold (SHA-256 prefix \texttt{e1819a8c930a}; full digest in
Appendix~\ref{app:artifact}).  R1 adds a verbatim raw-evidence capsule: one
\texttt{EPISODIC} memory per episode carrying the injected event content through the
normal retrieval gate.  R1b exempts
\texttt{EPISODIC} records from contradiction repair so raw evidence is not rewritten or
deleted by repair.  R2 excludes \texttt{CONTRADICTION}-type records from implementation
reads through the existing \texttt{allowed\_types} gate.

These repairs define the ladder used in the fold.  Reference-probe typed-only uses the derived
typed pipeline without the raw-evidence capsule.  Reference-probe raw-only disables typed
consolidation and exposes only the R1/R1b raw-evidence capsule, with R2 still excluding
contradiction records.  Reference-probe hybrid combines the typed pipeline and the raw-evidence
capsule.  By construction, reference-probe raw-only is near-equivalent to B5 on the dominant
recall-verbatim portion of \bench{}: both give the wake agent one verbatim injected-event
record, although B5 is the simpler deterministic event-memory substitute and raw-only
uses the same probe read path.  Here recall-verbatim identifies where the contract evidence
appears; it does not mean that the S3 patch can be copied or that implementation is unnecessary.
The historical v1 fold reflects this construction: 65/66 raw-only-vs.-B5
paired S3 outcomes agree, with one B5-only win.  This near-equivalence is descriptive,
not an equivalence test.

\input{figures/architecture}

\subsection{Typed Memory Schema}

The derived store uses a typed memory item with explicit provenance and lifecycle metadata. Required construction fields are content, type, provenance, and write reason. Provenance must reference at least one raw episode or raw artifact: trajectory id, task id, repository commit, file path observation, command output, diff, or human-feedback event. Scope fields constrain a memory to a repository, files, symbols, tasks, sequence type, or validity interval. Status and validity fields allow repair and stale suppression to block retrieval without erasing evidence. Confidence, utility, risk, and staleness scores are unit-bounded metadata used by read and maintenance policies. A derived memory without sufficient provenance is marked \texttt{requires\_review} and is ineligible when the provenance gate is enabled.

\begin{sloppypar}
The schema supports eight memory types: episodic, semantic-project,
procedural, failure, human-feedback, constraint, contradiction, and
dream-artifact memory. These types are not claimed to be universally optimal.
H7 tests whether the typed, provenance-linked schema improves
useful-memory precision relative to untyped summaries under an equal retrieval
budget; harmful-memory rate is logged as a diagnostic but is non-discriminating
in the reported fold because it is 0.000 for every condition.
\end{sloppypar}

\subsection{Sleep-Phase Operators}

Each sleep-time operation has a named intended ablation. This is deliberate: the paper's empirical spine is not that ``sleep'' helps, but which maintenance operation, if any, changes future software-agent behavior. The confirmatory fold, however, isolates only the ladder-level typed pipeline, raw-evidence capsule, and hybrid. Component ablations that were run are descriptive unless separately paired and tested. Table~\ref{tab:operator-ablation-map} therefore separates implemented operators from ablations isolated by the fold.

\begin{table}[!htbp]
\centering
\small
\resizebox{\textwidth}{!}{%
\begin{tabular}{p{0.20\linewidth}p{0.38\linewidth}p{0.18\linewidth}p{0.17\linewidth}}
\toprule
\textbf{Operation} & \textbf{Write or transition} & \textbf{Ablation} & \textbf{Fold status} \\
\midrule
Typed consolidation & Candidate semantic-project, procedural, human-feedback, constraint, and episodic-derived memories with scope, provenance, confidence, utility, risk, staleness, tags, and write reason & A1 no typed consolidation; A11 forced consolidation & A1 designed/not run; A11 run/descriptive; ladder isolates typed pipeline \\
Causal failure extraction & Failure memories grounded in failed trajectories, command output, tests, diffs, logs, or feedback & A3 no causal failure extraction & Designed/not run \\
Counterfactual replay & Dream artifacts and replay-derived failure or suppression decisions, accepted only when grounded in executable evidence, raw trace evidence, or human-audited causal support & A4 no counterfactual replay & Run/descriptive \\
Contradiction repair & Contradiction records, supersession links, status transitions, and validity-interval updates over derived memories & A2 no contradiction repair & Run/descriptive \\
Stale suppression and forgetting & \texttt{stale}, \texttt{superseded}, or logical \texttt{deleted} status for derived memories; retrieval exclusion without deleting raw evidence & A5 no stale suppression & Run/descriptive; stale activation evaluated through proxies \\
Provenance gate & \texttt{active} versus \texttt{requires\_review} transition based on whether the derived claim is supported by raw episodes & A8 no provenance gate & Designed/not run \\
Raw episodic fallback & Agent-visible raw-evidence capsule while preserving raw episodes for audit & A10 disables raw fallback & Designed/not run; ladder isolates raw capsule via DF-raw-only and DF-hybrid \\
Metadata retrieval gate & Admission by status, scope, provenance, confidence, staleness, risk, and supersession metadata & A6 no retrieval hard gate; A7 similarity-only; A12 no scope; A13 no validity intervals & A6 run/descriptive; A7/A12/A13 designed/not run \\
\bottomrule
\end{tabular}
}%
\caption{Implemented operators and ablation coverage. A0 episodic-only disables the derived store entirely and is run as a descriptive floor. The confirmatory fold isolates ladder-level channels; it does not statistically isolate every named operator.}
\label{tab:operator-ablation-map}
\end{table}

The sleep pipeline is:
\[
\begin{aligned}
\mathrm{typed} &= C_{\mathrm{type}}(E),\\
\mathrm{failures} &= C_{\mathrm{fail}}(E),\\
\mathrm{replay} &= C_{\mathrm{replay}}(E, M),\\
\mathrm{proposed} &= \mathrm{typed} \cup \mathrm{failures} \cup \mathrm{replay},\\
\mathrm{repaired} &= C_{\mathrm{repair}}(\mathrm{proposed}, M),\\
\mathrm{suppressed} &= C_{\mathrm{stale}}(\mathrm{repaired}, M),\\
\mathrm{validated} &= C_{\mathrm{prov}}(\mathrm{suppressed}, E).
\end{aligned}
\]
Here $E$ is the set of raw episodes selected by the sleep schedule and $M$ is the current derived store. The output is a set of derived writes and status transitions. The functions above are pure with respect to raw evidence: they may read $E$ and write provenance pointers into $M$, but they may not mutate $E$.

\subsection{Retrieval Gate}

The read path first retrieves broad candidates and then applies hard gates. A memory item is rejected if it is not \texttt{active}, outside the current repository scope, outside the task's file or symbol scope, beyond its validity interval, superseded by an active item, above a risk or staleness threshold, ungrounded when provenance gating is enabled, logically deleted, review-required, or excluded by the condition-specific type policy.

Let $G(q,m)\in\{0,1\}$ be the hard-gate decision for task context $q$ and memory $m$. If $G(q,m)=0$, $m$ is assigned score $-\infty$. Otherwise admitted candidates are ranked by
\[
\begin{aligned}
\mathrm{score}(q,m) ={}&
w_{\mathrm{sem}}\cos(e_q,e_m)
+ w_{\mathrm{tag}}J(T_q,T_m)
+ w_{\mathrm{file}}F(q,m)
+ w_{\mathrm{sym}}Y(q,m)\\
&+ w_{\mathrm{task}}K(q,m)
+ w_{\mathrm{type}}\pi_{\mathrm{type}}(m)
+ w_{\mathrm{conf}}c_m
+ w_{\mathrm{util}}u_m
+ w_{\mathrm{verify}}v_m
+ w_{\mathrm{prov}}p_m\\
&- \lambda_{\mathrm{stale}}s_m
- \lambda_{\mathrm{risk}}r_m
- \lambda_{\mathrm{age}}\log(1+\Delta t_m)
- \lambda_{\mathrm{overscope}}o(q,m).
\end{aligned}
\]
Here $J$ is tag-overlap similarity; $F$, $Y$, and $K$ are file-, symbol-, and task-scope compatibility scores; $\pi_{\mathrm{type}}$ is the condition-specific type prior; $c_m$, $u_m$, $v_m$, and $p_m$ are confidence, utility, human-verification, and provenance-strength metadata; $s_m$, $r_m$, $\Delta t_m$, and $o(q,m)$ are staleness, risk, age, and overscope penalties. The weights, thresholds, candidate pool size, and memory-token budget are fixed before held-out evaluation. H9 was designed to test whether this metadata-aware gate reduces risky or harmful retrievals relative to similarity-only retrieval, but the direct similarity-only comparator A7 was not run and HarmfulMemoryRate floors at 0.000 for every condition. A6 removes the hard gate and is reported descriptively; A7 keeps only semantic similarity, A12 ignores scope fields, and A13 ignores validity intervals but remain designed-not-run ablations in this fold.

\subsection{Condition Isolation}

Each model, seed, condition, run id, and repository receives a separate memory namespace. No condition can read another condition's memories. This isolation is necessary because the benchmark measures memory updates over time; cross-condition contamination would invalidate comparisons between the reference probe, baselines, and ablations.

The reported evaluation pairs this logical namespace isolation with the filesystem
isolation guarantee: scored benchmark artifacts are absent from the wake-agent and
sleep-judge container filesystems, but the run is not network-isolated. Wake agents and
sleep judges run in per-call Linux containers with only the task worktree and Codex
authentication material mounted as needed; benchmark oracles, reference solutions,
sequence records, analysis files, logs, and the paper repository are not mounted into the
agent environment. Section~\ref{sec:isolation-boundary} gives the experimental protocol
and the adversarial audit that motivated the containerized rerun.

\paragraph{Implementation repairs.}
The R1/R1b/R2 repairs are described above because they define the evaluated ladder
variants, not merely the isolation protocol.  Section~\ref{sec:results-capsule} verifies
that raw-evidence capsules and contradiction-record exclusion were active in the fold.

%% file: figures/architecture.tex
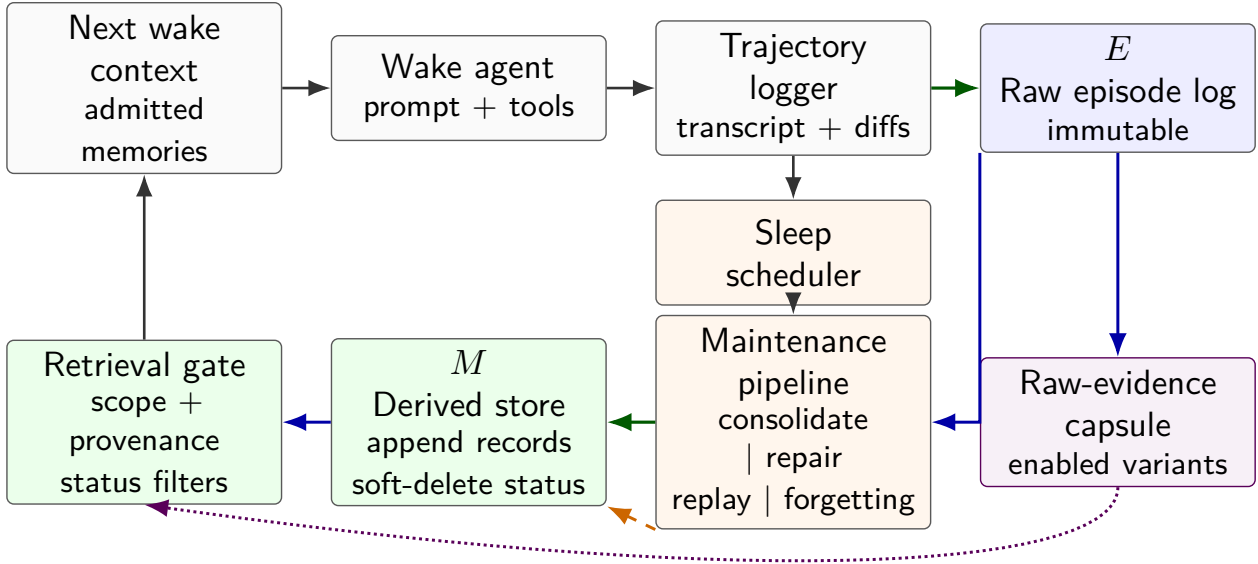
\begin{figure}[t]
\centering
\small
\resizebox{\linewidth}{!}{%
\begin{tikzpicture}[
  font=\sffamily,
  box/.style={
    draw=black!65,
    fill=black!2,
    rounded corners=2pt,
    align=center,
    minimum width=2.75cm,
    minimum height=1.05cm,
    text width=2.45cm
  },
  raw/.style={box, fill=blue!7},
  mem/.style={box, fill=green!8},
  sleep/.style={box, fill=orange!7},
  capsulebox/.style={box, draw=violet!70!black, fill=violet!6},
  flow/.style={-Latex, line width=0.75pt, draw=black!80},
  readflow/.style={-Latex, line width=0.85pt, draw=blue!65!black},
  writeflow/.style={-Latex, line width=0.9pt, draw=green!35!black},
  statusflow/.style={-Latex, line width=0.85pt, draw=orange!80!black, dashed},
  capsuleflow/.style={-Latex, line width=0.85pt, draw=violet!70!black, densely dotted}
]
\node[box] (context) at (0,2.8) {Next wake\\context\\[-0.1em]\footnotesize admitted memories};
\node[box] (wake) at (3.25,2.8) {Wake agent\\[-0.1em]\footnotesize prompt + tools};
\node[box] (logger) at (6.50,2.8) {Trajectory\\logger\\[-0.1em]\footnotesize transcript + diffs};
\node[raw] (episodes) at (9.75,2.8) {\textbf{$E$}\\Raw episode log\\[-0.1em]\footnotesize immutable};

\node[mem] (gate) at (0,-0.55) {Retrieval gate\\[-0.1em]\footnotesize scope + provenance\\status filters};
\node[mem] (memory) at (3.25,-0.55) {\textbf{$M$}\\Derived store\\[-0.1em]\footnotesize append records\\soft-delete status};
\node[sleep] (pipeline) at (6.50,-0.55) {Maintenance\\pipeline\\[-0.1em]\footnotesize consolidate | repair\\replay | forgetting};
\node[capsulebox] (capsule) at (9.75,-0.55) {Raw-evidence\\capsule\\[-0.1em]\footnotesize enabled variants};
\node[sleep] (scheduler) at (6.50,1.15) {Sleep\\scheduler};

\draw[flow] (gate.north) -- (context.south);
\draw[flow] (context) -- (wake);
\draw[flow] (wake) -- (logger);
\draw[writeflow] (logger) -- (episodes);
\draw[flow] (logger.south) -- (scheduler.north);
\draw[flow] (scheduler.south) -- (pipeline.north);

\draw[readflow] (episodes.south) -- (capsule.north);
\draw[readflow] (episodes.south west) |- (pipeline.east);
\draw[writeflow] (pipeline.west) -- (memory.east);
\draw[statusflow] (pipeline.south west) -- (memory.south east);
\draw[readflow] (memory.west) -- (gate.east);
\draw[capsuleflow] (capsule.south) .. controls +(0,-0.85) and +(6.0,-0.85) ..
  (gate.south);
\end{tikzpicture}%
}
\caption{Reference-probe system architecture for the wake/sleep/read loop. Raw episodes are append-only; sleep operators write or update only derived-memory records.}
\label{fig:arch}
\end{figure}

%% file: sections/06_experiments.tex
\section{Experimental Design}

The experiment tests whether offline memory maintenance improves multi-session software engineering reliability and memory hygiene under controlled budgets. It does not test whether metaphorical dreaming is useful. All empirical performance statements below refer to the pre-registered three-seed confirmatory fold reported in Section~\ref{sec:results}.

\subsection{\bench{}}
\label{sec:benchmark}

\bench{} is a dynamic multi-session benchmark. A benchmark instance contains a repository
id, initial commit, sequence type, sessions, injected events, oracle labels, verification
commands, and budgets. The historical v1 fold used 22 three-session sequences. The scaled
v2 confirmatory fold reported as the primary paper result uses 60 admitted three-session
traps, each with S1 and S2 setup or reinforcement sessions followed by an S3 trap session
that tests whether a later agent run uses prior evidence without reading hidden oracles or
reference solutions.

The historical 22-sequence v1 set is unevenly distributed across families (reviewer-preference 7,
convention-learning 5, stale-architecture 4, generated-files 3, flaky-test 3), so
per-family results are descriptive at small $n$ and reviewer-preference traps carry
disproportionate weight. We therefore describe \bench{} as a curated diagnostic stress
benchmark for SWE memory hygiene, not as broad coverage of production issue-resolution.
S3 traps turn on arbitrary tokens that are non-inferable from the public prompt and
recoverable only from earlier-session memory; this makes them genuine memory traps because
the scored material is hidden outside the container filesystem
(Section~\ref{sec:isolation-boundary}).

Continuation is lenient by design. S3 materializes from prior sequence state, and a
session advances when its patch applies with a non-empty production diff and no policy
rejection, even if the patch failed its oracle. S3 therefore models an ongoing project
with imperfect prior edits rather than a clean success chain. Figure~\ref{fig:templates}
sketches the sequence-template skeleton used for the reported fold.

The v1.0 trap set is mostly but not entirely a recall-verbatim benchmark.  The audit in
Appendix~\ref{app:verbatim} classifies 18 of 22 traps as recall-verbatim: the oracle-required
token, format, or rule is literally present in prior injected event text that B5 stores.
The remaining four are synthesis/apply traps involving generated-file or workflow
application.  This is an evidence-location taxonomy, not a patch-copyability taxonomy:
every S3 trap still requires repository-specific implementation, and a stricter semantic
audit found no pure copy-and-stop trap.  The taxonomy is central to interpreting B5: the
baseline is strong because most traps reward exact preservation of one prior event, not
because it models all forms of long-horizon SWE memory.  Section~\ref{sec:results-taxonomy}
reports the descriptive outcome split by this taxonomy.

Table~\ref{tab:unit-aggregation} fixes the denominators used below.

\begin{table}[!htbp]
\centering
\small
\begin{tabular}{p{0.28\linewidth}p{0.64\linewidth}}
\toprule
\textbf{Quantity} & \textbf{Definition in the confirmatory fold} \\
\midrule
Trap universe & V2 primary fold: 60 S3 traps, each the third session of a three-session sequence; historical v1 fold: 22 S3 traps. \\
Seeds & 3 seeds per condition. \\
Complete S3 cells & 180 seed-trap S3 task cells for complete v2 conditions; 66 seed-trap S3 task cells for complete historical v1 conditions. \\
Warmup denominator & Diagnostic pre-S3 trajectory counts reported by the analyzer; warmup rates do not replace the S3 task-cell denominator. \\
Hygiene denominator & Unique analyzer-contributing result records for the relevant fold; hygiene means and count sums are deduped by the canonical analyzer before paper tables are generated. \\
Validity exclusions & Record-level exclusions when \texttt{error\_type} contains \texttt{task\_exception}, \texttt{codex\_exec\_failed}, or \texttt{isolation\_unavailable}; exclusions are not automatically lost S3 cells. \\
Deduplication & Newest record per \texttt{(condition, seed, oracle\_id)} by result timestamp. \\
Known missingness & The v2 complete conditions in Section~\ref{sec:results} contribute 180 S3 cells with zero excluded records. Historical v1 missingness is reported only with the v1 rows; for example, typed-only had one \texttt{codex\_exec\_failed} exclusion and DF-strict was incomplete. \\
\bottomrule
\end{tabular}
\caption{Unit of analysis and aggregation rules. Source: \texttt{CONFIRMATORY-FOLD.md}.}
\label{tab:unit-aggregation}
\end{table}

The five sequence types are:

\begin{itemize}
  \item \textbf{Convention-learning:} later sessions require project-local coding, testing, naming, formatting, error-handling, or API conventions learned earlier.
  \item \textbf{Generated-files:} later sessions test whether the agent distinguishes source-of-truth files from generated artifacts after regeneration.
  \item \textbf{Stale-architecture:} later sessions make an older architecture fact false or scoped only to historical files.
  \item \textbf{Reviewer-preference:} later sessions test scoped use of human feedback and avoidance of overgeneralization after feedback changes.
  \item \textbf{Flaky-test:} later sessions test whether the agent avoids encoding an unsupported causal lesson from nondeterministic or environment-sensitive failures.
\end{itemize}

\input{figures/sequence_templates}

\subsection{Conditions}

Table~\ref{tab:condition-contract} is the condition contract used throughout the paper.
B3's canonical label is \emph{reflection-only vector retrieval}.  DF is the typed-only
reference-probe condition; the reported hybrid variant is named explicitly as reference-probe hybrid,
not with a generic full label.  The R1/R1b/R2 repairs are introduced in
Section~\ref{sec:method} and were frozen in \texttt{PREREGISTRATION.md} on
2026-07-03 before the confirmatory fold.

\begin{table}[!htbp]
\centering
\small
\resizebox{\textwidth}{!}{%
\begin{tabular}{p{0.13\linewidth}p{0.26\linewidth}p{0.55\linewidth}}
\toprule
\textbf{Code} & \textbf{Canonical label} & \textbf{Contract} \\
\midrule
B0 & No external memory & No external memory reads or writes. \\
B1 & Raw episodic retrieval & Retrieves raw prior trajectories. \\
B2 & Vector trace retrieval & Deterministic token-vector retrieval over prior trace chunks. \\
B3 & Reflection-only vector retrieval & Free-form verbal reflection lessons after sessions, retrieved by vector similarity. \\
B4 & Untyped summary memory & Deterministic untyped session summaries. \\
B5 & Verbatim event-memory substitute & Offline instance-memory substitute; one deterministic verbatim injected-event memory per trajectory; not stock Mem0. \\
B5-MEM0 & Pinned hosted Mem0 & Hosted Mem0 baseline with \texttt{mem0ai} 2.0.11, hosted platform, default extraction settings, \texttt{memory\_context=6}. \\
B5-MEM0-LIT & Pinned hosted Mem0 literal diagnostic & Same hosted Mem0 system and retrieval settings as B5-MEM0, but with fact inference disabled so sanitized raw event text is stored directly. \\
B6 & Subtask memory & Subtask-level memory over trajectory actions; inspired by but not a fidelity implementation of Shen et al. \\
B7 & Task-tracker memory & Sequence-local task-tracker memory. \\
DF & reference-probe typed-only & Typed consolidation, contradiction repair, counterfactual replay, local maintenance, and retrieval gating; no raw-evidence capsule. \\
DF-raw-only & reference-probe raw-only & Typed consolidation disabled; raw-evidence capsules enabled; contradiction records excluded from implementation reads. \\
DF-hybrid & reference-probe hybrid & DF plus raw-evidence capsules; contradiction records excluded from implementation reads. \\
DF-strict & reference-probe strict & DF with \(\theta_{\mathrm{admit}}=1.5\), read limit 3, and read-token budget 600. \\
DF-strict-hybrid & reference-probe strict-hybrid & DF-hybrid with the same strict retrieval settings as DF-strict. \\
A0 & Episodic-only & Raw episode writes only; no typed consolidation, repair, or replay. \\
A2 & No contradiction repair & Contradiction repair disabled. \\
A4 & No counterfactual replay & Counterfactual replay disabled. \\
A5 & No stale suppression & Read-side stale and superseded suppression disabled while repair remains enabled. \\
A6 & No retrieval hard gate & Retrieval hard gate disabled; memory admission is score-ranked. \\
A11 & Forced consolidation & Global maintenance scope forced during consolidation. \\
\bottomrule
\end{tabular}
}%
\caption{Condition and baseline contract for the main paper. Source definitions:
\texttt{CONFIRMATORY-FOLD.md} and \texttt{PREREGISTRATION.md}.}
\label{tab:condition-contract}
\end{table}

B5 requires clarification because it anchors the headline comparison. B5 is a verbatim
event-memory policy and offline instance-memory substitute, Mem0-inspired but
\emph{not} stock Mem0. B5 writes exactly one deterministic memory per trajectory and does
not implement Mem0's online add/update/delete/no-op extraction decisions or Mem0$^g$'s
graph memory and obsolete-triple handling (\texttt{src/benchmarks/baselines.py}
documents B0--B7 as synthetic harness implementations, not claims of fidelity to
third-party systems). B6 is likewise a subtask-granularity substitute inspired by
structurally aligned subtask-level memory \citep{shen2026structurally}, not a fidelity
reimplementation of that published system. Neither comparison is a fidelity claim: the
non-significant reference-probe hybrid-vs.-B5 comparison is not a claim of beating a tuned Mem0,
and B6's result is not a claim about Shen et al.'s method.
The scaled fold evaluates one hosted Mem0 system family through two supplemental
configurations after the namespace and payload fixes described in
\filepath{analysis/investigation-evidence/MEM0-ROW-FOLD-REPORT.md}.  Both use
\texttt{mem0ai} 2.0.11, one reset \texttt{user\_id} namespace per
run/condition/seed/sequence (colon-free and prefixed by
\filepath{DREAMBENCH_NAMESPACE_PREFIX}), \texttt{top\_k=6} retrieval
(\texttt{memory\_context=6}), and a 1,200-token memory-read budget.  B5-MEM0 uses
hosted-default fact extraction.  B5-MEM0-LIT disables fact inference and stores
sanitized raw event text directly as a diagnostic for exact-literal handling.  Both
adapters attach benchmark, condition, seed, sequence, and run metadata and search only
the corresponding \texttt{user\_id}.  B5-MEM0 reaches $21/180=0.117$, and
B5-MEM0-LIT reaches $20/180=0.111$, each with zero validity-gate exclusions.  The
historical v1 B5-MEM0 result of $6/66=0.091$ is used only in the historical failure
analysis in Appendix~\ref{sec:appendix-mem0-failure}; no v2 inference uses that
denominator.  These two configurations are not two independent products and do not
support a ranking claim about tuned Mem0 or Mem0$^g$.

In the historical v1 fold, synthetic B0--B7 conditions completed with 66 S3 records
and 0 exclusions; reference-probe typed-only (DF) had one validity-gate exclusion
(\texttt{codex\_exec\_failed}; $n{=}65$), and DF-strict had 11 record-level exclusions
and was treated as incomplete.  In the scaled v2 fold, the complete listed conditions in
Section~\ref{sec:results} contribute 180 S3 cells each with zero excluded records.  The
reported ablations start from the reference probe typed-only: A0 episodic-only, A2 no
contradiction repair, A4 no counterfactual replay, A5 no stale suppression, A6 no
retrieval hard gate, and A11 forced consolidation. Designed-but-not-run ablations are
listed in Table~\ref{tab:operator-ablation-map}.

\subsection{Controls}

\paragraph{Successor external-audit design.}
The v2.1 audit was designed after the original v2 study and is therefore post-hoc to
that historical result, but its conditions, conformance gates, repair budgets, two
comparison families, unavailable-slot rule, and analysis were frozen before successor
outcome inspection.  It reused the same 60 traps and three repeated seeds under a
committed Linux execution lock.  Four conditions cleared conformance and entered the
canonical run: B0, B5, DF-hybrid, and the pinned hosted B5-MEM0-LIT configuration.
Native B5-MEM0 failed the frozen cross-process normalized-context identity gate;
B5-SM-LOCAL and B5-SM-LOCAL-DOC exhausted their frozen repair budgets.  These three
conditions were rejected before evaluation and remain registered unavailable slots,
not observed performance rows.  Family A contains six comparisons against B0; Family B
contains two cross-configuration mechanism contrasts.  Unavailable slots retain
$p=1$, and neither Holm family is shrunk.  The primary statistical unit is the trap;
the exact sign/permutation test exchanges labels at the 60 trap clusters, with the
three seed cells treated as repeated observations.  Pooled and per-seed analyses are
sensitivities only.

Within each comparison table, all conditions use the same wake-phase base model, model version, endpoint, context limit, task order, repository checkout rule, verification commands, tool budget, wake-token budget, and stopping criteria. Each condition receives the same read-token budget for memory inserted into context. The reference probe runs an LLM-curated sleep pipeline that writes multiple derived items per condition and applies lifecycle operations, whereas B5 writes one deterministic record per trajectory. We therefore do not claim baseline parity beyond the wake prompt budget, and we report sleep/judge cost explicitly rather than treating it as free (Section~\ref{sec:limitations}, cost frontier).

Table~\ref{tab:run-config} records the fold setup.

\begin{table}[!htbp]
\centering
\small
\begin{tabular}{p{0.26\linewidth}p{0.64\linewidth}}
\toprule
\textbf{Item} & \textbf{Reported fold setting} \\
\midrule
Wake agent and model & Codex CLI agent using \texttt{gpt-5.5}. \\
Provider route & Hosted OpenAI/Codex route through the configured Codex CLI credentials. \\
Sleep/judge model & \texttt{--judge-model codex-gpt-5.5}, mapped by the harness to \texttt{gpt-5.5}. \\
Agent container & \texttt{@openai/codex} 0.142.0 in the \texttt{scripts/Dockerfile.codex-agent} image on \texttt{node:22-slim}. \\
Primary fold run date & 2026-07-06, recorded by run stamp \texttt{20260706T074759Z}. \\
Seeds & Seeds 1, 2, and 3 control the harness seed, condition namespace, task ordering within the recorded grid, and any seeded sampling in the harness; hosted model nondeterminism can still remain. \\
Task records & \texttt{experiments/env/sequences\_confirmatory\_v2.jsonl}; hidden oracles and reference solutions are outside the wake-agent container filesystem. \\
Result isolation & \filepath{DREAMBENCH_RESULTS_ROOT} isolates result directories; default fallback is \texttt{experiments/results}. \\
\bottomrule
\end{tabular}
\caption{D\&B setup essentials for the reported confirmatory fold.  Source
pointers are the launch script, Codex-agent Dockerfile, and reproduction
manifest.}
\label{tab:run-config}
\end{table}

The preregistration required B0 S1+S2 warmup pass rate to be at least 0.80 so
that S3 failures could be interpreted cleanly as memory failures.  The canonical
v2 analyzer reports B0 warmup at $287/360=0.797$, narrowly below that threshold.
The fold is complete and its executable S3 outcomes remain reportable, but the
paper therefore does not treat every B0 S3 failure as an isolated memory-only
failure.

Two disclosures affect all conditions equally but bound external validity. First, the wake-phase agent container is \emph{edit-only}: the Codex agent image installs \texttt{git} and the Codex CLI but not Python or pytest (\texttt{scripts/Dockerfile.codex-agent}), so the wake agent cannot run the repository's Python tests inside its own container; the oracle scores the agent's produced diff externally. This makes the wake agents weaker than a full self-testing SWE agent and may interact with which memories matter, though it applies identically to every condition. Second, sleep cost is reported as a cost frontier in Section~\ref{sec:limitations}, not omitted.

\subsection{Metrics}

\begin{sloppypar}
The main task metrics are Pass@1 and \texttt{final\_passed}. Pass@1 excludes timeout-then-pass records; \texttt{final\_passed} reports the final executable oracle result. The reported confirmatory-fold memory-hygiene metrics are RepeatedErrorRate, HarmfulMemoryRate, RegressionAfterUpdate, ContradictionRepairAccuracy, HumanFeedbackUseAccuracy, UsefulMemoryPrecision, ScopeAccuracy, and TransferScore. For RepeatedErrorRate, HarmfulMemoryRate, and RegressionAfterUpdate, lower is better. For the remaining metrics, higher is better. Hygiene values are deterministic offline re-scores from result records. These eight metrics are not independent: several are task-coupled, and CRA, HFUA, and TransferScore are numerically identical to one another for each ladder condition in the fold (a judge artifact; Section~\ref{sec:limitations}), so Section~\ref{sec:results} collapses them into independent families rather than counting eight separate wins. HarmfulMemoryRate is 0.000 for every condition and is therefore diagnostic only in this fold. RegressionAfterUpdate is a diagnostic heuristic that scans oracle stdout/stderr for regression-indicating substrings on failed records; we treat it as diagnostic, not as a headline hygiene win.
\end{sloppypar}

\subsection{Run Protocol}

Before evaluation, the benchmark instances, prompts, verification commands, oracle labels, model configurations, budgets, condition definitions, and three seeds were fixed. Each condition starts from an empty condition-specific memory namespace. For each session, the harness logs raw trajectory, memory reads, memory writes, task outcome, final repository state, and isolation metadata. Sleep-phase maintenance then runs according to the condition schedule. The scorer aggregates task and memory-hygiene metrics overall and by condition. Negative results and failure modes are reported rather than filtered.

The exact fold analysis can be replayed from frozen records with:

\begin{CodeBlock}
DREAMBENCH_ROOT=. PYTHONPATH=src python3 scripts/analyze_confirmatory.py \
  --results-root "${DREAMBENCH_RESULTS_ROOT:-experiments/results}"
\end{CodeBlock}

This command rewrites the checked-in confirmatory report and folded JSON under
\path{analysis/} from the result records.  Replay from frozen records is exact up to the
checked-in records and analysis script.  Live reruns can drift because the wake agent,
sleep pipeline, and primary judge all use hosted \texttt{gpt-5.5}.
Consequently, a new live rerun is a new experiment even with the same seeds.

\paragraph{Artifact and data availability.}
The frozen original-v2 benchmark package is archived at
\url{https://github.com/iroiro147/dreambench-swe/releases/tag/v2.0.5}.
The separately preregistered external-systems successor study and the manuscript that
reports it are archived in the additive \texttt{v2.1.0} release at
\url{https://github.com/iroiro147/dreambench-swe/releases/tag/v2.1.0}; the
\texttt{v2.0.5} tag and assets remain unchanged.  The original package publishes the
sequence records, harness code, analysis scripts, container recipe, tests, and frozen
result summaries.  The successor release adds its sanitized evidence bundle, paper PDF,
flat arXiv source, manifest, and detached checksum ledger.  Both releases keep hidden
oracle answers, scorer-only generated secrets, and hosted-service secrets out of public
task prompts and logs.
Appendix~\ref{app:artifact} lists the artifact checksums and run commands.  The
reproducible fold command above is the analysis entry point; live grid launch is recorded
in \filepath{ops/launch_confirmatory.sh}.  Hidden-oracle handling is deliberately
asymmetric: agents never receive oracle files or reference solutions in their container
filesystem, while the public paper reports aggregate task outcomes and selected sanitized
tokens needed to explain failure modes.

\paragraph{Pretraining and local-leakage stance.}
The scored S3 contracts use fresh, arbitrary, non-inferable tokens and local repository
rules, so base-model pretraining should not contain the answer unless it saw the benchmark
records themselves.  That addresses pretraining leakage at the token level.  Local
leakage is handled separately by the container-filesystem boundary: hidden oracles,
reference solutions, sequence records, analysis files, logs, and the paper repository are
not mounted into the wake-agent container.  This is not a proof against all external
network leakage; it is the specific claim tested by the canary and filesystem-isolation artifacts.

\subsection{Isolation and Contamination Control}
\label{sec:isolation-boundary}

The isolation guarantee is filesystem-scoped: scored benchmark artifacts are absent from
the relevant container filesystems, but the run is not network-isolated. Each wake agent
runs inside an isolated OrbStack Linux container with only the current task worktree
mounted read-write and a read-only copied Codex home mounted for the model call. The benchmark repository, hidden oracles, reference
solutions, sequence records, analysis files, and logs are physically absent from the
container filesystem. Sleep judges also run in isolated containers with Codex
authentication mounted, no benchmark repository mounted, and no access to task worktrees
or oracle trees. The container path is fail-closed: if isolation is unavailable, the run
records an isolation failure rather than falling back to an unjailed agent. We do
\emph{not} claim network isolation: the container has a default network bridge (required
for the hosted model call) and receives a read-only copied Codex home, so external network
state is reachable in principle even though the task prompt instructs the agent not to
make network calls. The guarantee we make is that the scored benchmark material is outside
the container filesystem wall; hidden oracles, reference solutions, and sequence files
still live in the host repository tree, so the guarantee rests on that container wall and
would be reintroduced as a risk by a future debug mount or non-container path
(\filepath{analysis/fold/HERMETICITY-MANIFEST.md} records the exact mount set and network
policy for the reported run).

This protocol was adopted after an adversarial audit invalidated the first non-isolated run. That audit found that wake agents could read outside the task worktree and that the sleep judge could run from the repository root, making hidden oracles and reference solutions reachable. The first result set is therefore not used as positive evidence. Early non-isolated and seed-1 intermediate tables produced unstable baseline values, including B5 values near 0.73 before the isolation fix and 0.59 in a superseded seed-1 filesystem-isolated table; those values are historical diagnostics only, not current results. The historical v1 B5 result is $48/66=0.727$ in \filepath{analysis/investigation-evidence/CONFIRMATORY-FOLD.md}; the current v2 B5 result is $89/180=0.494$ from the canonical analyzer artifacts in \filepath{analysis/fold/}.

Because ``0 contamination'' cannot be a trusted metadata field after a prior leak, we
audit that boundary two ways. First, a live container-isolation acceptance suite was
executed from an \emph{unsandboxed} host shell and archived verbatim in
\filepath{analysis/fold/CANARY-PROOF.txt}: three tests passed (3/3), showing that the
wake-agent container cannot see \texttt{/Users}, the harness repository, the oracle tree,
the reference-solution tree, or the sequence file, and that both the containerized wake
agent and the sleep judge are blocked from a planted canary yet still function. Second,
the fold's \texttt{contaminated} hygiene column is treated as a conservative scanner
diagnostic over contributing result files, not as a validity-gate exclusion or as a claim
that hidden benchmark files were read. Local inspection of the nonzero flags found
hidden-marker hits in visible task or memory material (for example \texttt{arity=2} and
\texttt{1-based}), while the container-hidden-read filter requires hidden-path/read-context
evidence and found no oracle, reference-solution, or sequence-file reads in the reported
container records. We therefore claim the run is container-filesystem isolated from scored
benchmark artifacts and not network-isolated, with no evidence of hidden benchmark-file
reads; we do not claim that the raw diagnostic column is zero.

For independent reproduction we pin the dependencies that govern model behavior: the wake and judge model is \texttt{gpt-5.5}, and the agent container uses the Codex CLI (\texttt{@openai/codex}) on a \texttt{node:22-slim} base.  The paths \path{scripts/Dockerfile.codex-agent} and \path{analysis/fold/REPRO-MANIFEST.md} record the node image digest and run commands. Because \texttt{gpt-5.5} is a hosted model that may be a mutable alias, exact model behavior may not be independently reproducible even with the harness pinned.

%% file: figures/sequence_templates.tex
\begin{figure}[t]
\centering
\scriptsize
\resizebox{\linewidth}{!}{%
\begin{tikzpicture}[
  font=\sffamily,
  border/.style={draw=black!45, line width=0.55pt},
  grid/.style={draw=black!22, line width=0.45pt},
  head/.style={align=center, font=\sffamily\bfseries\scriptsize, inner sep=2pt},
  cell/.style={align=left, font=\sffamily\scriptsize, inner sep=2pt},
  fam/.style={align=center, font=\sffamily\bfseries\scriptsize, inner sep=2pt}
]
\fill[black!8] (0,0) rectangle (18.5,-0.78);
\fill[blue!3] (0,-0.78) rectangle (18.5,-2.18);
\fill[green!3] (0,-3.58) rectangle (18.5,-4.98);
\fill[orange!3] (0,-6.38) rectangle (18.5,-7.78);

\draw[border] (0,0) rectangle (18.5,-7.78);
\foreach \x in {2.25,6.25,10.60,14.25} {
  \draw[grid] (\x,0) -- (\x,-7.78);
}
\foreach \y in {-0.78,-2.18,-3.58,-4.98,-6.38} {
  \draw[grid] (0,\y) -- (18.5,\y);
}

\node[head, text width=1.95cm] at (1.125,-0.39) {Family};
\node[head, text width=3.65cm] at (4.25,-0.39) {S1/S2 injected evidence};
\node[head, text width=3.95cm] at (8.425,-0.39) {Expected memory before S3};
\node[head, text width=3.25cm] at (12.425,-0.39) {S3 hidden-oracle dependency};
\node[head, text width=3.85cm] at (16.375,-0.39) {Expected failure mode};

\node[fam, text width=1.75cm] at (1.125,-1.48) {Convention-\\learning};
\node[cell, text width=3.55cm] at (4.25,-1.48) {S1 observes a project rule with \texttt{SYNTH-RULE}.\\S2 reinforces the same local convention.};
\node[cell, text width=3.90cm] at (8.425,-1.48) {\textbf{Admit:} scoped procedural or convention memory.\\\textbf{Suppress:} out-of-scope variant.};
\node[cell, text width=3.20cm] at (12.425,-1.48) {Oracle expects the redacted rule behavior, not an inferable default.};
\node[cell, text width=3.75cm] at (16.375,-1.48) {Missing, byte-wrong, or unscoped rule causes the S3 patch to violate the oracle.};

\node[fam, text width=1.75cm] at (1.125,-2.88) {Generated\\files};
\node[cell, text width=3.55cm] at (4.25,-2.88) {S1 shows generated output.\\S2 exposes source-of-truth plus regeneration workflow.};
\node[cell, text width=3.90cm] at (8.425,-2.88) {\textbf{Admit:} source file and generator procedure.\\\textbf{Suppress:} stale generated snapshot.};
\node[cell, text width=3.20cm] at (12.425,-2.88) {Oracle checks \texttt{REDACTED-GEN} after regeneration/application.};
\node[cell, text width=3.75cm] at (16.375,-2.88) {Editing generated artifacts or omitting the workflow leaves S3 behavior incomplete.};

\node[fam, text width=1.75cm] at (1.125,-4.28) {Stale\\architecture};
\node[cell, text width=3.55cm] at (4.25,-4.28) {S1 records an older ownership or routing fact.\\S2 moves the boundary.};
\node[cell, text width=3.90cm] at (8.425,-4.28) {\textbf{Admit:} current architecture with validity scope.\\\textbf{Suppress:} superseded owner/path.};
\node[cell, text width=3.20cm] at (12.425,-4.28) {Oracle depends on the current target, not the historical one.};
\node[cell, text width=3.75cm] at (16.375,-4.28) {Stale recall sends the implementation to the wrong module or invariant.};

\node[fam, text width=1.75cm] at (1.125,-5.68) {Reviewer-\\preference};
\node[cell, text width=3.55cm] at (4.25,-5.68) {S1 gives feedback with a redacted dialect token.\\S2 narrows or revises the preference.};
\node[cell, text width=3.90cm] at (8.425,-5.68) {\textbf{Admit:} latest scoped feedback memory.\\\textbf{Suppress:} over-scoped old preference.};
\node[cell, text width=3.20cm] at (12.425,-5.68) {Oracle expects \texttt{REDACTED-REV} only in its scoped context.};
\node[cell, text width=3.75cm] at (16.375,-5.68) {Overgeneralization or old feedback applies the wrong review dialect.};

\node[fam, text width=1.75cm] at (1.125,-7.08) {Flaky\\test};
\node[cell, text width=3.55cm] at (4.25,-7.08) {S1 captures nondeterministic failure evidence.\\S2 supplies rerun/context evidence.};
\node[cell, text width=3.90cm] at (8.425,-7.08) {\textbf{Admit:} uncertainty note and guardrail.\\\textbf{Suppress:} unsupported causal lesson.};
\node[cell, text width=3.20cm] at (12.425,-7.08) {Oracle checks that S3 avoids the spurious causal fix.};
\node[cell, text width=3.75cm] at (16.375,-7.08) {A false causal memory induces unnecessary or harmful code changes.};
\end{tikzpicture}%
}
\caption{\bench{} sequence-template skeleton used for the reported confirmatory fold. Tokens shown in the schematic are synthetic or redacted placeholders; the per-trap audit is in Appendix~\ref{app:verbatim}.}
\label{fig:templates}
\end{figure}

%% file: sections/07_results.tex
\section{Results}
\label{sec:results}

\paragraph{Scope.}
We report the scaled v2 confirmatory fold: 60 traps over three seeds, with every listed
condition contributing 180 validity-gated S3 cells and zero excluded records.  The fold
ran from the frozen confirmatory sequence file (SHA-256 prefix
\texttt{4966bad1e535}; full digest in Appendix~\ref{app:artifact}).  All v2 numbers in
this section come from the canonical analyzer artifacts in \path{analysis/fold/} and are
interpreted under the re-anchor: \bench{} is the headline artifact, while the evaluated
maintenance pipeline is a reference probe.

\subsection{V2 Baseline Sweep and Construct Strata}
\label{sec:results-baselines}
\label{sec:results-ladder}

The main v2 ladder is shown in Table~\ref{tab:v2-main-ladder}.  The strongest verbatim
baseline, B5, reaches $89/180=0.494$.  The hybrid reference probe reaches
$95/180=0.528$, typed-only reaches $80/180=0.444$, and raw-only reaches
$84/180=0.467$.  These pooled rates are descriptive; the primary inference is the
clustered P-family in Section~\ref{sec:results-mcnemar}.

\input{analysis/fold/v2_tables}

Table~\ref{tab:v2-construct-strata} is also descriptive.  It shows that the v2
instrument is not a single uniform construct: C1 remains mostly verbatim retention,
where B5 is strong; C8 and C9 are anti-hoarding or source-of-truth strata where
reference-probe variants can look directionally better, but both are small-n strata and C9
fails the strengthened B0-headroom validity rule below.  We therefore do not draw
directional per-construct claims for strata flagged as small-n.

\subsection{Clustered P-Family}
\label{sec:results-mcnemar}

Table~\ref{tab:v2-p-family} reports the preregistered clustered P-family.  No comparison
rejects after Holm correction.  The primary comparison, P1, is the result that controls
the paper's system interpretation: reference-probe hybrid does not significantly separate from B5.

\begin{table}[!htbp]
\centering
\small
\resizebox{\textwidth}{!}{%
\begin{tabular}{llrrrrr}
\toprule
ID & Comparison & Signed statistic & Permutation $p$ & Holm $p$ & Reject & First/second wins \\
\midrule
P1 & reference-probe hybrid vs.\ B5 & $+6$ & 0.518419 & 1.000 & no & 21 / 15 \\
P2 & reference-probe typed-only vs.\ B5 & $-9$ & 0.482597 & 1.000 & no & 22 / 31 \\
P3 & reference-probe raw-only vs.\ B5 & $-5$ & 0.489746 & 1.000 & no & 7 / 12 \\
P4 & reference-probe hybrid vs.\ reference-probe raw-only & $+11$ & 0.127867 & 0.767 & no & 22 / 11 \\
P5 & reference-probe hybrid vs.\ reference-probe typed-only & $+15$ & 0.134251 & 0.767 & no & 33 / 18 \\
P6 & reference-probe typed-only vs.\ B3 & $+13$ & 0.279096 & 1.000 & no & 32 / 19 \\
\bottomrule
\end{tabular}
}%
\caption{V2 trap-clustered P-family with Holm correction.  Signed statistic is
positive when the first condition has more trap-level wins.  Null outcomes are
failures to reject, not equivalence tests.  Source: canonical clustered
analyzer JSON.}
\label{tab:v2-p-family}
\end{table}

The P1 grid is complete: 60 traps, three seeds, 180 paired cells, no missing first cells,
no missing second cells, and zero excluded records for both reference-probe hybrid and B5.  The
descriptive pooled difference is small ($95/180$ vs.\ $89/180$), and the clustered test
does not reject.  The v1 result is consistent with this interpretation: in the 22-trap
fold, reference-probe hybrid reached $52/66=0.788$ versus B5 $48/66=0.727$, but the exact McNemar
comparison was also non-significant ($b{=}12$, $c{=}8$, $p{=}0.503$; Holm $p{=}1.0$).

\subsection{Admission Funnel and Construct Validity}
\label{sec:results-taxonomy}

The v2 admission funnel is clean: 30 net-new traps were authored, 30 were dry-valid, 30
were live-valid, and 30 were admitted.  There are no authored-not-admitted,
dry-valid-not-live-valid, or live-valid-not-admitted rows in the canonical funnel.  This
is evidence of process completion, not a license for broad validity language.

The strengthened stratum-validity check fails overall.  C2, C3, C5, C6, and C7 pass the
headroom/spread checks used by the analyzer.  C9 fails because B0 passes all C9 S3 cells
($12/12=1.000$), so the stratum does not have the intended no-memory headroom.  C10 also
fails B0 headroom ($6/6=1.000$).  C9 and C10 are therefore invalid as anti-hoarding or
abstention evidence in this fold: because B0 passes every cell in these strata, they
contain no measured memory-dependent signal.  After excluding C9 and C10, the fold retains
24 traps across C2, C3, C5, C6, and C7 that satisfy the preregistered headroom criterion, above
the 20-trap floor.  This count checks design coverage only and does not establish
anti-hoarding or abstention validity.
These failures are reported as benchmark limitations; they are not used to repair the null
P1 outcome.

\begin{table}[!htbp]
\centering
\small
\begin{tabular}{lrrrrl}
\toprule
Stratum & Max condition rate & Spread & B0 rate & Holds? & Note \\
\midrule
C2 & 0.833 & 0.833 & 0/18 = 0.000 & yes & retrieval precision \\
C3 & 0.667 & 0.667 & 0/15 = 0.000 & yes & staleness/supersession \\
C5 & 0.222 & 0.222 & 0/18 = 0.000 & yes & scope discipline \\
C6 & 0.250 & 0.250 & 0/12 = 0.000 & yes & contradiction/provenance \\
C7 & 0.333 & 0.333 & 0/9 = 0.000 & yes & cross-session synthesis \\
C9 & 1.000 & 1.000 & 12/12 = 1.000 & no & invalid: B0-headroom fail \\
C10 & -- & -- & 6/6 = 1.000 & no & invalid: B0-headroom fail \\
\bottomrule
\end{tabular}
\caption{V2 construct-validity headroom summary.  The overall strengthened F7
result is false because C9 fails; C10 B0 headroom also fails.  C9 and C10 are
invalid as anti-hoarding or abstention evidence in this fold by the B0-headroom
criterion.  Source: canonical post-fold analysis packet.}
\label{tab:v2-stratum-validity}
\end{table}

\subsection{Clean-start and Hygiene Diagnostics}
\label{sec:results-hygiene}

The clean-start subset contains 13 traps, or 39 cells per condition.  In that subset,
reference-probe hybrid reaches $19/39=0.487$ and B5 reaches $15/39=0.385$; paired cells are
first-wins 9, second-wins 5, both-pass 10, and both-fail 15.  This is descriptive and
not claim-rescuing: B3 reaches $22/39=0.564$ on the same subset.

Executable hygiene diagnostics expose failure modes across the full v2 fold.  The
aggregate oracle metrics are \texttt{IrrelevantImportRate} $1434/3024=0.474$,
\texttt{OverscopeRate} $378/2268=0.167$, \texttt{RepeatedErrorRate}
$1512/2646=0.571$, and \texttt{StaleUseRate} $756/3780=0.200$.  These are treated as
diagnostics for where the benchmark catches memory failures, not as a separate system-win
family.

\begin{table}[!htbp]
\centering
\small
\begin{tabular}{lrrr}
\toprule
Metric & Numerator & Denominator & Rate \\
\midrule
\texttt{IrrelevantImportRate} & 1434 & 3024 & 0.474 \\
\texttt{OverscopeRate} & 378 & 2268 & 0.167 \\
\texttt{RepeatedErrorRate} & 1512 & 2646 & 0.571 \\
\texttt{StaleUseRate} & 756 & 3780 & 0.200 \\
\bottomrule
\end{tabular}
\caption{Executable v2 hygiene-oracle aggregate metrics.  Source: canonical v2
hygiene-oracle JSON.}
\label{tab:v2-hygiene-oracle}
\end{table}

\subsection{Probe Ablations and Raw-evidence Checks}
\label{sec:results-ablations}
\label{sec:results-capsule}

The v2 fold includes several probe ablations as diagnostic conditions, all with 180 S3
cells.  A2 reaches $88/180=0.489$, A4 reaches $77/180=0.428$, A5 reaches
$79/180=0.439$, A6 reaches $72/180=0.400$, A11 reaches $70/180=0.389$, and A0 reaches
$20/180=0.111$.  These numbers do not establish independent operator causality; P4 and
P5, the planned marginal tests around raw-only, typed-only, and hybrid, do not reject.

The raw-evidence invariant is therefore best read as an auditability and measurement
property rather than a proven causal mechanism.  Raw episodes remain available for
inspection, and derived memories can be marked stale or gated, but the fold does not show
that the full typed pipeline significantly improves over raw-only or B5 under the primary
clustered family.

\input{sections/07_external_audit_v21}

%% file: analysis/fold/v2_tables.tex
% Generated by scripts/generate_tables_v2.py from a frozen v2 fold JSON.

\begin{table}[t]
\centering
\small
\begin{tabular}{llrrll}
\toprule
Condition & Wake backbone(s) & Traps & Cells & Pass@1 & Clustered 95\% CI \\
\midrule
B0 No memory & Codex / GPT-5.5 & 60 & 180 & 21/180 (0.117) & [0.050, 0.200] \\
B1 Raw episodic & Codex / GPT-5.5 & 60 & 180 & 21/180 (0.117) & [0.050, 0.200] \\
B2 Vector traces & Codex / GPT-5.5 & 60 & 180 & 20/180 (0.111) & [0.033, 0.194] \\
B3 Reflection-vector & Codex / GPT-5.5 & 60 & 180 & 67/180 (0.372) & [0.250, 0.489] \\
B4 Untyped summary & Codex / GPT-5.5 & 60 & 180 & 21/180 (0.117) & [0.050, 0.200] \\
B5 Instance memory & Codex / GPT-5.5 & 60 & 180 & 89/180 (0.494) & [0.378, 0.617] \\
B6 Subtask memory & Codex / GPT-5.5 & 60 & 180 & 21/180 (0.117) & [0.050, 0.200] \\
B7 Task tracker & Codex / GPT-5.5 & 60 & 180 & 19/180 (0.106) & [0.033, 0.189] \\
Probe typed-only & Codex / GPT-5.5 & 60 & 180 & 80/180 (0.444) & [0.333, 0.556] \\
Probe raw-only & Codex / GPT-5.5 & 60 & 180 & 84/180 (0.467) & [0.350, 0.578] \\
Probe hybrid & Codex / GPT-5.5 & 60 & 180 & 95/180 (0.528) & [0.422, 0.633] \\
Probe strict & Codex / GPT-5.5 & 60 & 180 & 42/180 (0.233) & [0.144, 0.328] \\
Probe strict-hybrid & Codex / GPT-5.5 & 60 & 180 & 78/180 (0.433) & [0.333, 0.533] \\
A0 & Codex / GPT-5.5 & 60 & 180 & 20/180 (0.111) & [0.033, 0.194] \\
A11 & Codex / GPT-5.5 & 60 & 180 & 70/180 (0.389) & [0.283, 0.489] \\
A2 & Codex / GPT-5.5 & 60 & 180 & 88/180 (0.489) & [0.383, 0.594] \\
A4 & Codex / GPT-5.5 & 60 & 180 & 77/180 (0.428) & [0.322, 0.539] \\
A5 & Codex / GPT-5.5 & 60 & 180 & 79/180 (0.439) & [0.322, 0.544] \\
A6 & Codex / GPT-5.5 & 60 & 180 & 72/180 (0.400) & [0.294, 0.511] \\
\bottomrule
\end{tabular}
\caption{V2 main ablation ladder S3 Pass@1 for the primary wake backbone. Hosted live-memory baselines are excluded from this headline table and reserved for failure analysis. Intervals are trap-cluster bootstrap 95\% CIs; seeds are repeated observations within traps.}
\label{tab:v2-main-ladder}
\end{table}

\begin{table}[t]
\centering
\scriptsize
\resizebox{\textwidth}{!}{%
\begin{tabular}{lrrllll}
\toprule
Construct & Traps & Cells & B5 & Probe hybrid & Best observed & Small-n \\
\midrule
C1 Verbatim retention & 18 & 1026 & 48/54 (0.889) & 46/54 (0.852) & B5 Instance memory 48/54 (0.889) & no \\
C2 Retrieval precision & 6 & 342 & 15/18 (0.833) & 12/18 (0.667) & B5 Instance memory 15/18 (0.833) & no \\
C3 Staleness/supersession & 7 & 399 & 13/21 (0.619) & 8/21 (0.381) & B5 Instance memory 13/21 (0.619) & no \\
C4 Update propagation & 1 & 57 & 0/3 (0.000) & 0/3 (0.000) & A4 1/3 (0.333) & yes \\
C5 Scope discipline & 6 & 342 & 4/18 (0.222) & 3/18 (0.167) & B5 Instance memory 4/18 (0.222) & no \\
C6 Contradiction/provenance & 5 & 285 & 3/15 (0.200) & 3/15 (0.200) & A2 4/15 (0.267) & yes \\
C7 Cross-session synthesis & 7 & 399 & 6/21 (0.286) & 7/21 (0.333) & Probe raw-only 7/21 (0.333) & no \\
C8 Procedural/source-of-truth & 4 & 228 & 0/12 (0.000) & 7/12 (0.583) & Probe hybrid 7/12 (0.583) & yes \\
C9 Spurious-lesson rejection & 4 & 228 & 0/12 (0.000) & 8/12 (0.667) & Probe typed-only 12/12 (1.000) & yes \\
C10 Abstention & 2 & 114 & 0/6 (0.000) & 1/6 (0.167) & Probe typed-only 6/6 (1.000) & yes \\
\bottomrule
\end{tabular}
}%
\caption{V2 per-construct S3 strata C1--C10, descriptive only. Hosted live-memory baselines are excluded from headline construct summaries. Small-n is flagged when a construct has fewer than 6 trap clusters with observed cells.}
\label{tab:v2-construct-strata}
\end{table}

% V2 cross-backbone table omitted: the GLM-5.2 backbone was scoped out, so no GLM S3 cells are present in the confirmatory fold.

%% file: sections/07_external_audit_v21.tex
\subsection{Separately Preregistered v2.1 External-Systems Audit}
\label{sec:results-v21-external}

After the original v2 study, we designed a successor external-systems audit and froze
its conditions, conformance gates, comparison families, multiplicity rules, and analysis
before inspecting successor outcomes.  This audit is additive to, not a replacement for,
the immutable v2.0.5 analysis above.

\paragraph{Completion and profile.}
The canonical audit completed all
$\vOneCompletedUnits/\vOnePlannedUnits$ planned work units and all
$\vOneValidCells/\vOnePlannedCells$
validity-gated S3 cells.  Each of B0, B5, DF-hybrid, and B5-MEM0-LIT contributed
$180/180$ valid cells.  There were no missing, invalid, off-corpus, retry-exhausted,
contaminated, forbidden-base-reference, or kill-violation records.  These predicates
establish execution admissibility; file and receipt counts are not performance outcomes.
Table~\ref{tab:v21-condition-profile} reports the four-condition profile.

\begin{table}[!htbp]
\centering
\small
\begin{tabular}{lrrrr}
\toprule
Condition & Passed & Valid S3 & Rate & Trap-clustered 95\% CI \\
\midrule
B0 & \vOneBZeroPassed & \vOneBZeroValid & \vOneBZeroRate & [\vOneBZeroCILow, \vOneBZeroCIHigh] \\
B5 & \vOneBFivePassed & \vOneBFiveValid & \vOneBFiveRate & [\vOneBFiveCILow, \vOneBFiveCIHigh] \\
DF-hybrid & \vOneDFHybridPassed & \vOneDFHybridValid & \vOneDFHybridRate & [\vOneDFHybridCILow, \vOneDFHybridCIHigh] \\
B5-MEM0-LIT & \vOneMemZeroLitPassed & \vOneMemZeroLitValid & \vOneMemZeroLitRate & [\vOneMemZeroLitCILow, \vOneMemZeroLitCIHigh] \\
\bottomrule
\end{tabular}
\caption{Canonical v2.1 successor condition profile.  Seeds are repeated
observations within 60 trap clusters.  Intervals are equal-trap-weighted normal
intervals over trap means, not pairwise equivalence intervals.  Marginal rates are
descriptive; inferential claims are limited to registered comparisons.}
\label{tab:v21-condition-profile}
\end{table}

\paragraph{Registered Family A: benchmark discrimination.}
Family A retained all six preregistered slots and used Holm correction at
$\alpha=0.05$; unavailable slots remained in the family with $p=1$.  Both concurrent
positive controls separated from no external memory: B5 versus B0 had signed statistic
$\vOneBFiveBZeroSigned$, raw clustered $p=\vOneBFiveBZeroRawP$, and
Holm-adjusted $p=\vOneBFiveBZeroHolmP$;
DF-hybrid versus B0 had signed statistic $\vOneDFHybridBZeroSigned$, raw clustered
$p=\vOneDFHybridBZeroRawP$, and Holm-adjusted $p=\vOneDFHybridBZeroHolmP$.
The one admitted external configuration also separated from B0: B5-MEM0-LIT versus
B0 had signed statistic $\vOneMemZeroLitBZeroSigned$, raw clustered
$p=\vOneMemZeroLitBZeroRawP$, and Holm-adjusted
$p=\vOneMemZeroLitBZeroHolmP$.  B5-MEM0, B5-SM-LOCAL, and
B5-SM-LOCAL-DOC were rejected by frozen pre-evaluation conformance gates and are
unavailable, not observed losses.

\begin{table}[!htbp]
\centering
\scriptsize
\resizebox{\textwidth}{!}{%
\begin{tabular}{lllrrl}
\toprule
Family & Registered comparison & Availability & Raw clustered $p$ & Holm $p$ & Interpretation \\
\midrule
A & B5 vs. B0 & available & \vOneBFiveBZeroRawP & \vOneBFiveBZeroHolmP & rejects \\
A & DF-hybrid vs. B0 & available & \vOneDFHybridBZeroRawP & \vOneDFHybridBZeroHolmP & rejects \\
A & B5-MEM0 vs. B0 & unavailable & 1 & 1 & pre-evaluation rejection \\
A & B5-MEM0-LIT vs. B0 & available & \vOneMemZeroLitBZeroRawP & \vOneMemZeroLitBZeroHolmP & rejects \\
A & B5-SM-LOCAL vs. B0 & unavailable & 1 & 1 & pre-evaluation rejection \\
A & B5-SM-LOCAL-DOC vs. B0 & unavailable & 1 & 1 & pre-evaluation rejection \\
B & B5-MEM0-LIT vs. B5-MEM0 & unavailable & 1 & 1 & no mechanism contrast \\
B & B5-SM-LOCAL-DOC vs. B5-SM-LOCAL & unavailable & 1 & 1 & no mechanism contrast \\
\bottomrule
\end{tabular}}
\caption{Preregistered successor comparison families.  Unavailable slots retain
raw and adjusted $p=1$; neither family was shrunk.  Family A supports benchmark
discrimination against B0.  Family B contains no evaluable mechanism contrast.}
\label{tab:v21-registered-families}
\end{table}

\paragraph{Family B and secondary comparisons.}
Neither registered mechanism contrast was evaluable.  Native B5-MEM0 failed the
frozen cross-process normalized-context identity gate, and both Supermemory conditions
exhausted their frozen repair budgets before evaluation.  Family B therefore supplies
no evidence about literal-versus-native Mem0 or documentation-mediated Supermemory
mechanisms; unavailable is neither a null result nor inferiority.

B5-MEM0-LIT was numerically higher than B5
($\vOneMemZeroLitPassed/\vOneMemZeroLitValid$ versus
$\vOneBFivePassed/\vOneBFiveValid$).  That secondary, nonconfirmatory clustered
comparison had signed statistic $\vOneLitBFiveSigned$ and raw
$p=\vOneLitBFiveRawP$, but it was not robust to the trap-majority-collapse sensitivity
($p=\vOneLitBFiveMajorityP$); per-seed values were
$\vOneLitBFiveSeedOneP$, $\vOneLitBFiveSeedTwoP$, and
$\vOneLitBFiveSeedThreeP$.  We therefore
do not claim superiority.  B5-MEM0-LIT was also numerically higher than DF-hybrid
($\vOneMemZeroLitPassed/\vOneMemZeroLitValid$ versus
$\vOneDFHybridPassed/\vOneDFHybridValid$), but the secondary clustered comparison did
not reject ($p=\vOneLitDFRawP$; majority-collapse $p=\vOneLitDFMajorityP$).  Failure to reject is not
equivalence.

\paragraph{Temporal and construct diagnostics.}
Concurrent B0 reproduced its historical reference rate exactly.  B5 and DF-hybrid
remained inside the frozen absolute temporal-drift bands, with absolute deviations
$\vOneBFiveDrift$ and $\vOneDFHybridDrift$, respectively, and both positive controls retained the expected
direction relative to B0.  These bands are diagnostics, not equivalence margins.
Successor construct rows cover only the 30 net-new traps carrying inline construct
labels; the other 30 carried traps were not remapped post hoc.  They therefore make no
full C1--C10 successor-profile claim.  Within that bounded subset, C2, C3, C5, C6, and
C7 are directional descriptions, while C9 and C10 retain the original B0-headroom
failure (B0 passes all
$\vOneCnineBZeroPassed/\vOneCnineBZeroValid$ C9 and
$\vOneCtenBZeroPassed/\vOneCtenBZeroValid$ C10 cells).  Neither construct profiles nor hygiene metrics substitute
for the unavailable Family B mechanism tests.

The successor result therefore supports \bench{} as a discriminating executable
profile benchmark and characterizes one exact hosted literal-storage configuration.
It does not establish an external-system mechanism, superiority among memory-bearing
conditions, equivalence, tuned-Mem0 performance, Supermemory performance, or broad
product generality.

%% file: sections/08_analysis.tex
\section{Analysis}

The v2 fold changes the paper's center of gravity.  The supported contribution
is \bench{} as a validity-gated benchmark and measurement protocol, with
the evaluated maintenance pipeline acting as a reference probe.  The scaled fold does not establish the
headline system claim that reference-probe hybrid is better than the strong verbatim
event-memory baseline B5.

\paragraph{Primary comparison.}
P1 compares \texttt{DF-hybrid} with B5 over 60 trap clusters and 180 paired S3
cells.  \texttt{DF-hybrid} is numerically higher in pooled S3 rate
($95/180=0.528$ versus $89/180=0.494$), but the trap-clustered exact
sign/permutation statistic is only \(+6\), with permutation
\(p=0.5184192657470703\), Holm-adjusted \(p=1.0\), and no rejection.  The
secondary pooled McNemar sensitivity is also non-significant
(\(b=21,c=15,p=0.405\), Holm \(p=0.815\)).  This is a failure to reject, not
equivalence and not evidence that the two systems are practically tied.

\paragraph{Family-level reading.}
All six pre-registered clustered comparisons are non-rejections after Holm
correction.  Typed-only memory does not beat B5 in v2
(\texttt{DF} \(80/180=0.444\) versus B5 \(89/180=0.494\); P2 Holm \(p=1.0\)).
Raw-only also does not beat B5 (\(84/180=0.467\); P3 Holm \(p=1.0\)).
\texttt{DF-hybrid} is numerically above raw-only and typed-only, but P4 and P5
remain non-significant after clustered Holm correction.  The v1 positive
structured-memory comparison against B3 does not replicate as a corrected v2
claim: \texttt{DF} reaches \(80/180=0.444\) versus B3 \(67/180=0.372\), but P6
has permutation \(p=0.279\), Holm \(p=1.0\), and no rejection.  The system
story is therefore diagnostic and mechanistic, not confirmatory.

\paragraph{Benchmark validity.}
The admission and completion gates support the benchmark as an executed
measurement artifact: 30 authored new traps passed dry validation, live
validation, and admission; the fold produced \(1890/1890\) condition-level result
files; completion dry plans had zero remaining \texttt{RUN} lines; and the
post-analyzer checker reported no failures.  The construct distribution also
does useful work: B0 is near floor overall (\(21/180=0.117\)), while strong
memory baselines separate on C2, C3, C5, C6, and C7.

The validity claim is constrained by the strengthened stratum gate.  C9 and C10
are invalid as anti-hoarding or abstention evidence in this fold by the
pre-registered B0-headroom criterion: B0 passes all 12 C9 S3 cells and all 6 C10
S3 cells, so neither stratum contains measured no-memory headroom.  The paper can
claim a benchmark that executed and exposed differentiated failure modes, but it
cannot claim that every anti-hoarding or abstention stratum is cleanly validated
at scale.  The remaining anti-hoarding constructs C2, C3, C5, C6, and C7 contribute
24 traps, still above the 20-trap floor.

\paragraph{Hygiene diagnostics.}
The hygiene panel is useful as error taxonomy, not as standalone proof of a
system win.  \texttt{DF-hybrid} improves \texttt{RepeatedErrorRate} relative to
B5 (0.142 versus 0.225) and is directionally higher on the task-coupled
memory-use diagnostics, but it misses important pre-registered hygiene
constraints.  \texttt{UsefulMemoryPrecision} fails as a system finding for
\texttt{DF-hybrid} (0.264 versus 0.556 for B5): the typed retrieval pipeline, as
implemented here, retrieves a lower fraction of useful memories than simple
verbatim event storage.  \texttt{RegressionAfterUpdate} also remains above the
strict target (0.106).  The construct-level oracle aggregates show large trip
rates for irrelevant imports, stale use, overscope, and repeated errors; the
aggregate \texttt{RepeatedErrorRate} includes low-memory and disabled-memory
conditions, so \texttt{DF-hybrid}'s per-system 0.142 rate is the relevant
pipeline diagnostic.  These are benchmark diagnostics, not evidence that the
maintenance pipeline has solved memory hygiene.

\paragraph{Pinned hosted Mem0.}
B5-MEM0 and B5-MEM0-LIT are retained only as supplemental failure analysis for
one hosted Mem0 system family under two pinned configurations.  In v2 they sit near the no-memory
floor: B5-MEM0 reaches \(21/180=0.117\), and B5-MEM0-LIT reaches
\(20/180=0.111\).  Those numbers are informative about this exact pinned
system family and configurations, including the historical token-normalization failure discussed
in Section~\ref{sec:appendix-mem0-failure}; they do not rank tuned Mem0 or
Mem0$^g$.

\paragraph{Cost diagnostics.}
The v2 cost frontier is not a rescue claim.  The analyzer estimates
the raw-only probe at \(0.03556\) USD per successful task, the hybrid probe
at \(0.03832\), and B5 at \(0.03953\).  That means the pre-registered
``more than \(10\times\) B5'' cost falsifier does not trip in v2, but it also
does not change the null task comparison.  Cost remains
a diagnostic constraint on future benchmark expansions and system variants.

\paragraph{Successor external-audit reading.}
The separately frozen v2.1 audit strengthens the benchmark claim while narrowing the
external-system claim.  In its registered six-slot Family A, both concurrent positive
controls and the one admitted hosted literal-storage configuration separate from B0
after Holm correction.  This establishes that the canonical successor run discriminates
memory-bearing behavior from no external memory.  It does not explain the external
configuration's numerical lead.  The native-versus-literal Mem0 contrast and the
Supermemory documentation contrast were both unavailable after frozen pre-evaluation
conformance rejection, so Family B supplies no mechanism evidence.  The literal
configuration's comparison with B5 is secondary and sensitivity-dependent, and its
comparison with DF-hybrid does not reject.  The defensible endpoint is therefore a
complete external-system profile with concurrent controls and transparent conformance
attrition, not superiority, equivalence, or causal attribution.

\paragraph{Bottom line.}
The v2 result is a credible negative result with useful instrumentation.  It
supports \bench{} as a reproducible stress test for multi-session SWE memory
hygiene, and it shows that simple verbatim event memory is a difficult baseline
for this fold.  It does not show that reference-probe hybrid beats B5, that typed memory
operators independently improve task success, that the result transfers to a
second wake model, or that the maintenance pipeline has a favorable cost profile.
The additive v2.1 result does not revise that original system comparison; it shows that
the benchmark discriminates the concurrent controls from B0 and characterizes one exact
hosted configuration under a separate preregistration.

%% file: sections/09_limitations.tex
\section{Limitations}
\label{sec:limitations}

The reference probe is an external memory-maintenance system, not a new foundation model.
The paper does not test whether larger context windows are unnecessary, whether
all agents need the same memory schema, or whether offline consolidation is
always beneficial.  The v2 fold explicitly shows that simple baselines can
remain competitive with a richer maintenance pipeline.

\paragraph{Failure to reject is not equivalence.}
The primary v2 result is null: \texttt{DF-hybrid} does not beat B5 under the
pre-registered trap-clustered Holm-corrected test.  The numerical direction
(\(95/180\) versus \(89/180\)) should not be read as a hidden win, and the
non-rejection should not be read as equivalence.  A future superiority or
equivalence claim would require a new preregistration with an appropriate
sample-size and equivalence-margin plan.

\paragraph{Benchmark-validity caveats.}
The fold executed cleanly, but not every construct stratum validates cleanly.
The preregistered B0 warmup interpretation gate also narrowly misses: B0 reaches
$287/360=0.797$ against the required 0.80.  The S3 results remain executable
outcomes, but B0 S3 failures cannot all be interpreted cleanly as isolated
memory failures.
The strengthened stratum-validity gate fails on C9 because B0 passes
\(12/12\) C9 cells; C10 also lacks no-memory headroom with B0 at \(6/6\).  These
strata are invalid as anti-hoarding or abstention evidence in this fold.  The
paper cannot claim uniformly validated anti-hoarding and abstention strata; it
can claim that the benchmark exposes differentiated failure modes and that the
disclosed C9/C10 failures constrain the interpretation.

\paragraph{Useful-memory precision failure.}
The pre-registered \texttt{UsefulMemoryPrecision} falsifier fails:
\texttt{DF-hybrid} achieves 0.264 versus B5's 0.556.  The typed retrieval
pipeline, as implemented in this probe, does not improve the fraction of
retrieved memories that are useful for the task.  This is a concrete system
limitation independent of the null P1 result.

\paragraph{Single wake model and no transfer result.}
The primary fold uses one wake model family, \texttt{gpt-5.5}, through the Codex
CLI harness.  As a scoped-out limitation, the planned second-backbone check did
not produce a full transfer fold: GLM-5.2 did not clear the warmup feasibility
path, so no cross-backbone table is reported and no broad model-generality
language is justified.  This does not affect the executable S3 oracles inside
the primary fold, but it does limit claims about whether the same memory
failures and relative rankings hold for other wake models.

\paragraph{Controlled benchmark scope.}
\bench{} uses curated fixture repositories, injected multi-session pathologies,
hidden oracles, and controlled memory events.  This makes memory failures
auditable, but it may underrepresent production repositories with larger
dependency graphs, ambiguous user intent, uncontrolled external services, and
open-ended issue resolution.  A future SWE-bench or SWE-bench Verified extension
would answer a different transfer question and should be reported separately.

\paragraph{Filesystem isolation, not network isolation.}
The isolation guarantee is filesystem-scoped: scored artifacts, reference
solutions, sequence files, and hidden oracles are absent from the relevant
wake-agent filesystems.  The evaluation is not network-isolated because hosted
model calls require network access.  The reported guarantee therefore rests on
container/filesystem boundaries and scanner/canary checks, not on a hermetic
air-gapped environment.

\paragraph{Hygiene labels are diagnostic.}
Executable S3 pass/fail is the primary outcome.  Hygiene metrics require
derived labels about usefulness, scope, stale use, repeated error, and
irrelevant import.  They are deterministic offline re-scores from result
records, but several are task-coupled or collinear, and the v2 fold does not
include a second full wake-model/judge transfer.  The hygiene panel should be
read as a diagnosis of failure modes, not as an independent proof that the
maintenance pipeline improves task success.

\paragraph{Baselines are bounded implementations.}
B5 is a deterministic verbatim event-memory substitute, not stock or tuned Mem0.
B5-MEM0 and B5-MEM0-LIT are two pinned configurations of one hosted Mem0
system family with specific package, service, namespace, and retrieval settings;
the latter is a diagnostic variant for the
historical exact-token failure.  Their low v2 rates do not imply a claim against
tuned Mem0 or Mem0$^g$, and the paper does not rank practical hosted-memory
systems.

For the successor audit, this historical two-configuration description must not be read
as two evaluated successor rows.  Only B5-MEM0-LIT cleared the frozen conformance gates.
Native B5-MEM0 failed the normalized-context identity gate before evaluation, and both
Supermemory conditions exhausted their repair budgets.  Thus no successor performance
result exists for those unavailable conditions.

\paragraph{External-system and benchmark-neighbor coverage.}
The original v2 fold includes one external memory-system family through two pinned
diagnostics; the successor evaluates only one admitted configuration from that family.
Its registered mechanism family is wholly unavailable.  The secondary
B5-MEM0-LIT-versus-B5 signal is not robust to trap-majority collapse, and the comparison
with DF-hybrid does not reject.  The audit therefore supports benchmark discrimination
and a bounded profile, not mechanism, superiority, non-inferiority, equivalence, or broad
ecological validity across deployed memory products.  MemoryArena, EvoMemBench,
WorldMemArena, MemoryAgentBench, and STATE-Bench also cover multi-session,
execution-oriented, lifecycle, or stateful-learning settings.  DreamBench-SWE's
claim is limited to hidden-evidence repository continuation with executable
software oracles and SWE-specific hygiene traps, not generic priority over those
benchmarks.

\paragraph{Successor historical and service scope.}
The v2.1 audit was prospectively frozen for its own outcomes but designed after the
original v2 study; it is additive and cannot retroactively enter the original P1--P6
family.  It uses one hosted wake-model route, one pinned Mem0 client/service
configuration, exact namespace filtering, a six-item retrieval cap, a 1,200-token read
budget, and controlled fixture repositories.  Hosted model and service behavior can
drift.  The frozen temporal bands are diagnostics rather than equivalence margins, and
the result does not generalize to tuned Mem0, Mem0$^g$, self-hosted deployments, other
retrieval policies, future service versions, production repositories, or other wake
models.

\paragraph{Strict-hybrid configuration.}
The strict-hybrid variant reaches \(78/180=0.433\), below the pre-registered
0.60 threshold.  Its strict admission settings (\(\theta=1.5\), read limit 3,
and 600-token read budget) are too conservative for the v2 trap distribution.
This is a failed tuning variant, not a validated strict retrieval design.

\paragraph{Cost is a constraint, not a headline.}
The v2 cost analyzer does not trip the pre-registered \(10\times\) cost
falsifier for \texttt{DF-hybrid} against B5, but the primary task comparison is
still null.  Estimated cost per successful task is useful for stress-testing
future memory policies and benchmark scale, yet it does not license favorable
cost, production, or resource claims.

\paragraph{Authoring and construct-history risk.}
The v2 trap set was designed after v1, the synthesis pilot, and mock-review
feedback.  The preregistration controls that history through construct quotas,
CSPRNG secret injection, admission gates, template caps, frozen sequence hashes,
and a published null-handling rule.  Those controls reduce but do not eliminate
authoring bias.  In particular, because the full second-backbone interaction is
scoped out and not reported, the paper avoids construction-independent or
model-general claims.

\paragraph{Terminology.}
The terms \emph{sleep} and \emph{dream} denote offline computation, replay, and
maintenance artifacts.  The paper does not claim biological equivalence,
consciousness, or cognitive fidelity.

%% file: sections/10_conclusion.tex
\section{Conclusion}

This paper reframes long-horizon software-agent memory as a maintenance and
measurement problem.  The primary contribution is \bench{}, a multi-session
software-engineering benchmark for memory hygiene: controlled repositories,
hidden executable oracles, frozen sequence records, CSPRNG-injected
non-inferable facts, validity gates, and trap-clustered inference.  The evaluated
maintenance pipeline is the reference probe used to stress that benchmark,
combining raw trajectory retention with typed, provenance-linked memory maintenance.

The v2 confirmatory fold is complete and gives a bounded negative result.  The
fold admits and evaluates 60 traps over three seeds, producing 180 S3 cells per
complete condition and \(1890/1890\) condition-level result files.  The primary
comparison does not reject: \texttt{DF-hybrid} reaches \(95/180=0.528\) versus
B5's \(89/180=0.494\), with signed statistic \(+6\), permutation
\(p=0.5184192657470703\), Holm-adjusted \(p=1.0\), and no rejection.  P2--P6
also do not reject.  The correct interpretation is not superiority and not
equivalence; it is that the scaled fold does not establish a system advantage
over strong verbatim event memory.

The benchmark result is still useful.  The executed fold yields descriptive
condition differences in the valid C2, C3, C5, C6, and C7 strata, while also
showing where the measurement is weak.  In particular, C9 fails stratum validity
because no-memory B0 passes all C9 cells, and C10 also lacks B0 headroom; neither
stratum supports spurious-lesson-rejection or abstention claims in this fold.
Those caveats narrow the benchmark claim rather than invalidate the full artifact.

The separately preregistered v2.1 successor audit adds a positive benchmark-discrimination
result without rewriting that original conclusion.  Across the same 60 traps and three
seeds, B0 achieved $\vOneBZeroPassed/\vOneBZeroValid$, B5
$\vOneBFivePassed/\vOneBFiveValid$, DF-hybrid
$\vOneDFHybridPassed/\vOneDFHybridValid$, and the admitted B5-MEM0-LIT configuration
$\vOneMemZeroLitPassed/\vOneMemZeroLitValid$.  All three available Family A
comparisons against B0 reject after fixed-six-slot Holm correction.  Both registered
Family B mechanism contrasts are unavailable after pre-evaluation conformance rejection.
The successor therefore supports \bench{} as a discriminating executable profile
benchmark and characterizes one exact hosted configuration; it does not establish
superiority over B5 or DF-hybrid, a causal mechanism, equivalence, or broad product
generality.

The resulting paper is therefore a benchmark-first result: \bench{} provides a
reproducible way to measure multi-session SWE memory hygiene, and the probe shows
how a richer memory-maintenance pipeline behaves under that protocol with one wake
model family and one seed schedule.  The current evidence does not justify broad
model-generality, generic benchmark-priority, tuned-Mem0 ranking,
production-reliability, or favorable cost claims.  It does justify a
transparent benchmark release with a null primary system result, explicit
construct caveats, an additive external-system profile, and canonical artifacts for
future memory systems to beat.

%% file: sections/appendix_verbatim.tex
\section{Recall-Verbatim Trap Audit}
\label{app:verbatim}

This appendix records the per-trap basis for the paper's statement that, under
its evidence-location taxonomy, 18 of 22 v1.0 S3 traps are recall-verbatim.  The
trap universe and reference-probe typed-only (DF)/B5 outcome counts come from
\path{analysis/investigation-evidence/per_trap_matrix_and_leakage.json}.
The semantic labels follow the Q-BENCH audit saved under
\path{analysis/investigation-evidence/REPORT-Q-BENCH.md} and agree with the
pre-registration disclosure that 18 of 22 traps are verbatim-event traps.  We
call a trap ``recall-verbatim'' when the required non-inferable token, format,
or rule is present verbatim in prior injected event text that B5 stores.  This
does not mean that answer code or a complete patch can be copied: all 22 traps
require implementation, and a stricter semantic audit found zero pure
copy-and-stop traps.  The four non-RV traps are generated-file/application
traps: their needed token is only partial without the generator workflow, so
they are listed as synthesis/apply rather than strict recall-verbatim.

\begin{table}[!htbp]
\centering
\small
\resizebox{\textwidth}{!}{%
\begin{tabular}{lllrr}
\toprule
Sequence & Class & Required S3 dependency & B5 S3 & reference-probe typed-only S3 \\
\midrule
config-convention-sections & Recall-verbatim & Section-list record and a redacted schema marker appear in event text & 3/3 & 3/3 \\
config-freeze-provenance & Recall-verbatim & \texttt{freeze\_tag()} must return the redacted freeze tag & 3/3 & 2/3 \\
config-reviewer-coerce-tag & Recall-verbatim & Bad-bool rejection format with a redacted reviewer code is stated in event text & 0/3 & 1/3 \\
config-reviewer-dupkey & Recall-verbatim & Duplicate-key line-numbered message format is stated in event text & 3/3 & 3/3 \\
config-reviewer-strict-csv & Recall-verbatim & Strict CSV width message format is stated in event text & 3/3 & 3/3 \\
config-stale-merge & Recall-verbatim & Redacted deep-merge source token is stated in event text & 3/3 & 2/3 \\
config-stale-schema-id & Recall-verbatim & \texttt{schema\_id()} must return the redacted schema id & 3/3 & 2/3 \\
expr-arity-contract & Recall-verbatim & \texttt{EXPR\_ARITY op=... expected=... got=...} format is stated in event text & 3/3 & 2/3 \\
expr-convention-opnaming & Synthesis/apply & Operator-family token must be applied through the generated-help workflow & 0/3 & 0/3 \\
expr-floordiv-category & Synthesis/apply & \texttt{quotient} category must be applied through the generated-help workflow & 0/3 & 3/3 \\
expr-generated-category & Synthesis/apply & \texttt{remainder} category must be applied through the generated-help workflow & 0/3 & 0/3 \\
expr-generated-precgroup & Synthesis/apply & Redacted precedence group must be applied through the generated-help workflow & 0/3 & 1/3 \\
expr-reviewer-divzero & Recall-verbatim & Redacted divide-by-zero message and signature are stated in event text & 3/3 & 1/3 \\
expr-stale-registry-stability & Recall-verbatim & Redacted registry stability token is stated in event text & 0/3 & 0/3 \\
todo-archive-bucket & Recall-verbatim & Redacted archive bucket token is stated in event text & 3/3 & 3/3 \\
todo-convention-aggregate & Recall-verbatim & \texttt{STATS} output contract and redacted batch marker are stated in event text & 3/3 & 3/3 \\
todo-convention-summary & Recall-verbatim & \texttt{TOP} summary contract and redacted tier token are stated in event text & 3/3 & 2/3 \\
todo-dedupe-keeplowest & Recall-verbatim & Keep-lowest-id/casefold and monotonic-\texttt{next\_id} rules are stated verbatim across prior events & 3/3 & 1/3 \\
todo-flaky-duetiebreak & Recall-verbatim & Due-date tie-breaker comparator is stated in event text & 3/3 & 3/3 \\
todo-flaky-monotonic-ids & Recall-verbatim & \texttt{purge-done} must preserve monotonic \texttt{next\_id} & 3/3 & 3/3 \\
todo-reviewer-export-dialect & Recall-verbatim & Redacted export dialect marker and CSV dialect are stated in event text & 3/3 & 3/3 \\
todo-reviewer-idformat & Recall-verbatim & Redacted review-only id format is stated in event text & 3/3 & 3/3 \\
\bottomrule
\end{tabular}
}%
\caption{Per-trap recall-verbatim audit for the 22 v1.0 S3 traps.  Outcome
columns count S3 successes across seeds 1, 2, and 3 from
\texttt{per\_trap\_matrix\_and\_leakage.json}.  The table has 18
recall-verbatim rows and four synthesis/apply rows, matching the paper's
18/22 evidence-location disclosure; the label does not imply that a solution
patch is available for copying.}
\label{tab:appendix-verbatim}
\end{table}

The reference-probe typed-only outcome cells are taxonomy-display counts across seeds 1--3, not the
validity-gated typed-only denominator.  One displayed non-success
(\path{expr-stale-registry-stability}, seed~3) is the reference-probe typed-only (DF)
\texttt{codex\_exec\_failed} validity exclusion, so the validity-gated reference-probe typed-only denominator
remains $44/65$, not $44/66$.

Two qualifications matter for interpretation.  First, the matrix's
token-overlap fields are a crude code-diff signal and are not the source of the
semantic classification; the classification question is whether the prior injected
event text contains the S3 contract that the hidden oracle later checks.
Second, B5 success is not automatic even on every recall-verbatim row:
\path{config-reviewer-coerce-tag} and
\path{expr-stale-registry-stability} still require enough implementation
work that both systems can fail despite the verbatim contract being available.

%% file: sections/appendix_artifact.tex
\section{Artifact and Reproducibility Details}
\label{app:artifact}

This appendix pins the artifact-level details used to reproduce the reported v2
fold summaries.  The numeric source of truth for the current paper is the
canonical analyzer bundle under \path{analysis/fold/}, not hand-copied tables.
The run stamp is \texttt{20260706T074759Z}.  The frozen sequence file and
raw-evidence repair preregistration have the following SHA-256 digests:

\begin{CodeBlock}
experiments/env/sequences_confirmatory_v2.jsonl
  4966bad1e535aaa0165c8aa7f2cb6fefd1d4e4afca78accfd868591a40448743
analysis/investigation-evidence/PREREGISTRATION.md
  e1819a8c930a3595727e27d11c4e50349c6dcc0059a3473cb6966996f412ccaf
\end{CodeBlock}

\paragraph{Recorded v2 inputs.}
The v2 fold evaluates the frozen 60-trap sequence file over seeds
\(\{1,2,3\}\).  The primary wake model is \texttt{gpt-5.5}; the sleep/judge
model identifier is \texttt{codex-gpt-5.5}.  Complete conditions contribute
180 S3 cells each.  B5-MEM0 and B5-MEM0-LIT are supplemental hosted-Mem0
failure-analysis rows and are excluded from the P1--P6 family.

\begin{CodeBlock}
PYTHONPATH=src python3 scripts/run_grid.py \
  --sequence-records experiments/env/sequences_confirmatory_v2.jsonl \
  --seeds 1,2,3 \
  --judge-model codex-gpt-5.5 \
  --run-stamp 20260706T074759Z
\end{CodeBlock}

The public package does not require live hosted-model reruns to verify the
reported paper numbers.  It ships the folded analyzer outputs and the scripts
used to regenerate the paper-facing summaries from the local result mirror.

\paragraph{Canonical analyzer bundle.}
The post-fold analyzer bundle consists of:
\begin{itemize}
\item \path{analysis/fold/v2_fold.json}
\item \path{analysis/fold/v2_confirmatory_clustered.json}
\item \path{analysis/fold/v2_confirmatory_clustered.stdout.json}
\item \path{analysis/fold/v2_hygiene_oracle.json}
\item \path{analysis/fold/v2_hygiene_oracle.tex}
\item \path{analysis/fold/v2_cost_frontier.json}
\item \path{analysis/fold/v2_cost_frontier.tex}
\item \path{analysis/fold/admission_funnel.json}
\item \path{analysis/fold/admission_funnel.md}
\item \path{analysis/fold/admission_funnel.tex}
\item \path{analysis/fold/v2_tables.tex}
\item \path{paper/figures/v2_construct_coverage.pdf},
  \path{paper/figures/v2_ladder.pdf}, and
  \path{paper/figures/v2_verbatim_vs_synthesis.pdf}
\end{itemize}

The completion and freshness gates are also file-backed:
\path{analysis/fold/v2_post_analyzer_completion_check.txt} ends with
\texttt{STATUS PASS failures=0}; the final public package freshness audit ends
with \texttt{STATUS PASS failures=0}.

\paragraph{Artifact/data availability and token handling.}
The public artifact for release \texttt{v2.0.5} is distributed at
\url{https://github.com/iroiro147/dreambench-swe/releases/tag/v2.0.5}.  Its frozen
digest is:

\begin{CodeBlock}
dreambench-swe-artifact.tar.gz
SHA-256: c7b5803d6c30f0b146ce91884b5609d6969c055b4f0160e6de1d0e7adca6eb23
\end{CodeBlock}

This digest binds the frozen public artifact tarball.  The tarball does not
contain this appendix, either manuscript PDF, the arXiv source archive, or the
detached release checksum ledger.  The release's
\path{RELEASE-CHECKSUMS.sha256} binds all release assets; PDF and arXiv-source
hashes are recorded only there because those assets contain this statement.
The \texttt{v2.0.5} tag and its assets are immutable historical evidence.  Any
manuscript correction or separately preregistered external-system audit must use
a distinct successor version and must not retag or replace this release.

The public package is built by \path{scripts/package_artifact.py} and consists of sequence
records, public fixtures, harness code, focused
analysis/smoke scripts, the container recipe, tests that do not require hidden
scoring assets, and frozen folded result summaries under the repository license.
It excludes \path{.git}, local agent instructions/state files, logs, credentials,
handoff material, full raw hosted-model result directories, and hidden scoring
assets under \path{experiments/env/oracles/} and \path{experiments/env/refsol/}.
The checksum table below reports the pre-packaging checkout artifacts used by
the manuscript; the public tarball manifest records package-internal hashes
after reviewer-only paths and local identifiers have been scrubbed.
A reviewer-only package built with
\filepath{scripts/package_artifact.py --private} additionally includes the hidden
oracle/reference-solution assets, still scrubbed of local paths and secrets.
Hosted service credentials are never included.

\paragraph{Pretraining and local-leakage stance.}
The S3 contracts depend on fresh arbitrary tokens and repository-local rules, so
the base model should not infer them from pretraining unless the benchmark
records themselves have leaked into training data.  Filesystem isolation
addresses a different threat: hidden oracles, reference solutions, sequence
records, analysis files, logs, and the paper repository are outside the
wake-agent container mount set.  The guarantee is therefore fresh-token
protection against ordinary pretraining inference plus container-filesystem
isolation against local leakage, not network isolation.

\paragraph{Isolation proof.}
The isolation claim is filesystem-scoped: scored benchmark artifacts are absent
from the relevant container filesystems, but the run is not network-isolated.
The proof pointers are \path{analysis/fold/HERMETICITY-MANIFEST.md} for the
mount set and boundary definition, and \path{analysis/fold/CANARY-PROOF.txt} for
the live 3/3 canary acceptance run.

\paragraph{Adjacent evidence artifacts.}
Cost-frontier numbers come from \path{analysis/fold/v2_cost_frontier.json}.
The hosted-Mem0 mechanism is scoped to the pinned configuration and is discussed
in \path{analysis/investigation-evidence/MEM0-ROW-FOLD-REPORT.md}: audited
failures show relevant memory retrieval but byte changes to non-inferable
tokens.  Historical cross-model hygiene rescore files, where present, are
limitation diagnostics only and are not v2 transfer-fold evidence.

Tables~\ref{tab:artifact-checksums}--\ref{tab:artifact-df-definitions} list
checksums and condition definitions.  The relevant v2 artifact checksums in this
checkout are listed below.  Package-level checksums are written to
\path{dist/CHECKSUMS.sha256} by the packaging step; they are not embedded here
because the package manifest records its generation time on rebuild.  The
release asset checksum sheet is \path{dist/RELEASE-CHECKSUMS.sha256}.

\begin{table}[htbp]
\centering
\scriptsize
\resizebox{\textwidth}{!}{%
\begin{tabular}{ll}
\toprule
Artifact & SHA-256 \\
\midrule
\path{analysis/fold/v2_fold.json} & \texttt{bfd63c8090c8b44650d56b582dac207faa99210bf25f66a911fe24c1d7f6f491} \\
\path{analysis/fold/v2_confirmatory_clustered.json} & \texttt{2294352db444fdcb0ee367f7c8b2a3682848cd66311f35ba04fb346ca4a71425} \\
\path{analysis/fold/v2_hygiene_oracle.json} & \texttt{77cd239c9d4097535d511aa6bc89adb0ec06f29c844cb6e4bf3b390c7740358f} \\
\path{analysis/fold/v2_cost_frontier.json} & \texttt{60e82b31126930642141c042f20052108147e4ab700d01bfdd77f4c4743b7662} \\
\path{analysis/fold/admission_funnel.json} & \texttt{229375cea802c0dcbc2576397ba73023b6e42c2fc8f59a616ef958904552fb52} \\
\path{analysis/fold/v2_tables.tex} & \texttt{2c1118dac674fbbb6d09875eb209cf621278cb8851b929901517cf52485f0b9a} \\
\path{paper/figures/v2_construct_coverage.pdf} & \texttt{9414a422cc390fd10029e4a186f1482d2129f1ead4ba821fff4aa1308203d54f} \\
\path{paper/figures/v2_ladder.pdf} & \texttt{7712cb8e258b492e15d4d2b8c8909ffa607004543dda1a7e96afb2044e290699} \\
\path{paper/figures/v2_verbatim_vs_synthesis.pdf} & \texttt{86573fc7f8b41bf50f91f377747db1a96226a4d1540441eb41da3e03de3e3a78} \\
\bottomrule
\end{tabular}
}%
\caption{Checksums for the v2 analyzer bundle artifacts in this checkout.}
\label{tab:artifact-checksums}
\end{table}

\paragraph{Condition definitions.}
The condition definitions below are the code-level definitions used by the v2
fold.

\begin{table}[htbp]
\centering
\small
\begin{tabular}{p{0.18\linewidth}p{0.74\linewidth}}
\toprule
Condition & Definition \\
\midrule
B0 & No external memory reads or writes. \\
B1 & Raw prior trajectory retrieval only. \\
B2 & Deterministic token-vector retrieval over prior trace chunks. \\
B3 & Reflection-only vector retrieval: free-form verbal reflection lessons after sessions, retrieved by vector similarity. \\
B4 & Deterministic untyped session summaries. \\
B5 & Offline instance-memory substitute; one deterministic verbatim injected-event memory per trajectory. \\
B6 & Subtask-level memory over trajectory actions. \\
B7 & Sequence-local task-tracker memory. \\
B5-MEM0 & Supplemental hosted Mem0 condition, client \texttt{mem0ai} 2.0.11, hosted platform, default extraction settings, and \texttt{memory\_context=6}. \\
B5-MEM0-LIT & Supplemental hosted Mem0 diagnostic variant for exact-literal handling. \\
\bottomrule
\end{tabular}
\caption{Baseline condition definitions.}
\label{tab:artifact-baselines}
\end{table}

\begin{table}[htbp]
\centering
\small
\begin{tabular}{p{0.18\linewidth}p{0.74\linewidth}}
\toprule
Condition & Definition \\
\midrule
DF & Typed consolidation, contradiction repair, counterfactual replay, local maintenance, and retrieval gating. \\
DF-raw-only & Typed consolidation disabled; raw-evidence capsules enabled; contradiction records excluded from implementation reads. \\
DF-hybrid & DF plus raw-evidence capsules; contradiction records excluded from implementation reads. \\
DF-strict & DF with \(\theta_{\mathrm{admit}}=1.5\), read limit 3, and read-token budget 600. \\
DF-strict-hybrid & DF-hybrid with the same strict retrieval settings as DF-strict. \\
A0 & Episodic-only raw episode writes; no typed consolidation, repair, or replay. \\
A2 & Contradiction repair disabled. \\
A4 & Counterfactual replay disabled. \\
A5 & Read-side stale and superseded suppression disabled while repair remains enabled. \\
A6 & Retrieval hard gate disabled; memory admission is score-ranked. \\
A11 & Global maintenance scope forced during consolidation. \\
\bottomrule
\end{tabular}
\caption{Reference-probe ladder and ablation definitions.}
\label{tab:artifact-df-definitions}
\end{table}

Designed-but-not-run ablations in this fold are A1 no typed consolidation, A3 no causal
failure extraction, A7 similarity-only retrieval, A8 no provenance gate, A10 no raw
episodic fallback at read time, A12 no scope fields, and A13 no validity intervals.

\subsection{v2.1 Successor Evidence Identity}
\label{app:v21-artifact}
The successor release is distinct from and additive to immutable v2.0.5.  Its canonical
public record and downloadable assets are at
\url{https://github.com/iroiro147/dreambench-swe/releases/tag/v2.1.0}.  The release
contains the sanitized successor artifact, paper PDF, flat arXiv source, manifest, and
detached checksum ledger; the manifest binds the evaluation, launcher, and release-source
commits without altering the historical \texttt{v2.0.5} tag or assets.  Its canonical
run stamp is \path{20260716T230000Z}; the implementation commit is
\path{ea86d4c45b5a1d47896b54aac7f7518985737056}, and the launcher head is
\path{9e0c949a07e6b783d00b421606c5ce3cf84e5a67}.  The lock SHA-256 is
\path{877412e82a1aa70c358f76f8cd3ab93439b89f3e6ac8ccc61f8ca6094819f082};
the canonical analyzer JSON file SHA-256 is
\path{288273980c26e9f97630be69443ea14e72bcbccae63889c10f02d4da378c97d1},
while the canonicalized JSON SHA-256 is
\path{94bb7d82962f8b1740e9579ae0574adc5571ecba9438622f7075a0e11cb9adcb}.
The deterministic surface-generation receipt binds both digests explicitly, and
\path{surfaces/external_audit_v21_artifacts.sha256} closes the generated surface set.
The manuscript consumes the generated macro surface through
\path{analysis/fold/v21_external_audit_macros.tex}; a fail-closed numeric-provenance
check requires both manuscript trees to carry exact copies of the generated macro and
table surfaces.  Private input ledgers, raw result/log/receipt roots, credentials, hidden
oracles, and local operational paths are excluded from the public package.

%% file: sections/appendix_v2_protocol.tex
\section{V2 Benchmark Protocol Preregistration}
\label{app:v2-protocol}

This appendix records the frozen V2 protocol for the scaled \bench{}
confirmatory fold.  The source of truth is
\filepath{analysis/investigation-evidence/PREREGISTRATION-V2.md} as reconciled by
\filepath{analysis/investigation-evidence/SCALE1-FREEZE-DECISIONS.md}.  The protocol
was frozen on 2026-07-04 before V2 trap authoring or any V2 performance run; any
later change requires a disclosed amendment log.  The v2 fold has now run; the
fields below record the resolved admission, analysis, falsifier, and artifact
status from the canonical v2 analyzer bundle.  The planned second wake-model
transfer did not proceed to a full fold, so transfer claims are explicitly scoped
out rather than filled with a post-hoc substitute.

\paragraph{Scope.}
V2 is a new scaled confirmatory fold for \bench{} and does not rewrite the V1
fold.  It fixes the trap set target, authoring controls, validity gates,
condition matrix, primary analysis, falsifiers, second wake-model transfer fold,
and go/no-go criteria before any memory-bearing V2 outcome is inspected.  The
transfer fold is now scoped out for the reported paper.

\paragraph{Definitions.}
A \emph{trap} is one multi-session sequence whose S3 task is the scored memory
trap.  In the primary analysis, a trap is the statistical cluster.  A \emph{cell}
is one \((\mathrm{trap}, \mathrm{seed}, \mathrm{condition})\) S3 outcome.  The
seed schedule is seeds \(1,2,3\).  The primary outcome is validity-gated S3
executable pass/fail.  The primary paired unit is the trap cluster; the three
seeds are repeated observations within a trap, not independent benchmark items.
The primary wake-model fold uses \texttt{gpt-5.5}.  The second wake-model fold
is a separate transfer check and is never pooled into the primary fold.

\paragraph{Hard authoring rule.}
No new V2 trap may be authored after looking at any V2 performance result.  The
only result-like information allowed before final trap freeze is the
pre-declared validity-gate verdict for candidate admission: B0 S3 failure,
reference-solution pass, oracle hiddenness, CSPRNG-secret compliance,
no-single-event or construct-specific gate status, clean-start flag, and
manifest/schema checks.  Once any memory-bearing condition has produced an S3
result on any V2 candidate trap, trap authoring is permanently closed for this
fold.

\subsection{Causal History and Bias Controls}
\label{app:v2-causal-history}

The V2 design is informed by the V1 fold, mock reviews, the 8-trap synthesis
pilot, and the scale-freeze review.  The report using V2 must disclose that
history.  The known V1 facts before V2 freeze include: 22 traps run on three
seeds for 66 S3 cells per complete condition; 18 of 22 V1 traps classified as
recall-verbatim and four as synthesis/apply; reference-probe hybrid leading B5
numerically but not significantly on the primary V1 comparison
\((52/66=0.788\) versus \(48/66=0.727\), exact McNemar
\(b=12,c=8,p=0.503\), Holm \(p=1.0)\); and the typed-only-versus-B3 positive
result surviving the trap-clustered exact test but not all conservative
sensitivity variants.  The corrected synthesis-pilot denominator is 24 S3 cells
\((8\) traps \(\times\) \(3\) seeds), and that pilot is cited only as a
pre-registered directional slice, never pooled into V2.

Those facts create authoring risks: authors could write traps to punish B5's
verbatim strategy, emphasize constructs where the reference probe looked strong, avoid V1
failure modes, tune hidden literals to condition read limits, or remove traps
after seeing early outcomes.  V2 controls those risks by allocating traps by
construct quota, recording construct and family labels separately, freezing
candidate hashes and validation reports before performance runs, using
CSPRNG-injected secrets, enforcing a template cap, disclosing the admission
funnel, and publishing the primary clustered result even if it is null.

\subsection{Trap Set and Construct Quotas}
\label{app:v2-trap-set}

The V2 target is 60 valid S3 traps, each run on three seeds, for 180 S3 cells per
complete condition:
\[
60 \ \mathrm{traps} \times 3 \ \mathrm{seeds} = 180 \ \mathrm{S3\ cells}.
\]
The target composition is 22 V1 traps, 8 synthesis-pilot traps, and
approximately 30 new traps.  Sixty valid traps is the confirmatory target because
the primary test clusters by trap; adding seeds without adding traps would not
solve the pseudo-replication concern.

The governing allocation unit is construct, not legacy family.  Every trap
receives one construct label \((\mathrm{C1}\) through \(\mathrm{C10})\) and one
separate family label.  Family labels support descriptive reporting but cannot
override construct quotas.

\begin{table}[!htbp]
\centering
\small
\resizebox{\textwidth}{!}{%
\begin{tabular}{llrp{0.39\linewidth}}
\toprule
Construct & Construct name & Target \(n\) & Role in V2 \\
\midrule
C1 & Verbatim retention & 18 & V1 continuity and B5 calibration anchor \\
C2 & Retrieval precision under interference & 6 & Anti-hoarding and executable precision under distractors \\
C3 & Staleness detection and supersession & 7 & Stale suppression and multi-hop supersession \\
C4 & Update propagation / regression avoidance & 1 & Derived update propagation probe \\
C5 & Scope discipline & 6 & Overgeneralization resistance \\
C6 & Contradiction handling / provenance conflict & 5 & Authority and precedence under conflict \\
C7 & Cross-session synthesis / paraphrase-retention application & 7 & Non-verbatim composition and paraphrase-retention control \\
C8 & Procedural / source-of-truth memory & 4 & Generated-source workflow and procedural memory \\
C9 & Spurious-lesson rejection / disconfirmation & 4 & Bad-memory rejection \\
C10 & Abstention / confabulation resistance & 2 & Irrelevant-memory rejection \\
\midrule
Total & & 60 & \\
\bottomrule
\end{tabular}
}%
\caption{Frozen V2 construct quota.  The arithmetic is
\(18+6+7+1+6+5+7+4+4+2=60\).}
\label{tab:v2-construct-quota}
\end{table}

For V2, the anti-hoarding constructs are C2, C3, C5, C6, C9, and C10, plus any
C7 trap explicitly labeled as paraphrase-retention where byte-copying is
insufficient.  The frozen target from C2, C3, C5, C6, C9, and C10 alone is
\[
6+7+6+5+4+2 = 30
\]
anti-hoarding traps, exceeding the binding floor of at least 20 anti-hoarding
traps.  C1 recall-verbatim has target 18, cap 24, and may not exceed 40\% of the
final valid set.

\paragraph{Construct definitions.}
C1 preserves a non-inferable exact token, format, or rule from one prior event.
C2 selects one relevant stored fact among distractors under a fixed read-token
budget.  C3 uses a superseding fact and suppresses the stale one, with at least
two C3 traps defeating a simple newest-event heuristic.  C4 propagates a changed
fact to a derived later use.  C5 applies feedback only inside its stated scope.
C6 resolves conflict by authority or precedence rather than raw recency.  C7
derives an answer from two or more events or applies a paraphrased retained
instruction so that no single event contains the decisive S3 answer as a
contiguous literal.  C8 remembers the repository's procedural source of truth.
C9 rejects an unsupported or later-disconfirmed lesson.  C10 abstains when no
prior event bears on the current task.

\paragraph{Manifest fields.}
The V2 manifest records, for every trap: sequence id, construct label, family
label, repository, template id, author model, skeleton id, anti-hoarding flag,
clean-start S3 flag, CSPRNG secret manifest hash, oracle hash, reference-solution
hash, and validation report path and hash.  The admitted/rejected manifest also
records first failing gate and whether a rejected candidate was discarded,
repaired before outcome inspection, or reserved before freeze.

\paragraph{Clean-start subset.}
At least 10 of the new traps must use clean-start S3:
\[
\geq 10 \ \text{new traps with } \mathtt{clean\_start\_s3}=\mathtt{true}.
\]
For clean-start traps, S3 does not inherit the S1/S2 production tree.  S3 starts
from the frozen clean base for the fixture repository while memory-bearing
conditions receive only admissible memory context from prior sessions.  The
analyzer must distinguish ordinary continuation traps from clean-start S3 traps
and must report the clean-start subset separately.

\subsection{CSPRNG Secret Injection and Validity Gates}
\label{app:v2-secrets-gates}

CSPRNG secret injection is a binding authoring rule.  The authoring LLM drafts
only a skeleton with named placeholders for non-inferable tokens, ids, marker
strings, format literals, and oracle literals.  The authoring LLM never invents a
secret token, never selects a random-looking literal, and never writes the final
hidden oracle literal by hand.  The harness generates all secret tokens with a
cryptographic random source and substitutes them into injected memory events,
hidden oracle expectations, reference-solution material, decisive-literal
metadata, and any public prompt field where the event itself is supposed to
reveal the token.  The manifest records a secret-manifest hash and
generator-code hash, but not secret values beyond those intentionally present in
public injected events.  The validator rejects any candidate whose decisive
non-inferable secret was authored directly by an LLM or a human rather than
injected by the harness.

Every non-C1 trap declares why verbatim replay is insufficient.  The automated
gate checks that no normalized decisive oracle literal is a substring of any
single injected event except for C1 or declared visible input literals that are
not themselves sufficient; that no single injected event covers all required
fact ids for non-C1 traps; that the S3 prompt does not reveal the decisive
answer; that public repository state before S3 does not contain the decisive
answer except in declared apply/derive cases; and that the trap is passable by
any system surfacing the needed facts rather than only by a probe-specific
operator.  ``Make B5 fail'' is not a valid design target.  B5 passing a non-C1
trap is an outcome, not an authoring defect.

Each new trap must pass the live validator before admission:
\begin{CodeBlock}
python3 experiments/validate_trap.py \
  --sequence-records <candidate-sequence-records> \
  --seq <seq_id> \
  --live
\end{CodeBlock}
The required verdict is: valid trap, B0 S3 failed, reference solution passes,
oracle hidden, CSPRNG secret injected, no public task leak of oracle commands or
hidden scoring material, construct-specific gate passes, and clean-start status
correctly represented in the manifest and validator.  For C7, the
no-single-event gate is required.  For C3, C5, C6, C9, and C10, the oracle must
check both correct behavior and absence of the construct-specific trap behavior.

\paragraph{Continuation and clean-start scoring.}
\begin{sloppypar}
Ordinary continuation traps must preserve the memory-trap invariant: S2 starts
from S1's scored production tree, and S3 starts from S2's scored production
tree.  Clean-start S3 traps instead start S3 from the frozen clean base by
design, while preserving valid S1 and S2 records as memory-source sessions.
Scoring applies only production diffs to a fresh scorer checkout and rejects
edits to tests, \texttt{.git}, import hooks, \texttt{conftest.py},
\texttt{pytest.ini}, \texttt{sitecustomize.py}, and \texttt{*.pth}.
\end{sloppypar}

\paragraph{Harness pre-fixes.}
Before any V2 performance run, \filepath{scan_completed_units} must verify full
per-sequence S1/S2/S3 coverage before skipping a unit, and analyzer warmup must
mean non-S3 sessions only.  For the now-scoped-out second wake-model fold, the launcher must
include wake model in work-unit keys and completed-unit detection so
\texttt{gpt-5.5} results cannot be mistaken for \texttt{glm-latest}, Kimi K2.7,
or frontier-peer results.

\subsection{Authoring, Admission, and Audit}
\label{app:v2-authoring-audit}

The v2 report discloses the new-trap admission funnel.  The canonical admission
funnel artifact is \filepath{analysis/fold/admission_funnel.md}; all 30 authored
new traps passed the recorded gates and were admitted:
\begin{CodeBlock}
C2:  6 authored ->  6 dry-valid ->  6 live-valid ->  6 admitted
C3:  5 authored ->  5 dry-valid ->  5 live-valid ->  5 admitted
C5:  6 authored ->  6 dry-valid ->  6 live-valid ->  6 admitted
C6:  4 authored ->  4 dry-valid ->  4 live-valid ->  4 admitted
C7:  3 authored ->  3 dry-valid ->  3 live-valid ->  3 admitted
C9:  4 authored ->  4 dry-valid ->  4 live-valid ->  4 admitted
C10: 2 authored ->  2 dry-valid ->  2 live-valid ->  2 admitted
Total: 30 authored -> 30 dry-valid -> 30 live-valid -> 30 admitted
\end{CodeBlock}
Together with the retained v1 and synthesis-pilot traps, the final v2 fold
evaluates 60 traps.

No more than four admitted traps may share the same trap skeleton:
\[
\leq 4 \ \text{admitted traps per } \mathtt{skeleton\_id}.
\]
A skeleton is defined by structural event pattern, oracle pattern, task type,
and fixture surface, not by token renaming or fixture-noun swaps.

Human audit is required before final trap freeze.  One hundred percent of
admitted traps receive human construct-label review.  At least 30\% of admitted
traps receive deep audit for construct fit, realism, leakage risk, template
duplication, and whether the stated insufficiency mechanism is load-bearing.
Raw agreement and Cohen's kappa, where applicable, are audit diagnostics rather
than confirmatory outcomes.  Disagreements are resolved before freeze.  Traps
with unresolved construct-label disagreement may remain as benchmark traps but
are excluded from per-construct claims; no unresolved construct-label
disagreement is used to support a v2 per-construct claim.

Every trap records \filepath{author_model}.  The frozen plan would have analyzed
the author-model \(\times\) wake-model interaction across both backbones: pass
rates by author model, wake model, condition, and construct;
primary-versus-transfer wake performance stratified by author model; a
trap-clustered descriptive interaction estimate for same-family author/solver
advantage; and, if enough traps existed, a sensitivity table excluding traps
authored by the same model family as the primary wake model.  Because the second
wake-model fold was scoped out, the reported paper does not present this
cross-backbone interaction or use it for a construction-independent validity
claim.  Such an analysis would be secondary and could not rescue or replace the
primary clustered comparison.

\subsection{Conditions and Run Freeze}
\label{app:v2-conditions}

The primary fold uses Codex CLI with \texttt{gpt-5.5} as the wake agent/model and
\texttt{codex-gpt-5.5} as the sleep/judge model wherever a judge is used.  The
trap set is the same frozen 60 valid traps, seeds are \(1,2,3\), task order is
fixed across conditions, budgets and stopping criteria are fixed, and each
condition receives an isolated memory namespace by run id, condition, seed, and
sequence.  Hidden oracles, reference solutions, sequence records, analysis
files, logs, and secret manifests are absent from the wake-agent filesystem.

The primary condition matrix is:
\begin{CodeBlock}
B0
B1
B2
B3
B4
B5
B6
B7
B5-MEM0
B5-MEM0-LIT
DF
DF-raw-only
DF-hybrid
DF-strict
DF-strict-hybrid
A0
A2
A4
A5
A6
A11
\end{CodeBlock}
\texttt{B5} is the deterministic verbatim event-memory / offline
instance-memory baseline, not stock or tuned Mem0.  \texttt{B5-MEM0} and
\texttt{B5-MEM0-LIT} are two pinned configurations of one hosted Mem0 family,
used only when package, service, namespace isolation, payload format, and key
are available.  They are supplemental and do not enter the primary Holm family.

Before the first primary-fold run, the following inputs are frozen and hashed:
this preregistration, sequence manifest, construct and family labels, condition
definitions, model ids, provider endpoints, decoding parameters, context limits,
budgets, seeds, judge prompts, calibration examples, scorer code, CSPRNG
generator code, and secret manifests.  Result roots, grid logs, Mem0 namespaces,
and Mem0 fixture/cache artifacts are isolated with
\filepath{DREAMBENCH_RESULTS_ROOT}, \filepath{DREAMBENCH_LOG_ROOT},
\filepath{DREAMBENCH_NAMESPACE_PREFIX}, and
\filepath{DREAMBENCH_MEM0_FIXTURE_ROOT}.

\subsection{Primary Analysis}
\label{app:v2-primary-analysis}

The primary outcome is validity-gated S3 executable pass/fail for each
\((\mathrm{condition}, \mathrm{seed}, \mathrm{trap})\) cell.  Warmup rates,
hygiene metrics, cost, construct slices, family slices, and authoring diagnostics
are reported but do not replace the primary S3 analysis.

The primary inferential test is an exact trap-clustered sign/permutation test.
For comparison \(A\) versus \(B\) and trap \(t\), define
\[
d_t =
\#\{\mathrm{paired\ valid\ seeds}: A \ \mathrm{passes}, B \ \mathrm{fails}\}
-
\#\{\mathrm{paired\ valid\ seeds}: A \ \mathrm{fails}, B \ \mathrm{passes}\}.
\]
Tied paired seed cells contribute zero, and
\[
D_{\mathrm{obs}}=\sum_t d_t .
\]
Under the null, the sign of each nonzero trap contribution is exchangeable.  The
two-sided exact p-value is
\[
p =
\Pr\left(\left|\sum_t s_t |d_t|\right| \geq |D_{\mathrm{obs}}|\right),
\qquad s_t \in \{-1,+1\}.
\]
Zero-contribution traps remain in descriptive counts but do not increase the
number of sign-flip states.  The implementation may enumerate signs directly or
use exact dynamic programming; no normal approximation is used for the primary
p-value.

\begin{table}[!htbp]
\centering
\small
\begin{tabular}{lll}
\toprule
ID & Comparison & Purpose \\
\midrule
P1 & \texttt{DF-hybrid} vs.\ \texttt{B5} & Headline comparison against strongest V1 baseline \\
P2 & \texttt{DF-typed-only} vs.\ \texttt{B5} & Typed consolidation alone against verbatim memory \\
P3 & \texttt{DF-raw-only} vs.\ \texttt{B5} & Raw evidence retention against B5 \\
P4 & \texttt{DF-hybrid} vs.\ \texttt{DF-raw-only} & Incremental value of typed maintenance above raw evidence \\
P5 & \texttt{DF-hybrid} vs.\ \texttt{DF-typed-only} & Incremental value of raw evidence above typed-only \\
P6 & \texttt{DF-typed-only} vs.\ \texttt{B3} & Structured typed memory against reflection/vector anchor \\
\bottomrule
\end{tabular}
\caption{Frozen V2 primary S3 comparison family.}
\label{tab:v2-primary-comparisons}
\end{table}

All six primary p-values are two-sided exact clustered p-values.
Holm-Bonferroni correction controls family-wise error at \(\alpha=0.05\) across
P1--P6.  The headline claim \texttt{DF-hybrid} \(>\) \texttt{B5} requires P1 to
have positive \(D_{\mathrm{obs}}\) and Holm-adjusted clustered \(p<0.05\).  If
P1 is null, the paper must state that the scaled fold does not establish
\texttt{DF-hybrid} \(>\) \texttt{B5}, even if pooled McNemar, construct slices,
family slices, hygiene metrics, or transfer analyses are significant.

No imputation is allowed.  A paired seed cell is included only if both
conditions have validity-gated S3 outcomes for that \((\mathrm{trap},
\mathrm{seed})\).  P1 requires at least 54 traps with at least two paired valid
seeds.  P1 cell-level validity exclusions must be at most 5\% for each side; with
180 cells this is at most nine excluded cells per side.  If exclusions exceed
5\% for either side, P1 may be reported only as incomplete.

Pooled exact McNemar is secondary and uses the seed-trap cell as the paired unit.
It is computed for P1--P6 with the same two-sided alternative and Holm
correction, but it can support only sensitivity language and cannot overturn a
null clustered P1.  Required sensitivity reports include per-seed exact McNemar,
trap-majority collapse, descriptive construct slices, descriptive family slices,
and the author-model \(\times\) wake-model interaction table.  Per-construct
strata are descriptive because \(n\) is small within strata.

\paragraph{Power interpretation.}
The primary clustered test depends on nonzero trap clusters, not merely on 180
pooled cells.  Under an equal-weight sign-test approximation, if all 60 trap
clusters are nonzero, a 38--22 split gives \(p=0.051894\) and a 39--21 split
gives \(p=0.027340\).  With one discordant seed per trap, 39--21 corresponds to
a net margin of \(18/180=0.100\).  With all three seeds discordant in the same
direction within each trap, it corresponds to \(54/180=0.300\).  Therefore, 60
traps is a reasonable target for large stable trap-level effects, not a
guarantee of detecting small or template-concentrated effects.  In the observed
v2 P1 vector, \(D_{\mathrm{obs}}=+6\), the exact trap-clustered permutation
p-value is \(0.5184192657470703\), and the Holm-adjusted p-value is \(1.0\).

\subsection{Second Wake-Model Fold (Scoped Out)}
\label{app:v2-second-backbone}

This scoped-out limitation record describes the frozen transfer plan.  The
second wake-model fold is a transfer check, not part of the primary confirmatory
sample.  The planned second backbone is GRID GLM-5.2, recorded in the harness
as \texttt{glm-latest} for this scoped-out limitation; it is treated as a pinned open-weights checkpoint as
exposed through GRID.  The judge remains \texttt{codex-gpt-5.5}.
Seeds, trap set, task order, budgets, and stopping criteria match the primary
fold.  The planned transfer conditions are:
\begin{CodeBlock}
B0
B3
B5
DF
DF-raw-only
DF-hybrid
\end{CodeBlock}
\texttt{B5-MEM0} may be included only if the hosted Mem0 configuration is
already working and frozen before the now-scoped-out transfer fold starts; it remains
supplemental.

Before the full transfer fold, GLM-5.2 must pass a 5-trap warmup pilot under the
same frozen harness and fixed judge.  If GLM-5.2 warmup pass rate is at least
0.60, the GLM-5.2 transfer fold proceeds.  If GLM-5.2 warmup pass rate is below
0.60, the floor is reported honestly and the fold falls back to Kimi K2.7.  If
Kimi K2.7 also floors under the same 5-trap pilot rule, the fold falls back to
one frontier peer and discloses the full fallback chain.  A floor result is
reported as second-backbone feasibility evidence, not hidden, pooled, or treated
as a failed implementation detail.  The GLM-5.2 warmup path did not clear the
feasibility gate, so no full second wake-model transfer fold is reported.  This
paper therefore makes no cross-backbone claim and no transfer claim.

Future work: a transfer fold may support or refute directional transfer to a non-\texttt{gpt-5.5}
wake model, identify wake-specific construct or family behavior, and test
whether B5 strength is a property of \texttt{gpt-5.5} in-context copying or of
the benchmark itself.  It cannot be pooled with the primary fold, rescue a null
primary P1, replace the primary Holm family, justify post-hoc trap removal, or
establish broad model generality beyond the one fixed second wake model and
seeds.

\subsection{Falsifiers, Go/No-Go, and Null Handling}
\label{app:v2-falsifiers}

Falsifiers constrain the causal story but do not authorize hiding the result.
If a falsifier fails, the V2 result still publishes with the corresponding claim
removed or narrowed.

\begin{table}[!htbp]
\centering
\scriptsize
\begin{tabular}{p{0.21\linewidth}p{0.49\linewidth}p{0.20\linewidth}}
\toprule
Falsifier & Frozen threshold or claim rule & V2 value \\
\midrule
B0 headroom & B0 S3 pass rate \(\leq 0.40\) preferred; \(\leq 0.50\) maximum acceptable. & Overall B0 \(21/180=0.117\) passes; C9 \(12/12\) and C10 \(6/6\) B0 headroom fail. \\
Useful memory precision & \texttt{DF-hybrid} \(\geq 0.45\) and no more than 0.15 below B5. & FAIL: \texttt{DF-hybrid}=0.264; B5=0.556. \\
Seed spread & \texttt{DF-hybrid} max(seed) minus min(seed) \(\leq 0.12\) preferred; \(\leq 0.16\) maximum for stability claim. & PASS: rates \(31/60,32/60,32/60\); spread \(0.017\). \\
Regression and repeated error & \texttt{DF-hybrid} repeated-error rate \(\leq\) B5 and \(\leq 0.15\); regression-after-update \(\leq 0.08\). & MIXED/FAIL: repeated error 0.142 passes relative to B5 0.225 and threshold 0.15; regression-after-update 0.106 fails. \\
Strict-hybrid recovery & If included, \texttt{DF-strict-hybrid} S3 pass rate \(\geq 0.60\) and no more than 0.12 below \texttt{DF-hybrid}. & FAIL: \(78/180=0.433\). \\
Construct dependence & General memory-hygiene claim requires nonnegative P1 direction on anti-hoarding aggregate and on C7. & No general memory-hygiene claim is made; primary P1 is null and F7 is false. \\
New construct floor & If three or more new construct strata have max-condition pass rate below 0.15, top-venue benchmark claim is killed. & No three-strata floor kill; benchmark claim is narrowed by C9/C10 B0-headroom failures. \\
Author \(\times\) solver contamination & Same-family author/solver advantage beyond wake-model difference requires re-authoring with separation before construction-independent validity claim. & Not used for a construction-independent claim; no full second-backbone fold is reported. \\
Second wake-model transfer & Negative transfer P1 direction forbids broad model-generalization language; ladder reordering reframes the paper as model-dependence evidence. & Scoped out: GLM-5.2 warmup path did not proceed to a full transfer fold; no model-generality claim. \\
Clean-start subset & If the primary direction holds only in inherited-state traps and disappears or reverses in clean-start S3, memory-quality isolation claim is forbidden. & Descriptive only: 13 traps; B5 \(15/39=0.385\), \texttt{DF-hybrid} \(19/39=0.487\), B3 \(22/39=0.564\). \\
Cost frontier & If \texttt{DF-hybrid} cost per successful S3 task is more than \(10\times\) B5, no favorable cost claim is allowed. & PASS as a falsifier only: \texttt{DF-hybrid} 0.03832 USD/success vs. B5 0.03953; no favorable cost claim is made. \\
\bottomrule
\end{tabular}
\caption{Frozen V2 falsifiers and claim constraints with resolved v2 values
from the canonical analyzer bundle.}
\label{tab:v2-falsifiers}
\end{table}

The primary fold may proceed only if the preregistration is reviewed, frozen,
and hashed; no trap was authored after viewing V2 performance results; 60 valid
traps or enough pre-freeze replacements are available; construct quota and
anti-hoarding floor are satisfied; C1 remains within cap; at least 10 new traps
are clean-start S3; every new trap uses CSPRNG secret injection; every new
non-C1 trap passes its construct gate; every new trap passes live validation; B0
headroom admission gate is satisfied; 100\% construct-label audit and at least
30\% deep audit are complete; template cap is satisfied; harness pre-fixes are
complete; and run manifests, condition definitions, model configs, seeds,
budgets, scorer hashes, and CSPRNG secret manifest hashes are frozen.

Abort the confirmatory fold, or downgrade it to pilot/descriptive, if fewer than
60 valid traps remain without a pre-freeze replacement path; construct quotas
cannot be satisfied; fewer than 20 anti-hoarding traps remain; C1 exceeds 24
traps or 40\%; fewer than 10 new clean-start traps remain; CSPRNG secret
injection is violated on any admitted trap; final B0 S3 pass rate exceeds 0.50;
P1 has fewer than 54 traps with at least two paired valid seeds; P1 exclusions
exceed 5\% for either \texttt{DF-hybrid} or B5; hidden oracle leakage is
discovered; one condition can read another condition's namespace; any trap is
edited or removed after outcome inspection; or either required harness pre-fix
is absent.

If P1 is null under the primary clustered Holm-corrected test, the null result is
published.  The paper must state that the scaled fold does not establish
\texttt{DF-hybrid} superiority over B5; it must not promote pooled McNemar,
construct slices, family slices, per-seed tests, scoped-out transfer-fold results, or
hygiene metrics into the headline.  If P1 is significant, the paper reports the
exact clustered p-value and Holm-adjusted p-value, reports pooled McNemar only
as secondary, applies all falsifier constraints before causal interpretation,
and avoids favorable cost or broad model-generality language unless the
corresponding falsifiers pass.

\paragraph{Binding kill criteria.}
The scale-freeze kill criteria are:
\begin{CodeBlock}
>=3 new construct strata floored (<0.15 max-condition) -> arXiv
measurement-study fallback; author x solver contamination -> re-author
w/ separation; ladder reordering across backbones (future-work limitation) -> model-dependence
analysis paper; instability at scale -> not top-venue. Never shrink
back to the verbatim stratum for a lesser venue.
\end{CodeBlock}
\texttt{DF-hybrid} failing to beat B5 is explicitly not a kill if the new
constructs discriminate among conditions.  A null or losing system result is
publishable as a benchmark finding if the benchmark measures what it claims to
measure.

\subsection{Required V2 Artifacts and Claim Boundaries}
\label{app:v2-artifacts-claims}

The scaled fold must preserve: frozen preregistration path and hash; frozen
sequence manifest and hash; construct/family manifest; CSPRNG generator-code
hash; CSPRNG secret-manifest hashes; hidden-oracle hash manifest; reference
solution hash manifest; validation reports for every new trap; no-single-event
or construct-specific reports for every non-C1 trap; clean-start subset
manifest; author-model, skeleton-id, and template-id manifest; human-audit
report with agreement; admission-funnel report; run manifest with models,
conditions, seeds, budgets, task order, and scorer hash; harness pre-fix evidence
for \filepath{scan_completed_units} and warmup definition; per-condition result
records; primary clustered analysis script output; secondary pooled McNemar
output; per-seed and trap-majority sensitivity output; construct and family
slice tables; clean-start subset analysis; author \(\times\) solver interaction
analysis; falsifier table with raw numerators and denominators; second
wake-model pilot report and transfer-fold report or fallback/not-run
explanation; cost report; contamination/isolation manifest; and excluded-cell
report with reasons.  The reported v2 artifact bundle includes
\filepath{analysis/fold/v2_fold.json},
\filepath{analysis/fold/v2_confirmatory_clustered.json},
\filepath{analysis/fold/v2_hygiene_oracle.json},
\filepath{analysis/fold/v2_cost_frontier.json},
\filepath{analysis/fold/admission_funnel.md}, generated paper tables and
figures.  The current v2 analyzer and figure checksums are enumerated in
Appendix~\ref{app:artifact}; older timestamped hash ledgers under
\filepath{analysis/fold/} are historical diagnostics, not the paper-facing
checksum authority.

Allowed claims after V2 are bounded by the frozen analyses.  A benchmark-scale
claim is allowed only if the scaled trap set is frozen and valid.  The primary
system claim is allowed only if P1 passes the clustered Holm-corrected test.
Typed-maintenance and raw-evidence claims require their corresponding
pre-registered comparisons.  Construct-specific claims are descriptive and only
for constructs where the frozen descriptive direction supports them.  Transfer
claims are separate second-wake-model claims.  Clean-start claims require the
clean-start subset to support the direction.

Forbidden regardless of outcome: presenting failure to reject as equivalence;
calling B5 stock or tuned Mem0; pooling any scoped-out second wake-model fold with the
primary fold; removing traps after results; redesigning traps toward a probe-favorable
win; using LLM-authored secrets; claiming broad production generality from this
fixture-repository benchmark; claiming a favorable cost interpretation if the cost falsifier
fails; and claiming container-equivalent hermeticity for a host-side GRID wake
fold.

No result from the second wake-model fold, pooled McNemar analysis, construct
slice, family slice, authoring analysis, or hygiene analysis can replace the
primary clustered P1 result.

The v2.1 external-systems audit is governed by its own preregistration, erratum,
conformance lock, fixed Family A/Family B rules, and canonical analyzer.  It is
prospectively frozen for successor outcomes but post-hoc to original v2, and therefore
cannot revise, pool with, or replace this appendix's P1--P6 protocol.

%% file: sections/appendix_construct_taxonomy.tex
\section{v2 Construct Taxonomy}
\label{app:construct-taxonomy}

This appendix records the construct-level allocation used for the v2
\bench{} trap set.  The taxonomy and pure-trap criteria are drawn from
\filepath{analysis/investigation-evidence/FABLE-CONSTRUCTS.md}; the target
counts come from
\filepath{analysis/investigation-evidence/AUTHORING-WORKLIST.md} and the
frozen quota table in
\filepath{analysis/investigation-evidence/PREREGISTRATION-V2.md}.  These are
design and authoring counts, not live v2 outcomes; live v2 outcomes are
reported separately by the canonical analyzer artifacts and the results
section.

\begin{table}[htbp]
\centering
\scriptsize
\resizebox{\textwidth}{!}{%
\begin{tabular}{lp{0.18\textwidth}p{0.28\textwidth}p{0.32\textwidth}cc}
\toprule
Construct & Construct name & One-line definition & Pure-trap criterion & Anti-hoarding? & v2 trap count \\
\midrule
C1 & Verbatim retention &
Preserve a non-inferable exact token, format, or rule from one prior event and reproduce it when relevant. &
A single prior event contains the decisive arbitrary token or format, and storing that event is necessary and sufficient. &
No & 18 \\
C2 & Retrieval precision under interference &
Select the one relevant stored fact among many similar stored facts under a fixed read-token budget. &
The sequence injects many same-vocabulary distractor events, only one event is load-bearing, and the budget admits only a small subset. &
Yes & 6 \\
C3 & Staleness detection and supersession &
Use the superseding fact and suppress the stale fact, even when the stale fact is more salient. &
An older token or rule is repeated and lexically attractive, a later event supersedes it, and the oracle fails any use of the old value. &
Yes & 7 \\
C4 & Update propagation / regression avoidance &
Propagate a changed fact to a derived later location that the superseding event did not directly patch. &
The superseding event gives the new rule but not the S3 answer location; the oracle requires applying the rule consistently downstream. &
No & 1 \\
C5 & Scope discipline &
Apply feedback inside its stated scope and refuse to apply it outside that scope. &
The trap includes an in-scope check where memory must be used and an out-of-scope check where the same memory becomes a failure. &
Yes & 6 \\
C6 & Contradiction handling / provenance conflict &
Resolve conflicting sources by an established authority or precedence rule rather than recency or arbitrary choice. &
Two sources conflict without temporal order deciding the answer; a separate precedence rule determines which source the oracle accepts. &
Yes & 5 \\
C7 & Cross-session synthesis / paraphrase-retention application &
Derive the answer from two or more events, or apply a paraphrased retained instruction, with no single event containing the decisive S3 literal. &
The S3 output must compose rule, parameter, and/or runtime data split across events, so copying any one event is insufficient. &
No & 7 \\
C8 & Procedural / source-of-truth memory &
Remember how this repository expects a change to be made, such as editing a generator rather than a generated artifact. &
The oracle checks both the user-visible output and the diff shape, failing artifact-only edits when the source-of-truth workflow is required. &
No & 4 \\
C9 & Spurious-lesson rejection / disconfirmation &
Reject an unsupported or later-disconfirmed lesson, even when stored memory suggests a wrong causal story. &
A prior event records a false causal lesson, a later event disconfirms it, and S3 fails if the agent acts on the false lesson. &
Yes & 4 \\
C10 & Abstention / confabulation resistance &
Recognize that no prior event bears on the current task and avoid fabricating benchmark-style tokens or importing irrelevant memories. &
The S3 prompt resembles a memory trap, but no prior event is relevant; the oracle fails confabulated tokens or irrelevant memory use. &
Yes & 2 \\
\bottomrule
\end{tabular}
}%
\caption{Construct taxonomy and frozen v2 design quota.  ``Anti-hoarding''
records the preregistered allocation: C2, C3, C5, C6, C9, and C10 contribute
\(6+7+6+5+4+2=30\) designed traps, exceeding the 20-of-60 floor.  Individual C7
traps can also be tagged anti-hoarding when they meet the preregistered
paraphrase-retention rule.  This is allocation arithmetic, not an outcome
guarantee: the observed C9 and C10 strata fail B0-headroom validity and do not
support anti-hoarding or abstention claims.}
\label{tab:construct-taxonomy}
\end{table}

The taxonomy is measurement-theoretically distinct because each construct
defines a different failure mechanism, not merely a different topic label.
C1 tests single-event exact recall.  C2 tests selection under interference.  C3
tests stale suppression, while C4 tests propagation of a changed rule to a
derived location.  C5 turns memory into a scope hazard, C6 turns memory into an
authority-resolution problem, C7 requires composition across events, C8 requires
remembering the repository's procedural source of truth, C9 requires rejecting a
disconfirmed stored lesson, and C10 requires abstaining when memory is
irrelevant.  The construct label is therefore tied to the oracle's commission
and omission checks: what counts as a pass or trap-trip differs by memory
hygiene mechanism.

This also corrects a v1 labeling weakness.  The v1 family names were continuity
labels such as \texttt{flaky-test}, \texttt{reviewer-preference}, and
\texttt{stale-architecture}; they did not reliably name the measured construct.
For example, the worklist maps all v1 \texttt{flaky-test} traps audited as
recall-verbatim to C1 unless redesigned and revalidated.  In v2, family labels
are retained only for continuity, while each trap carries a separate C1--C10
construct label and the confirmatory quota is governed by construct rather than
legacy family.  The live construct-profile results are reported from the
canonical v2 analyzer bundle and remain descriptive unless tied to a frozen
comparison.

%% file: sections/appendix_clustered_stats.tex
\section{Clustered Statistical Methods}
\label{app:clustered-stats}

This appendix records the clustered inference model used for the scaled
\bench{} confirmatory fold and explains how it relates to the pooled McNemar
analysis reported for the 22-trap v1 fold.  The method source of truth for the
v1 sensitivity check is
\filepath{analysis/investigation-evidence/CLUSTERED-STATS.md}, regenerated by
\filepath{scripts/clustered_sensitivity.py}.  The scaled-fold power arithmetic
is the frozen arithmetic in
\filepath{analysis/investigation-evidence/PREREGISTRATION-V2.md}.

\subsection{Trap-as-Cluster Model}
\label{app:clustered-stats-model}

The statistical cluster is the trap, not the seed-trap cell.  A trap is one
multi-session sequence whose S3 task is the scored memory trap.  Seeds
\(s \in \{1,2,3\}\) are repeated observations of that same trap.  For a paired
comparison between conditions \(A\) and \(B\), let \(Y_{Ats}\in\{0,1\}\) and
\(Y_{Bts}\in\{0,1\}\) be the validity-gated S3 pass/fail outcomes for trap
\(t\) and seed \(s\).  A seed contributes to the comparison only when both
conditions have valid S3 outcomes for that \((t,s)\) cell; no imputation is
used.

For each trap, define the signed trap contribution
\[
d_t =
\sum_{s \in \mathcal{S}_t}
\left[
\mathbf{1}\{Y_{Ats}=1,Y_{Bts}=0\}
-
\mathbf{1}\{Y_{Ats}=0,Y_{Bts}=1\}
\right],
\]
where \(\mathcal{S}_t\subseteq\{1,2,3\}\) is the set of paired valid seeds for
trap \(t\).  Tied paired seed cells, where both systems pass or both fail,
contribute zero.  A trap with no paired valid seeds for the comparison is
excluded from that comparison and listed as missing.  The observed signed
effect is
\[
D_{\mathrm{obs}}=\sum_t d_t,
\]
which equals the cell-level discordance difference \(b-c\), where \(b\) is the
number of paired seed cells where \(A\) passes and \(B\) fails, and \(c\) is
the number where \(B\) passes and \(A\) fails.  The reported two-sided test
statistic is
\[
T_{\mathrm{obs}}=\left|D_{\mathrm{obs}}\right|.
\]

In the v2 design, a complete condition has
\[
60\ \mathrm{traps}\times 3\ \mathrm{seeds}=180
\]
S3 cells.  The primary paired unit remains the 60 trap clusters.  The 180 cells
are descriptive repeated measurements inside those 60 clusters, not 180
independent benchmark items.  In the observed v2 P1 comparison,
\texttt{DF-hybrid} versus B5 has \(b=21\), \(c=15\),
\(D_{\mathrm{obs}}=+6\), 60 contributing clusters, 180 paired seed cells,
permutation \(p=0.5184192657470703\), and Holm-adjusted \(p=1.0\).

\subsection{Exact Trap-Clustered Sign/Permutation Test}
\label{app:clustered-stats-test}

The null hypothesis for a paired comparison is condition-label exchangeability
at the trap level.  For each trap \(t\), the permutation swaps the labels
\(A\) and \(B\) for all available paired seed cells in that trap together.  It
does not independently swap seed 1, seed 2, and seed 3 within the same trap.
Equivalently, write \(\epsilon_t\in\{-1,+1\}\) for the trap-level label-swap
sign.  The exact randomization distribution is
\[
D(\epsilon)=\sum_t \epsilon_t d_t,
\]
with all sign assignments equally likely.  For the complete v2 design this is
a \(2^{60}\)-assignment trap-level permutation over the
\(60\times 3=180\) seed-trap cells: each assignment flips or preserves all
three seed cells for each trap as one block.  Zero-contribution traps may be
retained in the enumeration for descriptive consistency; they do not change
the resulting \(p\)-value.  The exact two-sided \(p\)-value is
\[
p =
\Pr_{\epsilon}\left(
  \left|\sum_t \epsilon_t d_t\right|
  \geq
  \left|D_{\mathrm{obs}}\right|
\right).
\]
The implementation may enumerate signs directly for small \(T\) or compute the
same distribution by exact dynamic programming over the possible signed sums.
No normal approximation is used for the primary \(p\)-value.  The primary
family correction for v2 is Holm-Bonferroni over the six pre-registered
clustered comparisons.  The observed adjusted values are: P1 \(1.0\), P2
\(1.0\), P3 \(1.0\), P4 \(0.7672004699707031\), P5
\(0.7672004699707031\), and P6 \(1.0\).

\subsection{Successor Families and Unavailable Slots}
\label{app:clustered-stats-v21}

The separately preregistered v2.1 audit uses the same trap-level signed statistic and
exact sign-flip distribution, but it does not join the original P1--P6 family.  Its
Family A has six fixed benchmark-discrimination slots and Family B has two fixed
mechanism slots.  A condition rejected by the frozen pre-evaluation conformance gate is
unavailable: its registered comparison remains in its family with raw and adjusted
$p=1$, rather than being dropped, imputed, or interpreted as an observed loss.  Holm
correction is applied separately within the six- and two-slot families.  The two
literal-storage comparisons against B5 and DF-hybrid are explicitly secondary,
unadjusted, and nonconfirmatory; their raw clustered, per-seed, pooled-cell, and
trap-majority results are sensitivity descriptions, not additional family members.

\subsection{Why Pooled McNemar Over-Counts Evidence}
\label{app:clustered-stats-mcnemar}

Pooled exact McNemar treats every paired seed-trap cell as an exchangeable
paired unit.  In the v1 fold, that meant up to \(22\times 3=66\) cells; in the
scaled v2 fold it would mean up to \(60\times 3=180\) cells.  This is useful as
a secondary sensitivity calculation, but it over-counts evidence for primary
inference because the three seeds for a trap share the same hidden contract,
repository pathology, oracle, and implementation burden.  If condition \(A\)
wins the same trap on all three seeds, pooled McNemar counts three
same-direction discordant observations.  The clustered test counts one
trap-level directional contribution with magnitude three and assigns one
exchangeable sign to the trap.  The distinction is exactly the pseudo-
replication issue: 66 or 180 seed-trap cells are correlated within 22 or 60
trap clusters, respectively.

For this reason, pooled McNemar cannot replace the primary clustered test in
v2.  It is reported as a secondary cell-level sensitivity check, with the same
two-sided alternative and family correction, while the headline claim is tied
to the trap-clustered exact test.

\subsection{v1 Clustered Sensitivity Result}
\label{app:clustered-stats-v1}

The v1 sensitivity analysis used the same trap contribution
\(d_t=\#(A\mathrm{\ only})-\#(B\mathrm{\ only})\) and the same exact sign-flip
test, applied to the 22 v1 traps and three seeds.  In the pre-registered
five-comparison v1 family, the one result that survived the clustered
sensitivity check was reference-probe typed-only (DF) versus B3 reflection-only vector
retrieval.  The pooled McNemar row had \(n=65\) paired cells because the
reference-probe typed-only condition was missing
\filepath{seed3:expr-stale-registry-stability}; it reported \(b=24\), \(c=3\),
exact \(p=4.92334365845\times 10^{-5}\), and Holm-adjusted
\(p=0.000246167182922\) (rounded in the paper as \(0.0002\)).  The
trap-clustered exact sign/permutation test for the same comparison had 22
contributing trap clusters, 65 paired seed cells, observed \(b-c=21\), raw
clustered \(p=0.00732421875\), and Holm-adjusted
\(p=0.03662109375\) (rounded as \(0.037\)).

This survival is narrow.  The same v1 result did not pass every conservative
sensitivity variant: all three per-seed tests did not reject after seed-family
Holm correction, and the complete three-seed trap-majority collapse had
Holm-adjusted \(p=0.1953125\).  Therefore the v1 paper-level reading is not
that pooled inference was adequate, but that the structured-memory comparison
against B3 remained positive under the exact trap-clustered test while carrying
seed-concentration and trap-collapse caveats.

\subsection{Power and Minimum Detectable Effects at 180 Cells}
\label{app:clustered-stats-power}

The 60-trap v2 design is not powered by 180 independent cells.  Its primary
power depends on the number of traps with nonzero discordance and on how
discordant seeds cluster within each trap.  For intuition, if \(k\) nonzero
traps have equal weight, the trap-clustered test reduces to an exact sign test.
If \(b\) of those \(k\) traps favor the reference probe and \(c=k-b\) favor B5, with
\(b>k/2\), the two-sided exact sign-test \(p\)-value is
\[
p = 2 \sum_{i=b}^{k} {k \choose i} 2^{-k}.
\]
The equal-weight rejection thresholds at \(\alpha=0.05\), using the frozen
180-cell denominator for descriptive cell margins, are:

\begin{table}[htbp]
\centering
\scriptsize
\resizebox{\textwidth}{!}{%
\begin{tabular}{rrrrrr}
\toprule
Nonzero trap clusters \(k\) & Minimum split & Exact \(p\) &
One discordant seed/trap & Net margin &
All three seeds discordant/trap \\
\midrule
20 & 15--5 & 0.041389 & \(20/180=0.111\) & \(10/180=0.056\) & \(60/180=0.333\), net \(30/180=0.167\) \\
30 & 21--9 & 0.042774 & \(30/180=0.167\) & \(12/180=0.067\) & \(90/180=0.500\), net \(36/180=0.200\) \\
40 & 27--13 & 0.038477 & \(40/180=0.222\) & \(14/180=0.078\) & \(120/180=0.667\), net \(42/180=0.233\) \\
50 & 33--17 & 0.032839 & \(50/180=0.278\) & \(16/180=0.089\) & \(150/180=0.833\), net \(48/180=0.267\) \\
60 & 39--21 & 0.027340 & \(60/180=0.333\) & \(18/180=0.100\) & \(180/180=1.000\), net \(54/180=0.300\) \\
\bottomrule
\end{tabular}
}%
\caption{Equal-weight sign-test thresholds for the 60-trap, 180-cell v2
design.  The cell rates are descriptive translations of trap-level splits, not
the primary independent sample size.}
\label{tab:clustered-power-thresholds}
\end{table}

For the full \(k=60\) equal-weight case, the exact threshold computation is:
\[
2 \sum_{i=38}^{60} {60 \choose i}2^{-60}
=0.051894,
\]
so a 38--22 trap split is not below \(0.05\).  The next split is:
\[
2 \sum_{i=39}^{60} {60 \choose i}2^{-60}
=0.027340,
\]
so 39--21 is the smallest equal-weight split below \(0.05\).  If exactly one
seed is discordant in each of those 60 traps, the descriptive cell-level net
margin is
\[
(39-21)/180 = 18/180 = 0.100.
\]
If all three seeds are discordant in the same direction within each nonzero
trap, the corresponding cell counts are \(39\times 3=117\) DF-favoring cells
and \(21\times 3=63\) B5-favoring cells, giving
\[
(117-63)/180 = 54/180 = 0.300.
\]
Thus the same trap-level rejection boundary can correspond to a 10 percentage
point or 30 percentage point cell-level net margin, depending on within-trap
seed clustering.

Under the same equal-weight sign model, the frozen approximate power values
are:

\begin{table}[htbp]
\centering
\small
\begin{tabular}{rrrrr}
\toprule
Nonzero clusters \(k\) & Rejection threshold &
Power at 0.65 & Power at 0.70 & Power at 0.75 \\
\midrule
20 & \(\geq 15/20\) & 0.245 & 0.416 & 0.617 \\
30 & \(\geq 21/30\) & 0.358 & 0.589 & 0.803 \\
40 & \(\geq 27/40\) & 0.441 & 0.703 & 0.897 \\
50 & \(\geq 33/50\) & 0.506 & 0.782 & 0.945 \\
60 & \(\geq 39/60\) & 0.559 & 0.838 & 0.970 \\
\bottomrule
\end{tabular}
\caption{Frozen approximate power under the equal-weight trap-level sign
model, parameterized by the true probability that a nonzero trap favors
the reference probe.}
\label{tab:clustered-power}
\end{table}

The design therefore has reasonable power only for a large, stable trap-level
reference-probe advantage.  It does not guarantee detection of small effects, sparse
discordance, or effects concentrated in a small number of trap templates.  The
reported fold computes the observed \(d_t\) vectors and primary p-values in
\filepath{analysis/fold/v2_confirmatory_clustered.json}.

%% file: sections/appendix_reproducibility.tex
\section{Reproducibility Appendix}
\label{app:reproducibility}

This appendix is an executable runbook for the \bench{} confirmatory fold.  The
data-card pointer for the benchmark release is \texttt{docs/DATASHEET.md}.  This
checkout also keeps artifact-level notes in \texttt{artifact/README.md} and
source-level validation rules in \filepath{docs/trap_skeleton_spec.md}.

\paragraph{Repository layout.}
The source tree separates paper text, benchmark code, frozen records, and live
rerun machinery:

\begin{itemize}
  \item \texttt{paper/}: manuscript source.  \texttt{paper/main.tex} defines the
  reference probe and \bench{} macros and inputs appendix files.
  \item \texttt{src/}: Python package code for agents, benchmark adapters,
  experiment runners, and memory/scoring modules.
  \item \filepath{src/experiments/run_bench.py}: the benchmark runner used by
  \filepath{scripts/run_grid.py}; invoke it as \filepath{python3 -m experiments.run_bench}
  with \texttt{PYTHONPATH=src}.
  \item \texttt{scripts/}: validation, grid, analysis, packaging, and container
  support scripts.  The container recipes are
  \filepath{scripts/Dockerfile.codex-agent} and
  \filepath{scripts/Dockerfile.api-agent}.
  \item \texttt{experiments/env/}: sequence records and benchmark environment
  assets.  \filepath{experiments/env/sequences_confirmatory_v2.jsonl} is the
  scaled v2 confirmatory sequence file; \filepath{experiments/env/sequences.jsonl}
  is retained for the historical v1/public-smoke fixture set. Hidden scoring assets live under
  \filepath{experiments/env/oracles/} and \filepath{experiments/env/refsol/}.
  \item \texttt{experiments/results/}: default result root.  Override with
  \filepath{DREAMBENCH_RESULTS_ROOT} to keep new reruns separate from frozen
  evidence.
  \item \texttt{logs/grid/}: default grid subprocess log root.  Override with
  \filepath{DREAMBENCH_LOG_ROOT}.
  \item \texttt{analysis/fold/} and \texttt{analysis/investigation-evidence/}: folded
  analysis JSON, table inputs, and human-readable investigation reports.
  \item \texttt{artifact/} and \texttt{dist/}: public artifact runbook and packaged
  tarball/checksum outputs.
\end{itemize}

\paragraph{Container images.}
The wake and judge boundary is container-filesystem isolation, not network
isolation.  The Codex wake/judge image defaults to
\texttt{dreambench-swe-codex-agent:latest}; the OpenAI-compatible GRID/API wake
image defaults to \texttt{dreambench-swe-api-agent:latest}.  The harness can build
missing images automatically, but explicit builds are:

\begin{CodeBlock}
docker build -t dreambench-swe-codex-agent:latest -f scripts/Dockerfile.codex-agent .
docker build -t dreambench-swe-api-agent:latest -f scripts/Dockerfile.api-agent .
\end{CodeBlock}

The default image names can be overridden with
\filepath{DREAMBENCH_CODEX_AGENT_IMAGE} and
\filepath{DREAMBENCH_GRID_AGENT_IMAGE}.  The corresponding container networks
can be overridden with \filepath{DREAMBENCH_CODEX_CONTAINER_NETWORK} and
\filepath{DREAMBENCH_GRID_CONTAINER_NETWORK}.  The API-agent container receives
\filepath{GRID_BASE_URL} and \filepath{GRID_API_KEY} as environment variables;
hidden oracles, reference solutions, the harness repository, and local
credentials must not be mounted into the wake worktree.

\paragraph{Seeds and source freezing.}
The confirmatory fold uses seeds \(1,2,3\), passed to \filepath{scripts/run_grid.py}
with \texttt{--seeds 1,2,3}.  The grid runner forwards each seed to
\filepath{experiments.run_bench} as \texttt{--seed}.  Direct invocations of
\filepath{experiments.run_bench} default to seed \texttt{1729} unless
\texttt{--seed} is supplied.

Each sequence record binds the task to a frozen repository state through
\texttt{repo} and \filepath{initial_commit}.  The runner materializes the
sequence from that commit, advances later sessions through continuation commits,
and records per-session \filepath{start_commit}, \filepath{end_commit},
\filepath{previous_session_end_commit}, \filepath{oracle_id}, condition, seed,
run id, model, and budget metadata in \texttt{results.json}.  A result record is
therefore evidence for the sequence source, hidden oracle/refsol assets, harness
code, container image, model route, and result directory that produced it.
Editing sequence records, oracle files, reference solutions, harness code, or
container recipes after a fold has been frozen creates a new experiment rather
than an exact replay of the paper fold.  Use a fresh
\filepath{DREAMBENCH_RESULTS_ROOT} for such runs.

\paragraph{CSPRNG secret protocol.}
New traps are authored as skeletons with placeholders and finalized by
\filepath{scripts/inject_secrets.py}.  The injector uses Python's
\texttt{secrets} module, writes finalized sequence records, hidden oracles,
reference-solution diffs, and a \filepath{secret_provenance.json} sidecar.  The
sidecar records the skeleton id, sequence ids, generator path/hash, RNG class,
and placeholder format classes; it does not record generated secret values.
There is deliberately no \texttt{--seed}, \texttt{--deterministic}, or replay
flag for the injector:

\begin{CodeBlock}
python3 scripts/inject_secrets.py <skeleton-dir-or-jsonl> <output-dir-or-jsonl>
\end{CodeBlock}

Optional injector flags are \texttt{--manifest}, \texttt{--leakage-root}, and
\texttt{--provenance-name}.  A rerun of the injector over the same skeleton
should produce fresh non-inferable literals, so it is a new benchmark instance.
The hidden oracles stay stable within a finalized instance because the same
generated value replaces each placeholder everywhere that placeholder occurs:
the public sequence entry, hidden oracle file, reference solution, and any
decisive-literal metadata for that trap.  Post-injection checks fail on unused
or undeclared placeholders, leftover \texttt{\{\{...\}\}} tokens, same-run value
collisions, placeholder reuse across traps, or generated values leaking outside
the trap's allowed sequence/oracle/refsol assets.

\paragraph{Batch validation.}
\filepath{scripts/batch_validate.py} accepts a JSON/JSONL sequence file or a
directory of candidate sequence files.  Dry validation parses records, checks
oracle-file existence, checks reference-solution application, and applies the
no-single-event gate for synthesis records:

\begin{CodeBlock}
PYTHONPATH=src python3 scripts/batch_validate.py \
  experiments/env/sequences_confirmatory_v2.jsonl \
  --dry
\end{CodeBlock}

Live validation is the default mode when \texttt{--dry} is absent; there is no
\texttt{--live} flag in the script.  A live validation command is:

\begin{CodeBlock}
PYTHONPATH=src python3 scripts/batch_validate.py \
  experiments/env/sequences_confirmatory_v2.jsonl \
  --workers 4 \
  --timeout-seconds 1800
\end{CodeBlock}

\begin{sloppypar}
The validator writes batch summaries under \filepath{experiments/validation/} by
default.  Use \texttt{--validation-root} or \texttt{--output} to choose another
location.  Use \texttt{--resume} to skip matching prior \filepath{seq_id} plus
\filepath{record_hash} entries from existing batch summaries.  To name an
explicit prior summary, pass \texttt{--resume-from} with that path.
\end{sloppypar}

\paragraph{Grid launch.}
For a launch plan without subprocesses, use \texttt{--dry} on the grid runner:

\begin{CodeBlock}
PYTHONPATH=src python3 scripts/run_grid.py \
  --conditions B0,DF \
  --seeds 1,2,3 \
  --group-size 2 \
  --max-parallel 8 \
  --judge-model codex-gpt-5.5 \
  --sequence-records experiments/env/sequences_confirmatory_v2.jsonl \
  --dry
\end{CodeBlock}

The current live-rerun launcher \filepath{ops/launch_confirmatory.sh} enumerates
the complete 19-condition non-live matrix and both hosted-Mem0 configurations:

\begin{CodeBlock}
PYTHONPATH=src python3 scripts/run_grid.py \
  --conditions "B0,B1,B2,B3,B4,B5,B6,B7,DF,DF-strict,DF-hybrid,DF-raw-only,DF-strict-hybrid,A0,A2,A4,A5,A6,A11" \
  --seeds 1,2,3 \
  --group-size 2 \
  --max-parallel 16 \
  --judge-model codex-gpt-5.5 \
  --sequence-records experiments/env/sequences_confirmatory_v2.jsonl

PYTHONPATH=src python3 scripts/run_grid.py \
  --conditions "B5-MEM0,B5-MEM0-LIT" \
  --seeds 1,2,3 \
  --group-size 2 \
  --max-parallel 4 \
  --judge-model codex-gpt-5.5 \
  --sequence-records experiments/env/sequences_confirmatory_v2.jsonl
\end{CodeBlock}

This launcher is a corrected replay-ready runbook, not a claim that one shell
invocation produced the historical fold.  The reported v2 records were assembled
across resumable launch waves and isolated result roots.  Their completeness is
established by the canonical 1,890-file completion audit and folded records, not
by reconstructing command history from this script.  Running the script now
creates a new hosted-model experiment rather than an exact replay.

\begin{sloppypar}
\filepath{scripts/run_grid.py} scans the chosen results root for completed
units and skips matching \filepath{(condition, seed, seq_ids)} groups that
already have a completed \texttt{results.json} plus gate report.  The B5-MEM0
and B5-MEM0-LIT conditions are special-cased by the grid runner: it appends
\texttt{--include-live-baselines} when launching the child benchmark process.
If invoking \filepath{experiments.run_bench} directly for either condition, pass that
flag yourself.
\end{sloppypar}

A direct dry-run smoke through the benchmark runner is:

\begin{CodeBlock}
PYTHONPATH=src python3 -m experiments.run_bench \
  --conditions B0,DF \
  --sequence-records experiments/env/sequences_confirmatory_v2.jsonl \
  --dry-run \
  --seed 1 \
  --limit 2
\end{CodeBlock}

\paragraph{Analyzer path and S3 definition.}
The canonical v2 path first folds raw result records with
\filepath{scripts/analyze_confirmatory.py} and then runs the clustered v2 analyzer:

\begin{CodeBlock}
DREAMBENCH_ROOT=. PYTHONPATH=src python3 scripts/analyze_confirmatory.py \
  --results-root "${DREAMBENCH_RESULTS_ROOT:-experiments/results}"

cp analysis/fold/confirmatory.json analysis/fold/v2_fold.json

PYTHONPATH=src python3 scripts/analyze_confirmatory_v2.py \
  --analysis-json analysis/fold/v2_fold.json \
  --permutation-seed 20260705 \
  --primary-pair DF-hybrid B5 \
  --family DF-hybrid B5 \
  --family DF B5 \
  --family DF-raw-only B5 \
  --family DF-hybrid DF-raw-only \
  --family DF-hybrid DF \
  --family DF B3 \
  --output-json analysis/fold/v2_confirmatory_clustered.json
\end{CodeBlock}

The first-stage analyzer discovers \filepath{<results-root>/*-SLICE-*/*/results.json}, keeps the newest
record per \filepath{(condition, seed, oracle_id)} by the timestamp embedded in
the \filepath{results/<YYYYMMDDTHHMMSSZ>-SLICE-*} directory name, and rewrites
\filepath{analysis/investigation-evidence/CONFIRMATORY-FOLD.md} plus
\filepath{analysis/fold/confirmatory.json}; the v2 analyzer resolves the
pre-registered clustered P-family from \filepath{analysis/fold/v2_fold.json}.
The canonical S3 predicate is:

\begin{CodeBlock}
oracle_id endswith "-s3" OR ordinal == 3
\end{CodeBlock}

Do not replace this with an ordinal-only filter.  Grouped jobs can place a
sequence's S3 record at a later ordinal inside a multi-sequence child run, so
the analyzer's \filepath{oracle_id} suffix rule is the source of truth.

\paragraph{Pinned hosted-Mem0 failure analysis.}
\label{sec:appendix-mem0-failure}
The pinned hosted-Mem0 configurations are reported only as supplemental failure
analysis, not as headline memory-system comparisons and not as claims against
tuned Mem0 or Mem0$^g$.  In the v2 diagnostic fold, B5-MEM0 reaches
$21/180=0.117$ and B5-MEM0-LIT reaches $20/180=0.111$, close to the B0
no-memory rate of $21/180=0.117$.  They are excluded from the generated
headline ladder and construct tables.

The historical v1 hosted-Mem0 row helps explain the mechanism.  In that fold,
the pinned hosted-Mem0 configuration completed 66 S3 records with 0
validity-gate exclusions and reached pooled S3 pass@1 $6/66=0.091$.  Sampled
failures show that hosted Mem0 retrieved the relevant memory at S3, but its
LLM-based fact extraction did not preserve exact non-inferable tokens; for
example, a redacted sampled failure had the form
\texttt{TOKEN-AAAA-BBBB}, while the retrieved fact rewrote the ASCII hyphen
(U+002D) to a Unicode non-breaking hyphen (U+2011), yielding a byte-wrong
marker that the hidden oracle rejected.  This is a diagnostic failure mode for
one pinned configuration, not a practical ranking claim.

\paragraph{Successor replay boundary.}
The v2.1 analyzer was executed exactly once after terminal completion and immutable
input-ledger validation.  It must not be rerun against the admitted run identity.
Post-analysis regeneration is deterministic and read-only with respect to canonical
results:

\begin{CodeBlock}
python3 scripts/generate_external_audit_v21_artifacts.py \
  --analysis-json analysis/fold/v21_external_audit_20260716T230000Z.json \
  --output-dir <absent-output-directory>

python3 scripts/check_v21_artifact_freshness.py \
  --analysis-json analysis/fold/v21_external_audit_20260716T230000Z.json \
  --surface-dir <generated-surface-directory>

python3 scripts/check_paper_numeric_provenance_v21.py \
  --analysis-json analysis/fold/v21_external_audit_20260716T230000Z.json \
  --surface-dir surfaces --source-root paper --source-root paper_arXiv
\end{CodeBlock}

Live reruns are new experiments because the wake model and hosted external service can
change.  The public verifier checks path-neutral evidence projections and source/hash
bindings; it does not expose private raw result, log, receipt, credential, or oracle
material.

\paragraph{Machine-generated successor quantitative surface.}
The following ordered panels are included directly from the analyzer-bound generated
table artifact: condition profile, Family A, Family B, temporal control, bounded
construct diagnostics, and operational diagnostics.  The operational cost column
contains recorded benchmark cost only; unavailable external-service prices are not
silently imputed.
\begingroup
\scriptsize
\setlength{\tabcolsep}{3pt}
\input{analysis/fold/v21_external_audit_tables}
\endgroup

%% file: analysis/fold/v21_external_audit_tables.tex
% Generated from validated dreambench-external-audit-analysis-v1 JSON.
% Do not edit values by hand.
\begin{tabular}{lrrrrl}
Condition & Passed & Valid & Planned & Rate & Eligible \\
\hline
B0 & 21 & 180 & 180 & 0.116667 & yes \\
B5 & 82 & 180 & 180 & 0.455556 & yes \\
DF-hybrid & 83 & 180 & 180 & 0.461111 & yes \\
B5-MEM0-LIT & 97 & 180 & 180 & 0.538889 & yes \\
\end{tabular}

% family_a
\begin{tabular}{lllll}
Comparison & Availability & Confirmatory eligible & Raw $p$ & Holm $p$ \\
\hline
B5 vs B0 & available & yes & 0.000324063 & 0.00129625 \\
DF-hybrid vs B0 & available & yes & 1.56049e-06 & 9.36294e-06 \\
B5-MEM0 vs B0 & unavailable & no & 1 & 1 \\
B5-MEM0-LIT vs B0 & available & yes & 7.85204e-06 & 3.92602e-05 \\
B5-SM-LOCAL vs B0 & unavailable & no & 1 & 1 \\
B5-SM-LOCAL-DOC vs B0 & unavailable & no & 1 & 1 \\
\end{tabular}

% family_b
\begin{tabular}{lllll}
Comparison & Availability & Confirmatory eligible & Raw $p$ & Holm $p$ \\
\hline
B5-MEM0-LIT vs B5-MEM0 & unavailable & no & 1 & 1 \\
B5-SM-LOCAL-DOC vs B5-SM-LOCAL & unavailable & no & 1 & 1 \\
\end{tabular}

\begin{tabular}{lrrl}
Temporal control & Concurrent rate & Absolute delta & Within band \\
\hline
B0 & 0.116667 & 0 & yes \\
B5 & 0.455556 & 0.0388889 & yes \\
DF-hybrid & 0.461111 & 0.0666667 & yes \\
\end{tabular}

\begin{tabular}{lrrrrl}
Construct & B0 & B5 & DF-hybrid & B5-MEM0-LIT & Reading \\
\hline
C1 & -- & -- & -- & -- & descriptive only \\
C2 & 0 & 0.666667 & 0.444444 & 0.833333 & bounded directional set \\
C3 & 0 & 0.533333 & 0.266667 & 0.733333 & bounded directional set \\
C4 & -- & -- & -- & -- & descriptive only \\
C5 & 0 & 0.166667 & 0.166667 & 0.0555556 & bounded directional set \\
C6 & 0 & 0.166667 & 0.25 & 0.583333 & bounded directional set \\
C7 & 0 & 0.333333 & 0.444444 & 0.555556 & bounded directional set \\
C8 & -- & -- & -- & -- & descriptive only \\
C9 & 1 & 0 & 0.75 & 0.25 & descriptive only \\
C10 & 1 & 0 & 0.333333 & 0 & descriptive only \\
\end{tabular}

\begin{tabular}{lrrrrrr}
Condition & Records & Judge calls & Receipt calls & Cost & Tokens & Latency p95 \\
\hline
B0 & 540 & 0 & 0 & 10.5326 & 2.10652e+06 & 533.218 \\
B5 & 540 & 0 & 0 & 10.8971 & 2.20936e+06 & 505.46 \\
DF-hybrid & 540 & 12655 & 0 & 11.7601 & 1.74205e+08 & 3503.39 \\
B5-MEM0-LIT & 540 & 0 & 1080 & 13.1401 & 2.86524e+06 & 510.82 \\
\end{tabular}

%% file: sections/appendix_notation.tex
\section{Notation and Glossary}
\label{app:notation}

This appendix is a reference index for notation, condition codes, construct
labels, metrics, and statistical terms used in the paper.  It introduces no new
empirical claims.  Source definitions are the paper sections, especially
Sections~\ref{sec:method}, \ref{sec:benchmark}, and \ref{sec:results};
\filepath{paper/sections/appendix_construct_taxonomy.tex};
\filepath{paper/sections/appendix_artifact.tex}; and the fold analyzer
\filepath{scripts/analyze_confirmatory.py}.  Metric helper formulas that are named
but not fully expanded in the paper are mirrored from
\filepath{src/dream_memory/evaluation.py} and the label-grounded scorer notes in
\filepath{src/dream_memory/slice_scorer.py}.  V2 construct-profile outcomes are
reported by the canonical analyzer artifacts and remain descriptive unless tied
to a pre-registered comparison.

\subsection{Core Notation}

\begin{table}[htbp]
\centering
\small
\begin{tabular}{p{0.20\linewidth}p{0.72\linewidth}}
\toprule
\textbf{Symbol or term} & \textbf{Meaning} \\
\midrule
\(A\) & Software-engineering agent under evaluation. \\
\(R\) & Repository for a task or sequence. \\
\(T\) & Task prompt or task instance. \\
\(S_i\) & The \(i\)th bounded session in a multi-session sequence.  In the
memory-hygiene objective, \(S_i \subseteq R_i\) is also used for the subset of
retrieved memories that should have been suppressed; context disambiguates the
two uses. \\
Sequence & Ordered list of sessions over one repository or a controlled family
of related repositories. \\
\(\tau\) & Agent trajectory,
\(\tau=(\mathrm{meta}, e_1, e_2,\ldots,e_n,\mathrm{outcome})\). \\
\(\mathrm{meta}\) & Trajectory identifiers and budgets: task id, session id,
repository id, starting commit, model id, condition id, seed, and budgets. \\
\(e_i\) & Observed trajectory event: state, model message, tool call, tool
result, file edit, test result, memory read, memory write, reviewer message, or
termination event. \\
\(\mathrm{outcome}\) & Final diff, tests run, pass/fail status when executable
verification exists, cost, latency, and labels. \\
Raw episode & Persisted trajectory or trajectory segment with immutable evidence
semantics. \\
\(E\) & Append-only raw episode log.  reference-probe operators may read \(E\) but may
not overwrite, truncate, summarize in place, hard-delete, or garbage-collect raw
episodes during an experiment. \\
\(M\) & Derived memory store.  Sleep operators append derived memories, append
audit records, or change lifecycle status of derived memories. \\
\(m\) & Derived memory item with content, type, provenance, write reason, id,
status, scope fields, timestamps, validity interval, confidence, utility, risk,
staleness, contradiction and supersession links, retrieval tags, usage
accounting, and optional outcome impact. \\
\(q\) & Task context used for memory retrieval and admission. \\
\(R_i\) & Set of memory items admitted into session \(i\). \\
\(U_i \subseteq R_i\) & Admitted items judged useful. \\
\(H_i \subseteq R_i\) & Admitted items judged harmful. \\
\(G(q,m)\) & Hard-gate admission decision for candidate memory \(m\) under task
context \(q\).  If \(G(q,m)=0\), the item receives score \(-\infty\). \\
\(\mathrm{score}(q,m)\) & Retrieval ranking score for admitted memories under
the fixed memory-token budget. \\
\bottomrule
\end{tabular}
\caption{Base notation from the problem formulation and method sections.}
\label{tab:notation-core}
\end{table}

\begin{table}[htbp]
\centering
\small
\begin{tabular}{p{0.22\linewidth}p{0.70\linewidth}}
\toprule
\textbf{Read-score symbol} & \textbf{Meaning} \\
\midrule
\(J(T_q,T_m)\) & Tag-overlap similarity between task-context tags and memory
tags. \\
\(F(q,m)\) & File-scope compatibility score. \\
\(Y(q,m)\) & Symbol-scope compatibility score. \\
\(K(q,m)\) & Task-scope compatibility score. \\
\(\pi_{\mathrm{type}}(m)\) & Condition-specific memory-type prior. \\
\(c_m\), \(u_m\), \(v_m\), \(p_m\) & Confidence, utility, human-verification,
and provenance-strength metadata. \\
\(s_m\), \(r_m\), \(\Delta t_m\), \(o(q,m)\) & Staleness, risk, age, and
overscope penalty terms. \\
Retrieval-score weights \(w_{\cdot}\) & Positive weights for semantic, tag,
file, symbol, task, type, confidence, utility, verification, and provenance
terms; fixed before held-out evaluation. \\
Penalty weights \(\lambda_{\cdot}\) & Penalty weights for staleness, risk, age,
and overscope terms; fixed before held-out evaluation. \\
\(\theta_{\mathrm{admit}}\) & Strict retrieval admission threshold used by
DF-strict and DF-strict-hybrid; in the reported strict setting,
\(\theta_{\mathrm{admit}}=1.5\). \\
\bottomrule
\end{tabular}
\caption{Retrieval-gate notation from the method section.}
\label{tab:notation-retrieval}
\end{table}

\FloatBarrier
\clearpage
\subsection{Sessions, Outcomes, and Aggregation}

\begingroup
\small
\begin{longtable}{@{}p{0.25\linewidth}p{0.67\linewidth}@{}}
\caption{Session, outcome, and aggregation terms used in the fold.}
\label{tab:notation-outcomes}\\
\toprule
\textbf{Term} & \textbf{Definition} \\
\midrule
\endfirsthead
\caption[]{Session, outcome, and aggregation terms used in the fold, continued.}\\
\toprule
\textbf{Term} & \textbf{Definition} \\
\midrule
\endhead
S1, S2 & Setup or reinforcement sessions that create prior sequence state and
memory evidence before the scored trap. \\
S3 & Trap session.  In the analyzer, S3 means
\filepath{oracle_id} ends with \texttt{-s3} or \texttt{ordinal == 3}.  The
reported headline outcome is S3 \filepath{pass_at_1}. \\
Warmup & Valid non-S3 records.  Grouped runs can contribute warmup ordinals
beyond 1 and 2; warmup is not the S3 headline outcome. \\
\filepath{pass_at_1} & Result-record field used by
\filepath{scripts/analyze_confirmatory.py} for S3 outcomes. \\
Pass@1 & Fraction of executable sessions whose first completed attempt passes
the configured verification suite.  The paper states that Pass@1 excludes
timeout-then-pass records. \\
\filepath{final_passed} & Final executable oracle result for a record. \\
TaskSuccess & Fraction of sessions whose final patch satisfies the task oracle. \\
V1 trap universe & The v1 confirmatory fold uses 22 S3 traps, each the third
session of a three-session sequence. \\
V2 trap universe & The scaled v2 fold uses 60 S3 traps, each evaluated over
three seeds. \\
Complete V1 S3 cells & 66 seed-trap S3 task cells for a complete v1 condition
(22 sequences \(\times\) 3 seeds). \\
Complete V2 S3 cells & 180 seed-trap S3 task cells for a complete v2 condition
(60 sequences \(\times\) 3 seeds). \\
Seed-trap S3 cell & Paired unit for pooled McNemar comparisons, keyed in the
analyzer as \filepath{seed<seed>:<sequence_id>}. \\
Warmup denominator & 360 valid non-S3 records for a complete v2 condition
(60 sequences \(\times\) 3 seeds \(\times\) S1/S2). \\
Hygiene denominator & 90 unique contributing result files per complete v2
condition; the analyzer reports metric means and count sums over files that
contributed at least one deduped record. \\
Validity gate & Excludes records whose \filepath{error_type} contains
\filepath{task_exception}, \filepath{codex_exec_failed}, or
\filepath{isolation_unavailable}. \\
Exclusion & Record-level validity-gate exclusion.  Exclusions are not
automatically lost S3 cells, and record-level excluded counts need not sum to
the S3 denominator. \\
Deduplication & The analyzer keeps the newest record per
\filepath{(condition, seed, oracle_id)} by the timestamp embedded in the
result-directory name. \\
Recall-verbatim & Trap class where the oracle-required token, format, or rule
is literally present in prior injected event text that B5 stores.  This labels
the evidence location, not solution-patch copyability; implementation is still
required. \\
Synthesis slice & Separate DreamBench-SWE-Synth directional slice.  It is not
pooled with the v1 or v2 confirmatory fold. \\
\bottomrule
\end{longtable}
\endgroup

\FloatBarrier
\clearpage
\subsection{Condition Codes}

\begin{table}[htbp]
\centering
\scriptsize
\resizebox{0.97\textwidth}{!}{%
\begin{tabular}{p{0.14\linewidth}p{0.24\linewidth}p{0.54\linewidth}}
\toprule
\textbf{Code} & \textbf{Canonical label} & \textbf{Definition} \\
\midrule
B0 & No external memory & No external memory reads or writes. \\
B1 & Raw episodic retrieval & Retrieves raw prior trajectories. \\
B2 & Vector trace retrieval & Deterministic token-vector retrieval over prior
trace chunks. \\
B3 & Reflection-only vector retrieval & Free-form verbal reflection lessons
after sessions, retrieved by vector similarity. \\
B4 & Untyped summary memory & Deterministic untyped session summaries. \\
B5 & Verbatim event-memory substitute & Offline instance-memory substitute;
one deterministic verbatim injected-event memory per trajectory; not stock
Mem0. \\
B5-MEM0 & Pinned hosted Mem0 & Hosted Mem0 baseline using
\texttt{mem0ai} 2.0.11, the hosted platform, default extraction settings, and
\filepath{memory_context=6}. \\
B5-MEM0-LIT & Pinned hosted Mem0 literal diagnostic & The same hosted Mem0
system and retrieval settings with fact inference disabled and sanitized raw
event text stored directly. \\
B6 & Subtask memory & Subtask-level memory over trajectory actions; inspired
by, but not a fidelity implementation of, structurally aligned subtask memory. \\
B7 & Task-tracker memory & Sequence-local task-tracker memory. \\
DF & reference-probe typed-only & Typed consolidation, contradiction repair,
counterfactual replay, local maintenance, and retrieval gating; no raw-evidence
capsule. \\
DF-raw-only & reference-probe raw-only & Typed consolidation disabled; raw-evidence
capsules enabled; contradiction records excluded from implementation reads. \\
DF-hybrid & reference-probe hybrid & DF plus raw-evidence capsules; contradiction
records excluded from implementation reads. \\
DF-strict & reference-probe strict & DF with \(\theta_{\mathrm{admit}}=1.5\), read
limit 3, and read-token budget 600. \\
DF-strict-hybrid & reference-probe strict-hybrid & DF-hybrid with the same strict
retrieval settings as DF-strict. \\
A0 & Episodic-only & Raw episode writes only; no typed consolidation, repair,
or replay. \\
A2 & No contradiction repair & Contradiction repair disabled. \\
A4 & No counterfactual replay & Counterfactual replay disabled. \\
A5 & No stale suppression & Read-side stale and superseded suppression disabled
while repair remains enabled. \\
A6 & No retrieval hard gate & Retrieval hard gate disabled; memory admission is
score-ranked. \\
A11 & Forced consolidation & Global maintenance scope forced during
consolidation. \\
\bottomrule
\end{tabular}
}%
\caption{Condition-code glossary from the condition contract and artifact appendix.}
\label{tab:notation-conditions}
\end{table}

\begin{table}[htbp]
\centering
\small
\begin{tabular}{p{0.18\linewidth}p{0.72\linewidth}}
\toprule
\textbf{Designed-not-run ablation} & \textbf{Meaning in this fold} \\
\midrule
A1 & No typed consolidation. \\
A3 & No causal failure extraction. \\
A7 & Similarity-only retrieval. \\
A8 & No provenance gate. \\
A10 & No raw episodic fallback at read time. \\
A12 & No scope fields. \\
A13 & No validity intervals. \\
\bottomrule
\end{tabular}
\caption{Ablation labels mentioned as designed-but-not-run in the reported fold.}
\label{tab:notation-designed-not-run}
\end{table}

\FloatBarrier
\clearpage
\subsection{Repairs and Operators}

\begin{table}[htbp]
\centering
\small
\begin{tabular}{p{0.25\linewidth}p{0.67\linewidth}}
\toprule
\textbf{Term} & \textbf{Definition} \\
\midrule
R1 & Adds a verbatim raw-evidence capsule: one \texttt{EPISODIC} memory per
episode carrying \filepath{injected_memory_event.content} through the normal
retrieval gate. \\
R1b & Exempts \texttt{EPISODIC} records from contradiction repair so raw
evidence is not rewritten or deleted by repair. \\
R2 & Excludes \texttt{CONTRADICTION}-type records from implementation reads
through the existing \filepath{allowed_types} gate. \\
Typed consolidation & Writes candidate semantic-project, procedural,
human-feedback, constraint, and episodic-derived memories with scope,
provenance, confidence, utility, risk, staleness, tags, and write reason. \\
Causal failure extraction & Writes failure memories grounded in failed
trajectories, command output, tests, diffs, logs, or feedback. \\
Counterfactual replay & Produces dream artifacts and replay-derived failure or
suppression decisions, accepted only when grounded in executable evidence, raw
trace evidence, or human-audited causal support. \\
Contradiction repair & Creates contradiction records, supersession links,
status transitions, and validity-interval updates over derived memories. \\
Stale suppression and forgetting & Applies \texttt{stale},
\texttt{superseded}, or logical \texttt{deleted} status for derived memories,
blocking retrieval without deleting raw evidence. \\
Provenance gate & Marks memories \texttt{active} or
\filepath{requires_review} based on whether the derived claim is supported by raw
episodes. \\
Raw episodic fallback & Agent-visible raw-evidence capsule while preserving raw
episodes for audit. \\
Metadata retrieval gate & Admission by status, scope, provenance, confidence,
staleness, risk, and supersession metadata. \\
\bottomrule
\end{tabular}
\caption{Repair labels and sleep/read operators used by reference-probe variants.}
\label{tab:notation-operators}
\end{table}

\FloatBarrier
\clearpage
\subsection{Construct Labels}

\begin{table}[htbp]
\centering
\scriptsize
\resizebox{0.97\textwidth}{!}{%
\begin{tabular}{L{0.08\linewidth}L{0.26\linewidth}L{0.58\linewidth}}
\toprule
\textbf{ID} & \textbf{Name} & \textbf{Definition} \\
\midrule
C1 & Verbatim retention & Preserve a non-inferable exact token, format, or
rule from one prior event and reproduce it when relevant. \\
C2 & Retrieval precision under interference & Select the one relevant stored
fact among many similar stored facts under a fixed read-token budget. \\
C3 & Staleness detection and supersession & Use the superseding fact and
suppress the stale fact, even when the stale fact is more salient. \\
C4 & Update propagation / regression avoidance & Propagate a changed fact to a
derived later location that the superseding event did not directly patch. \\
C5 & Scope discipline & Apply feedback inside its stated scope and refuse to
apply it outside that scope. \\
C6 & Contradiction handling / provenance conflict & Resolve conflicting sources
by an established authority or precedence rule rather than recency or arbitrary
choice. \\
C7 & Cross-session synthesis / paraphrase-retention application & Derive the
answer from two or more events, or apply a paraphrased retained instruction,
with no single event containing the decisive S3 literal. \\
C8 & Procedural / source-of-truth memory & Remember how this repository expects
a change to be made, such as editing a generator rather than a generated
artifact. \\
C9 & Spurious-lesson rejection / disconfirmation & Reject an unsupported or
later-disconfirmed lesson, even when stored memory suggests a wrong causal
story. \\
C10 & Abstention / confabulation resistance & Recognize that no prior event
bears on the current task and avoid fabricating benchmark-style tokens or
importing irrelevant memories. \\
\bottomrule
\end{tabular}
}%
\caption{Construct labels from the construct-taxonomy appendix.
V2 construct-profile outcomes are reported separately by the canonical analyzer
artifacts; this table defines the labels.}
\label{tab:notation-constructs}
\end{table}

\FloatBarrier
\clearpage
\subsection{Metrics and Hygiene Terms}

\begingroup
\scriptsize
\begin{longtable}{@{}L{0.20\linewidth}L{0.43\linewidth}L{0.14\linewidth}L{0.09\linewidth}@{}}
\caption{Task, hygiene, and cost metric glossary.  Hygiene values in the
confirmatory fold are deterministic offline re-scores from result records; the
paper treats hygiene claims as diagnostic because several metrics are
task-coupled, collinear, or judge-dependent.}
\label{tab:notation-metrics}\\
\toprule
\textbf{Metric or count} & \textbf{Definition} & \textbf{Fold handling} &
\textbf{Direction} \\
\midrule
\endfirsthead
\caption[]{Task, hygiene, and cost metric glossary, continued.}\\
\toprule
\textbf{Metric or count} & \textbf{Definition} & \textbf{Fold handling} &
\textbf{Direction} \\
\midrule
\endhead
TaskSuccess & \(\sum_i \mathrm{TaskSuccess}_i/N\); in prose, the fraction of
sessions whose final patch satisfies the task oracle. & Task outcome & Higher \\
Pass@1 / \filepath{pass_at_1} & \(\sum_i \mathrm{Pass@1}_i/N_{\mathrm{exec}}\);
the analyzer uses S3 \filepath{pass_at_1} as the headline outcome. & Primary S3
outcome & Higher \\
\filepath{final_passed} & Final executable oracle result. & Reported task field
& Higher \\
\metricbreak{UsefulMemory}{Precision} & \(\sum_i |U_i|/\sum_i |R_i|\): useful admitted
memories divided by admitted memories.  The label-grounded scorer also describes
the v2 form as useful retrieved memories divided by all retrieved memories for
human-feedback opportunities. & Hygiene panel & Higher \\
\metricbreak{HarmfulMemory}{Rate} & \(\sum_i |H_i|/\sum_i |R_i|\): harmful admitted
memories divided by admitted memories.  The v2 scorer definition is harmful
memory retrieved into active context and acted on. & Diagnostic in this fold;
0.000 for every ladder condition & Lower \\
\metricbreak{RepeatedError}{Rate} & Count of repeated error events divided by
opportunities to avoid repetition.  The v2 scorer definition is that the same
labeled error appears before the active session and recurs in the active patch.
& Hygiene panel & Lower \\
\metricbreakthree{StaleMemory}{Activation}{Rate} & Count of stale or superseded memories used
divided by tasks with stale memory available. & Tracked diagnostically; not a
standalone confirmatory-fold metric in the reported hygiene panel & Lower \\
\metricbreak{RegressionAfter}{Update} & Count of memory-update regressions divided by
memory updates with future dependency.  In the reported fold it is a diagnostic
heuristic that scans oracle stdout/stderr for regression-indicating substrings
on failed records. & Hygiene panel; diagnostic, not a headline hygiene win &
Lower \\
\metricbreakthree{Contradiction}{Repair}{Accuracy} & Count of correct contradiction repairs
divided by evaluable contradictions; the problem formulation defines correctness
as expected old-item status, new-item status, and relation all correct. & Hygiene
panel; collinear with HFUA and TransferScore for ladder conditions in this fold
& Higher \\
\metricbreakthree{HumanFeedback}{Use}{Accuracy} & V2 scorer definition: useful in-scope
feedback retrieved and reflected in the active patch. & Hygiene panel;
collinear with CRA and TransferScore for ladder conditions in this fold &
Higher \\
\metricbreak{Scope}{Accuracy} & Count of retrieved memories whose scope matches task
repo/file/symbol scope divided by retrieved memories; the v2 scorer notes it as
a mixed aggregate of human-feedback use and harmful-memory scope behavior. &
Hygiene panel & Higher \\
\metricbreak{Transfer}{Score} & \(\mathrm{mean}_j(\mathrm{outcome\_with\_prior\_memory}_j
- \mathrm{outcome\_without\_prior\_memory}_j)\); the v2 scorer definition is
that the S3 patch uses useful in-scope memory with provenance before S3, not
gated by \filepath{final_passed}. & Hygiene panel; collinear with CRA and HFUA
for ladder conditions in this fold & Higher \\
\metricbreakthree{AdmittedMemory}{TokensPerTask}{(AdmTok)} & Mean admitted memory tokens per
task, averaged by the analyzer as \texttt{AdmTok/task}. & Hygiene panel count
summary & Lower is not asserted; cost and context diagnostic \\
\texttt{contaminated} & Conservative scanner diagnostic over contributing
result files, not a validity-gate exclusion and not a claim that hidden
benchmark files were read. & Count sum & Diagnostic \\
\filepath{retrieved_memories} & Count of retrieved memories summed over
contributing result files by the analyzer. & Count sum & Diagnostic \\
\filepath{sleep_writes} & Count of sleep memory writes summed over contributing
result files by the analyzer. & Count sum & Diagnostic \\
\texttt{TotalTokens} & Wake tokens plus sleep tokens plus judge tokens. & Cost
frontier & Lower for cost \\
\metricbreakthree{CostPer}{Successful}{Task} & Total cost divided by successful tasks. & Cost
frontier & Lower for cost \\
\metricbreak{SleepCost}{Share} & Sleep cost divided by total cost. & Cost diagnostic &
Lower is not asserted; cost diagnostic \\
\metricbreak{Memory}{Bloat} & Total memory tokens divided by useful memory tokens. &
Impl. metric; not a reported independent win & Lower \\
\bottomrule
\end{longtable}
\endgroup

\FloatBarrier
\clearpage
\subsection{Clustered and Paired-Test Terms}

\begin{table}[htbp]
\centering
\small
\begin{tabular}{p{0.25\linewidth}p{0.67\linewidth}}
\toprule
\textbf{Term} & \textbf{Definition} \\
\midrule
Pooled exact McNemar family & Secondary paired exact two-sided McNemar
comparisons over S3 seed-trap cells, with Holm correction across P1--P6. \\
Supplemental McNemar comparison & A comparison rendered by the analyzer but not
part of the pre-registered Holm-corrected McNemar family; it does not change the
family's adjusted \(p\)-values. \\
Paired \(n\) & Number of S3 seed-trap cells present for both conditions in a
comparison after the validity gate. \\
\(b\) & Discordant-pair count where the first condition passes and the second
condition fails. \\
\(c\) & Discordant-pair count where the second condition passes and the first
condition fails. \\
Discordant pairs & \(b+c\), the paired cells where the two conditions differ. \\
Exact \(p\) & Two-sided exact McNemar \(p\)-value computed from \(b\) and \(c\). \\
Holm \(p\) & Holm-adjusted \(p\)-value for the comparison family. \\
Pooled McNemar & Secondary seed-trap-cell sensitivity analysis; it cannot
replace the primary trap-clustered P1--P6 family. \\
Per-seed McNemar & Exact two-sided McNemar run within each seed; Holm adjustment
is applied separately within each seed's six-comparison P1--P6 family. \\
Trap-majority collapse & Cluster sensitivity variant that requires all three
seed outcomes per condition for a trap, majority-votes each condition at the
trap level, then runs exact two-sided McNemar over complete trap pairs with Holm
adjustment across the six collapsed P1--P6 comparisons. \\
Trap-clustered sign/permutation test & Cluster sensitivity variant that, for
each trap, computes \(d_t=\#\mathrm{first\mbox{-}only\ seeds}
-\#\mathrm{second\mbox{-}only\ seeds}\) over available paired seeds and uses
exact sign-flip dynamic programming over the nonzero trap contributions (up to
60 traps in v2).  The two-sided statistic is
\(\left|\sum_t d_t\right|\), with Holm adjustment across the six clustered
P1--P6 comparisons.  This is the primary inferential family. \\
Failure to reject & Statistical conclusion used for the non-significant
reference-probe hybrid-vs.-B5 comparison; it is not an equivalence claim. \\
\bottomrule
\end{tabular}
\caption{Paired and clustered sensitivity terminology from the confirmatory and
clustered-sensitivity analyzers and Section~\ref{sec:results-mcnemar}.}
\label{tab:notation-clustered}
\end{table}

\paragraph{Successor comparison terms.}
In the v2.1 audit, \emph{Family A} is the fixed six-slot discrimination family
comparing registered memory-bearing conditions with B0.  \emph{Family B} is the fixed
two-slot cross-configuration mechanism family.  A \emph{conformance-unavailable} slot
is a condition rejected before evaluation; it remains in its registered Holm family at
$p=1$ and is neither an observed null nor an observed loss.  A \emph{secondary}
comparison is unprotected for a confirmatory superiority claim and must be read with its
trap-majority and per-seed sensitivities.  A \emph{temporal-drift band} is a diagnostic
tolerance around a historical control rate, not an equivalence margin.

%% file: sections/appendix_ethics_impact.tex
\FloatBarrier
\clearpage
\section{Ethics, Broader Impact, and Benchmark Integrity}
\label{app:ethics-impact}

\paragraph{Benchmark-contamination resistance.}
\bench{} is designed so that decisive S3 literals cannot be learned from public
artifacts or pretraining corpora that contain only those artifacts.  The hidden
contracts use arbitrary tokens generated by a CSPRNG at injection time.  The
generator records provenance classes, skeleton identifiers, and run metadata, but not
the generated values themselves.  Public task prompts, paper examples, and the public
artifact package therefore do not contain the exact hidden tokens needed by the scored
oracles.  For token-dependent traps, an agent that has not received the relevant
earlier-session event through the benchmark memory channel cannot infer the literal
answer from the visible S3 repository state except by guessing a fresh high-entropy
string.  This is the benchmark's integrity property: memorizing the released benchmark
description, code, or public fixture text is insufficient to solve the hidden-token
portion of the benchmark, because the load-bearing strings are sampled after those
public artifacts are fixed and are never released in them.

This property is narrower than a general anti-cheating guarantee.  It does not prove
network isolation, does not protect a private reviewer package after disclosure, and
does not prevent contamination if hidden oracle or reference-solution files are
accidentally mounted into the agent environment.  Those risks are handled separately by
the filesystem-isolation protocol in Section~\ref{sec:isolation-boundary} and by the
artifact policy in Appendix~\ref{app:artifact}.  The precise claim is that public
training-set leakage of the benchmark paper, harness, fixture repositories, and
redacted task records should not reveal the fresh CSPRNG literals that determine
success on the hidden-token traps.

\paragraph{Dual-use scope.}
The work measures memory hygiene for software-engineering agents: whether an agent
retains, retrieves, updates, scopes, and suppresses information from its own prior
coding sessions.  Better memory maintenance can improve benign software maintenance,
debugging continuity, and reproducibility of long-horizon agent workflows.  The same
capability is dual-use in the ordinary sense that more capable coding agents can also
be used for unwanted automation.  The benchmark does not evaluate exploitation,
malware, credential theft, surveillance, or social-engineering behavior, and it should
not be read as a safety evaluation for those domains.

The evaluated tasks use synthetic fixture repositories and synthetic reviewer or
maintainer events.  The hidden tokens are artificial benchmark strings, not real
credentials.  The study does not require personal data, user conversations, production
repositories, or private customer code.  The public artifact policy excludes hosted
service credentials, local agent state, raw logs, and hidden scoring assets; selected
examples are redacted or synthetic when exact-token mechanisms are discussed.  As a
result, the main privacy risk is not subject-data exposure, but accidental artifact
leakage of hidden benchmark answers or local operational material.  The packaging and
isolation checks are intended to make that risk auditable.

\paragraph{Compute cost and environmental honesty.}
The reported reference-probe configuration still requires explicit compute accounting.  The v2
cost artifact \filepath{analysis/fold/v2_cost_frontier.json} estimates reference-probe hybrid at
0.03832 USD per successful task, B5 at 0.03953, and reference-probe raw-only at 0.03556.  These
figures are diagnostic constraints, not grounds for a favorable cost claim, and the
primary task comparison remains null.

We also do not report carbon emissions or energy consumption, because the experiment
uses hosted model endpoints and the repository does not contain verified power draw,
hardware utilization, datacenter energy mix, or per-request energy accounting.  Converting
token counts into emissions would create false precision.  The honest environmental
statement is therefore qualitative and operational: sleep-phase maintenance materially
increases model-call volume; this cost limits the practical claim; and any future
deployment-oriented version should report token budgets, call counts, wall-clock
runtime, and, where available, provider-backed energy or emissions estimates before
claiming environmental efficiency.

\paragraph{Hosted-Mem0 limitation.}
B5-MEM0 and B5-MEM0-LIT are two pinned hosted configurations of one Mem0
system family, not a ranking claim against Mem0 as a system family.  Both use
the hosted API through \texttt{mem0ai} 2.0.11 with the fixed harness settings
described in the paper, including \filepath{memory_context=6}; B5-MEM0-LIT
disables fact inference for the exact-literal diagnostic.  Their value in this paper is failure analysis: audited rows
show relevant memory retrieval followed by exact-token corruption, which is informative
for \bench{} because the hidden oracles require byte-level preservation of arbitrary
tokens.

That result should not be generalized to tuned Mem0, Mem0$^g$, self-hosted deployments,
different extraction prompts, different graph or vector settings, larger retrieval
contexts, or variants configured explicitly for verbatim byte preservation.  The paper's
claim is only that these pinned hosted configurations fail these exact-token traps under
the reported harness.  A practical comparison against tuned Mem0 would require a new
pre-registered evaluation with configuration search rules, cost accounting, and the same
benchmark-integrity constraints applied to all systems.

The successor audit changes none of those disclosure boundaries.  It admits only the
literal-storage configuration, reports native Mem0 and both Supermemory conditions as
pre-evaluation conformance-unavailable, and publishes only path-neutral aggregate
evidence.  The literal configuration's operational profile includes hosted-service calls
and therefore depends on provider availability, mutable service behavior, and nonzero
resource use.  Those dependencies are disclosed as limitations and are not converted
into production-readiness or environmental-efficiency claims.